\documentclass[11pt,journal]{article}
 \pdfoutput=1
\usepackage{mdwmath}
\usepackage{mdwtab}
\usepackage{amsmath,amssymb}
\usepackage{graphicx}
\usepackage{geometry}
\usepackage{subfig}
\newcommand{\argmax}{\operatornamewithlimits{arg\,max}}
\newcommand{\argmin}{\operatornamewithlimits{arg\,min}}
\DeclareMathOperator{\diag}{diag}
\providecommand{\abs}[1]{\lvert#1\rvert}

\providecommand{\diag}{diag}
\providecommand{\norm}[1]{\lVert#1\rVert}

\def\R{\mathbb{R}}
\def\N{\mathcal{N}}

\def\GP{\mathcal{GP}}

\newtheorem{theorem}{Theorem}

\begin{document}
\title{Gaussian Process Morphable Models} 



\author{Marcel L{\"u}thi, Christoph Jud, Thomas Gerig and Thomas Vetter \\
  \\[5mm]
  \small Department of Mathematics and Computer Science, University of Basel \\
  \small \ttfamily\{marcel.luethi, christoph.jud, thomas.gerig, thomas.vetter\}@unibas.ch
}
\date{~}




\maketitle

\begin{abstract}
  Models of shape variations have become a central component for the
  automated analysis of images. Statistical shape models (SSMs) represent a class of shapes as a normal distribution of point variations, whose
  parameters are estimated from example shapes. Principal component analysis (PCA) is applied to obtain a low-dimensional
  representation of the shape variation in terms of the leading principal components.
  In this paper, we propose a generalization of SSMs, which we refer
  to as \emph{Gaussian Process Morphable Models} (GPMMs). We model the
  shape variations with a Gaussian process, which we represent using
  the leading components of its Karhunen-Lo\`eve expansion. To compute the expansion, we make use of an approximation scheme
  based on the Nystr\"om method. The resulting model can be seen as a continuous analogon of a standard SSM. However,
  while for SSMs the shape variation is restricted to the linear span
  of the example data, with GPMMs we can define the shape variation using any Gaussian process. For example, we can build shape models
  that correspond to classical spline models, and thus do not require any example data. Furthermore, Gaussian processes make it possible
  to combine different models. For example, an SSM can be extended with a spline model, to obtain a model that
  incorporates learned shape characteristics, but is flexible enough
  to explain shapes that cannot be represented by the SSM. 

  We introduce a simple algorithm for fitting a GPMM to a surface or image.
  This results in a non-rigid registration approach, whose
  regularization properties are defined by a GPMM. We show how we can obtain
  different registration schemes, including
  methods for multi-scale, spatially-varying or hybrid registration,
  by constructing an appropriate GPMM. As our approach strictly separates modelling from the fitting process, this is all achieved without changes to the fitting algorithm. 

  We show the applicability and versatility of GPMMs on a clinical use
  case, where the goal is the model-based segmentation of 3D forearm
  images. In a first experiment we show how we can build GPMMs that are specially tailored to the task of forearm registration. We demonstrate how GPMMs can be used to advance standard Active Shape
  Models based segmentation. Finally, we show the application of GPMMs
  for the problem of 3D image to image registration.  To complement the paper, we
  have made all our methods available as open source.
\end{abstract}




\section{Introduction}

The automatic interpretation and analysis of objects in an image is
the core of computer vision and medical image analysis. A popular
approach is analysis by synthesis \cite{grenander2007pattern}, which asserts that in order to explain an image, we need to be able to synthesize its
content. This is achieved by fitting a probabilistic model to an
image, such that one-to-one correspondence between the model and the
image is achieved. The image can then be explained using the model
information. The better the model represents the structure of the
objects to be analyzed, the easier it becomes to fit the model. For
this reason statistical shape models have become very popular. The
most important examples of statistical shape models are the Active
Shape Model \cite{cootes_active_1995} and the Morphable model
\cite{blanz_morphable_1999}, which learn the shape variation from given
training examples, and represent the shape variation using the leading principal components. In the following, we refer to this type of model as PCA-based statistical shape models, or in short SSMs. These models are linear, parametric models and
hence are mathematically convenient and easy to incorporate in
image-analysis algorithms. Since they can represent only shapes that
are in the linear span of the given training examples, they lead to
algorithms that are robust towards artifacts and noise. The
downside of this specificity is that to learn a model that can
express all possible target shapes, a lot of training data is needed.

The main contribution of this work is that we introduce a
generalization of SSMs, which we refer to as Gaussian Process
Morphable Models (GPMM). We model 
a shape as a deformation $u$ from a reference shape $\Gamma_R \subset \R^3$, 
i.e.\ a shape $s$ can be represented as
\[
s = \{x + u(x) | x \in \Gamma_R\}
\]
for some deformation $u : \Omega \to \R^3$, with $\Omega \supseteq \Gamma_R$.  We model the
deformation as a Gaussian process $u \sim GP(\mu, k)$ where $\mu :
\Omega \to \R^3$ is a mean deformation and $k : \Omega \times \Omega \to 
\R^{3 \times 3}$ a covariance function or kernel. The core idea behind our approach is that we
obtain a parametric, low-dimensional model by representing the Gaussian process using the $r$ 
leading basis function $\phi_i : \Omega \to \R^3$ of its Karhunen-Lo{\`e}ve expansion:
\begin{equation} \label{eq:basic-equation}
u =  \mu + \sum_{i=1}^r \alpha_i \sqrt{\lambda_i}\phi_i, \; \alpha_i \in \mathcal{N}(0,1)
\end{equation}
(here, $\lambda_i$ is the variance associated with each basis function
$\phi_i$). As we usually assume strong smoothness of the deformation
when modelling shapes, it is often possible to achieve good
approximations using only a few leading eigenvalues, which makes the
representation practical. The main difficulty of this approach is to
efficiently compute the leading eigenfunction/eigenvalue pairs. To this end, we propose to 
use a Nystr\"om approximation, and make use of a recently introduced computational approach, which is able to use a large number of input points for computing the approximation  \cite{li_making_2010}. We study the
approximation properties of the numerical method and provide a detailed discussion of how the approximation quality is
influenced by different chooses of covariance functions. Furthermore, we discuss how to chose the parameters of our method in order to reach a given approximation quality.

The biggest advantage of GPMMs compared to SSMs is that we have much more freedom in defining the covariance
function. As a second main contribution we will show in
Section~\ref{sec:modeling-with-kernels} how expressive prior models
for registration can be derived, by leveraging the modeling power of
Gaussian processes. By estimating the covariances from example data
our method becomes a continuous version of an SSM. When we have no or
only little training data available, arbitrary kernel functions can be
used to define the covariances. In particular we can define models of
smooth deformations using spline models or radial basis functions,
which are frequently used in registration approaches. We show how a
simple registration approach, whose regularization properties are
defined in terms of a GPMM, allows us to use these models for actual
surface and image registration. Besides these simple models, GPMMs
also make it possible to combine different covariance functions (or
kernels) to mimic more sophisticated registration schemes. We show
how to construct priors that have multi-scale properties, are
spatially-varying or incorporate landmark constraints. We will also show how 
to combine models learned from training data with analytically defined covariance functions, 
in order to increase the flexibility of SSMs in cases where not sufficient training data is available. 
Although in contrast to SSMs, GPMMs  model deformations defined on a continuous domain,  we can always discretize it to obtain a model that is mathematically equivalent to an SSM. This makes it possible to leverage the modelling flexibility of GPMMs also in classical shape modelling algorithms, such as for example the Active Shape Model fitting \cite{cootes_active_1995} algorithm or the coherent point drift method \cite{myronenko2010point}.

We have implemented our method for modeling with Gaussian processes as
part of the open source software statismo
\cite{luthi_statismo-framework_2012} and scalismo \cite{scalismo}. We perform registration experiments on a real world 
data-set of forearm images with associated ground-truth segmentations. In a first experiment we show how GPMMs can be specially tailored to the task of forearm registration, and perform surface-to-surface registration experiments with different models. In the second experiment we show how Active Shape Model fitting can be improved by using GPMMs as a shape prior. In the last experiment, we present an application of GPMMs for 3D image-to-image registration and compare the result to the popular B-spline registration method implemented in Elastix \cite{klein2010elastix}.

\subsection{Related work}

Our work can be seen as the unification of two different concepts: On one hand we extend SSMs, 
such that they become more expressive, on the other hand we model prior distributions for surface and image registration. There are works from both the shape modelling and the registration community, which are conceptually similar or have the same goals as we pursuit with our approach. Most notably, the work of Wang and Staib \cite{wang_boundary_2000}, which aims for extending the flexibility of shape models, and the work by Grenander et al. \cite{grenander_computational_1998}, who use Gaussian processes as priors for registration are very close in spirit to our model.  
The idea of Wang and Staib is to extend the flexibility of a SSM by combining a learned covariance matrix used in a statistical shape model with covariance matrices that represent other, synthetic deformations. This corresponds exactly to our idea for combining covariance functions in the GP setting. However, their method requires that the full covariance matrix can be represented, which is only feasible for very coarsely discretized shapes. In contrast, our method yields a continuous representation, and allows for an arbitrarily fine discretization once the prior is evaluated in the final registration procedure.
On the registration side, the use of Gaussian processes for image registration has been extensively studied in the 90s by
Grenander et al.\ (see the overview article
\cite{grenander_computational_1998} and references therein). Similar
to our approach, they propose to use a basis function representation
to span the model space. However, in all these works the basis functions have to be known
analytically \cite{amit_structural_1991}, or the initial model needs
to be of finite rank \cite{joshi1997gaussian}. In our method we use of the Nystr\"om approximation to numerically approximate the leading eigenfunctions, which makes it possible to approximate any Gaussian process and thus to allow us to use arbitrary combinations of kernels in our models. We believe that this modelling flexibility is what makes this approach so powerful.

There are many other works that propose to model the admissible deformations for non-rigid registration by means of a kernel (i.e. as a Reproducing Kernel Hilbert Space (RKHS)). Especially for landmark based registration, spline based models and radial basis functions have been widely used
\cite{holden2008review}.  The algorithm for solving a standard spline-based landmark registration problem corresponds to the MAP
solution in Gaussian process regression \cite{rasmussen_gaussian_2006}. Using Gaussian process regression directly for image registration has been proposed by Zhu et al \cite{zhu2009nonrigid}. A similar framework for surface registration, where kernels are used for specifying the admissible deformation was proposed by Steinke et al.\ \cite{scholkopf_object_2005}. While they do not provide a
probabilistic interpretation of the problem, their approach results in
the same final registration formulation as our approach.
The use of Reproducing Kernel Hilbert Spaces for modeling admissible deformation also plays an important role for diffeomorphic image registration (see
e.g. \cite{younes2010shapes}, Chapter 9). In this context, it has also
been proposed to combine basic kernels for multi-scale
\cite{bruveris2012mixture,sommer2013sparse} and spatially-varying
models \cite{schmah2013left} for registration. However, the work focuses more on the mathematical and algorithmic aspects of 
enforcing diffeomorphic mappings, rather than modelling aspect.

Besides the work of Wang and Staib \cite{wang_boundary_2000} there have been many other works for extending the flexibility of SSMs.  This is typically achieved by adding artificial training data \cite{cootes_combining_1995} or by segmenting the model either spatially \cite{zhao2005novel,blanz_morphable_1999} or in the frequency domain \cite{davatzikos2003hierarchical,nain2007multiscale}. The use of Gaussian processes to model the covariance structure is much more general and subsumes all these methods. Another set of work gives shape model based algorithms more flexibility for explaining a target solution \cite{albrecht2008statistical,kainmueller2013omnidirectional,le_pdm-enlor:_2013}. Compared to our model, these approaches have the disadvantage that the model is not generative anymore and does not admit a clear probabilistic interpretation.

This paper is a summary and  extension of our previous conference publications \cite{luthi_using_2011,gerig2014spatially,luthi2013unified}.
It extends our previous work in several ways:
1) It provides an improved presentation of the basic method and in particular its numeric implementation. 2) It provides an analysis of the approximation properties of this scheme. 3) It proposes new combination of kernels to combine
statistical shape models and analytically defined kernels. 4) It features a more detailed validation including surface and image registration, as well as Active Shape Model fitting.




\section{Gaussian Process Morphable Models}

Before describing GPMMs, we summarize the main concepts of PCA-based statistical shape models, on which we will build up our work.

\subsection{PCA-based statistical Shape models}

PCA-based statistical shape models assume that the space of all possible shape deformation can be learned from a 
set of typical example shapes $\{\Gamma_1, \ldots, \Gamma_n\}$. Each shape $\Gamma_i$ is represented as a discrete set of landmark points, i.e.\
\[
\Gamma_i = \{x_k^i\,|\, x_k \in \R^3, k=1,\ldots, N\}, 
\]
where $N$ denotes the number of landmark points. In early approaches, the points typically denoted anatomical landmarks, and $N$ was consequently small (in the tens). Most modern approaches use a dense set of points to represent the shapes. In this case, the number of points is typically in the thousands.
The crucial assumption is that the points are in correspondence among the examples. 
This means that the $k$-th landmark point $x_k^i$ and $x_k^j$ of two shapes $\Gamma_i$ and $\Gamma_j$ represent the same anatomical point of the shape. These corresponding points are either defined manually, or automatically determined using a registration algorithm. 
To build the model a shape $\Gamma_i$ is represented as a vector $s_i \in \R^{3N}$, where the $x,y,z-$ components of each point are stacked onto each other:
\[
\vec{s_i} = (x_{1x}^i, x_{1y}^i, x_{1z}^i, \ldots, x_{Nx}^i, x_{Ny}^i, x_{Nz}^i).
\]
This vectorial representation makes it possible to apply the standard
multivariate statistics to model a probability distribution over
shapes. 
The usual assumption is that the shape variations can be modelled using a normal
distribution \[
s \sim \N(\mu, \Sigma)\]
 where the mean $\mu$ and
covariance matrix $\Sigma$ are estimated from the example data:
\begin{align}
\mu &= \overline{s} :=  \frac{1}{n} \sum_{i=1}^n \vec{s}_i   \\
\Sigma &= S := \frac{1}{n-1} \sum_{i=1}^n (\vec{s}_i - \overline{s} )(\vec{s}_i - \overline{s})^T.
\end{align}
As the number of points $N$ is usually large, the covariance matrix $\Sigma$ cannot be 
represented explicitly. Fortunately, as it is determined completely by the $n$ example data-sets, it has at most rank $n$ and can therefore be represented using $n$ basis vectors. This is achieved by performing a Principal Component Analysis (PCA) \cite{jolliffe2002principal}. In its probabilistic interpretation, PCA leads to a model of the form
\begin{equation}\label{eq:pca-repr}
s = \overline{s} + \sum_{i=1}^n \alpha_i \sqrt{d_i} \vec{u}_i 
\end{equation}
where $(u_i, d_i), \, i=1, \ldots, n$, are the eigenvectors and eigenvalues of the covariance
matrix $S$. Assuming that $\alpha_i \sim \N(0,1)$ in \eqref{eq:pca-repr}, it is easy to check that 
$s \sim \N(\overline{s}, S)$. Thus, we have a efficient, parametric representation of the distribution.

\subsection{Gaussian Process Morphable Models}\label{sec:gpmms}
The literature of PCA based statistical shape models usually
emphasizes the shapes that are modelled. Equation~\ref{eq:pca-repr}
however, gives rise to different interpretation: A statistical shape
model is a model of deformations $\vec{\phi} = \sum_{i=1}^n \alpha_i
\sqrt{d_i}\vec{u}_i \sim \N(0, S)$ which are added to a
mean shape $\overline{s}$. The probability distribution is on the
deformations. This is the interpretation we use when we generalize
these models to define Gaussian Process Morphable Models. We define a probabilistic model directly on the deformations. To stress that we are modelling deformations (i.e.\ vector fields defined on the reference domain $\Gamma_R$), and to become independent of the discretization, we model the deformations as a Gaussian process. 

Let $\Gamma_R \subset \R^3$ be a reference shape and denote by $\Omega \subset \R^3$ a domain, such that $\Gamma_R \subseteq \Omega$. We define a Gaussian process $u \in
\GP(\mu, k)$ with mean function $\mu : \Omega
\to \R^3$ and covariance function $k : \Omega \times \Omega \to
\R^{3 \times 3}$. Note that any deformation $\hat{u}$ sampled from
$\GP(\mu, k)$, gives rise to a new shape by warping the reference shape $\Gamma_R$:
\[
\Gamma = \{x + \hat{u}(x) \, |\, x \in \Gamma_R\}.
\]
Similar to the PCA representation of a statistical shape model used in (Equation~\ref{eq:pca-repr}), a Gaussian process $\GP(\mu, k)$ can be represented in terms of an orthogonal set of basis functions $\{\phi_i\}_{i=1}^\infty$
\begin{equation}\label{eq:kle}
	u(x) \sim \mu(x) + \sum_{i=1}^\infty \alpha_i \sqrt{\lambda}_i \phi_i(x), \, \alpha_i \in \N(0,1),
\end{equation}
where $(\lambda_i, \phi_i)$ are the eigenvalue/eigenfunction pairs of the integral operator
\[
\mathcal{T}_k f(\cdot) := \int_\Omega k(x,\cdot) f(x) \, d\rho(x), 
\]
where $\rho(x)$ denotes a measure.
The representation \eqref{eq:kle} is known as the Karhunen-Lo{\`e}ve expansion of the Gaussian process \cite{berlinet2004reproducing}. Since the random coefficients $\alpha_i$ are uncorrelated, the variance of $u$ is given by the sum of the variances of the individual components. Consequently, the eigenvalue $\lambda_i$ corresponds to the variance explained by the $i$-th component. 
This suggests that if the $\lambda_i$ decay sufficiently quickly, we can use the low-rank approximation
\begin{equation}\label{eq:low-rank}
\tilde{u}(x) \sim \mu(x) + \sum_{i=1}^r \alpha_i \sqrt{\lambda}_i \phi_i(x)
\end{equation}
to represent the process. The expected error of this approximation is given by the tail sum
\[
\sum_{i=r+1}^\infty \lambda_i.
\]
The resulting model is a finite dimensional, parametric model, similar to a standard statistical model. Note, however, that there is no restriction that the covariance function $k$ needs to be the sample covariance matrix. Any valid positive definite covariance function can be used. As we will show in Section~\ref{sec:modeling-with-kernels} this makes it possible to define powerful prior models, even when there is little or no example data available. 

\subsection{Computing the eigenfunctions}\label{sec:eigendecomposition}
The low-rank approximation \eqref{eq:low-rank} can only be performed if we are able to compute the eigenfunction/eigenvalue pairs $(\phi_i, \lambda_i)_{i=1}^r$. Although for some kernel functions analytic solutions are available (see e.g. \cite{amit_structural_1991,grenander_computational_1998}) for most interesting models we need to resort to numeric approximations. A classical method, which has recently received renewed attention from the machine learning community, is the Nystr\"om method \cite{rasmussen_gaussian_2006}. 
The goal of the Nystr\"om method is to obtain a numerical estimate for
the eigenfunctions/eigenvalues of the integral operator
\begin{equation}\label{eq:integral-operator}
\mathcal{T}_k f(\cdot) := \int_\Omega k(x,\cdot) f(x) d\rho(x).
\end{equation}
The pairs $(\phi_i, \lambda_i)$,
satisfying the equation
\begin{equation}\label{eq:eigenfunction-equation}
\lambda_i \phi_i(x') = \int_{\Omega} k(x, x') \phi_i(x) \,d\rho(x),  \; \forall x' \in \Omega
\end{equation}
are sought. The Nystr\"om method is intended to approximate the integral in \eqref{eq:eigenfunction-equation}. 
This can, for example, be achieved by letting $d\rho(x)=p(x)\,dx$ where $p(x)$ is a density function defined on the domain $\Omega$,
and to randomly
sample points $X = \{x_1, \ldots, x_n\}, x_l$ according to $p$. 
The samples $(x_l)_{l=1,\ldots, N}$ for $x'$ in \eqref{eq:eigenfunction-equation} lead to the matrix eigenvalue problem
\begin{equation}\label{eq:eigenvalue-problem-small}
K u_i = \lambda_i^{mat} u_i,
\end{equation}
where $K_{il}=k(x_i, x_l)$ is the kernel matrix, $u_i$ denotes the $i-$th eigenvector and $\lambda_i^{mat}$ the corresponding eigenvalue.
Note, that since the kernel is matrix valued ($k : \Omega \times \Omega \to \R^{d \times d}$), the matrices $K$ and $k_X$ are block matrices: $K \in \R^{nd \times nd}$ and $k_X \in \R^{nd \times d}$.
The eigenvalue $\lambda_i^{mat}$ approximates $\lambda_i$, while the eigenfunction $\phi_i$ in turn is approximated 
with
\begin{equation} \label{eq:nystrom-phi}
\tilde{\phi}_i(x) = \frac{\sqrt{n}}{\lambda_i^{mat}} k_X(x) u_i \approx \phi_i(x),
\end{equation}
where $k_X(x) = (k(x_1, x), \ldots ,k(x_n, x))$.

Clearly, the quality of this approximation improves with the number of points $n$, which are sampled
(see Appendix~\ref{sec:low-rank} for a detailed discussion). As $n$ becomes larger (i.e.\ exceeds a few thousand points),  deriving the eigenvalue
problem \eqref{eq:eigenvalue-problem-small} might still be
computationally infeasible. Following Li et al.\ \cite{li_making_2010}, we therefore
apply a random SVD \cite{halko_finding_2011} for efficiently
approximating the first eigenvalues/eigenvectors without having to
compute the eigenvalues of the full $N \times N$ matrix. Theoretical
bounds of the method \cite{halko_finding_2011}, as well as its
application for the Nystr\"om approximation \cite{li_making_2010} show
that it leads to accurate approximation for kernels with a fast decaying spectrum. For our application, the error induced by the random SVD is negligible compared to the  approximation error caused by the low-rank approximation and the Nystr\"om method. 

\subsection{Accuracy of the low-rank approximation}
It is clear that our method depends crucially on the quality of the
low-rank approximations. Ideally, we would like to see the low-rank model 
as a convenient reparametrization of the original process, which would not affect the shape variations that are spanned by our model. Unfortunately, and not surprisingly, this is not always the case. There are two sources of error: 1) There is no good representation of the Gaussian process in terms of the leading basis-functions 2) Even if there is, it might still happen that the numerical computation of the eigenvalues using the Nystr\"om method leads to approximation errors. 
In order not to digress from the main theme of the paper, we have put a detailed discussion of these issues into the appendix (cf. Appendix~\ref{sec:low-rank}), and summarize here only the main results.

Intuitively, the ability to represent a model using only a few leading
basis functions is dependent on how much the individual points of the
models correlate. This becomes clear when we consider the two extreme
cases: 1) All points are perfectly linearly correlated and 2) every
point can move independently. We can simulate these extreme using a
Gaussian kernel $k(x,x')=\exp(-\frac{\norm{x-x'}^2}{\sigma^2})$, where
we choose $\sigma^2 \to \infty$ to simulate the first case. In this
case all points are perfectly correlated and the process can be
represented using a single basis function. The second extreme case is
attained if we let $\sigma^2$ approach $0$ (i.e. the kernel becomes the $\delta$-function $\delta_{xx'}$). In this case, any point
can move independently, and to faithfully represent the process, we
need one basis function per point. From these considerations we can
see that the larger the assumed smoothness of the process (and hence
the correlation between the points), the fewer basis functions are
needed to approximate the process. It is also intuitively clear that
the Nystr\"om approximation yields much better results in the first
case. As there are strong correlations between the points, the
knowledge of the true eigenfunction value at the Nystr\"om points
$x_1, \ldots, x_n$ will also determine the value of the basis function
at the correlated points. This is no longer the case where we don't
have correlations. In this case, knowing the value at the points $x_1,
\ldots, x_n$ will not give us any information about the value of the
basis function at other points of the domain.



\section{Modeling with kernels} \label{sec:modeling-with-kernels}

The formalism of Gaussian processes provides us with a rich language to model shape variations. In this section we explore some of these modelling possibilities, with a focus on models that we find most useful in our work on model-based image analysis and in particular surface and image registration. Many more possibilities for modelling with Gaussian processes have been explored in the machine learning community (see e.g. Duvenaud, Chapter 2 \cite{duvenaud2014automatic}).

\begin{figure}
\center{\includegraphics[trim=2cm 2cm 2cm 2cm,width=0.28\columnwidth]{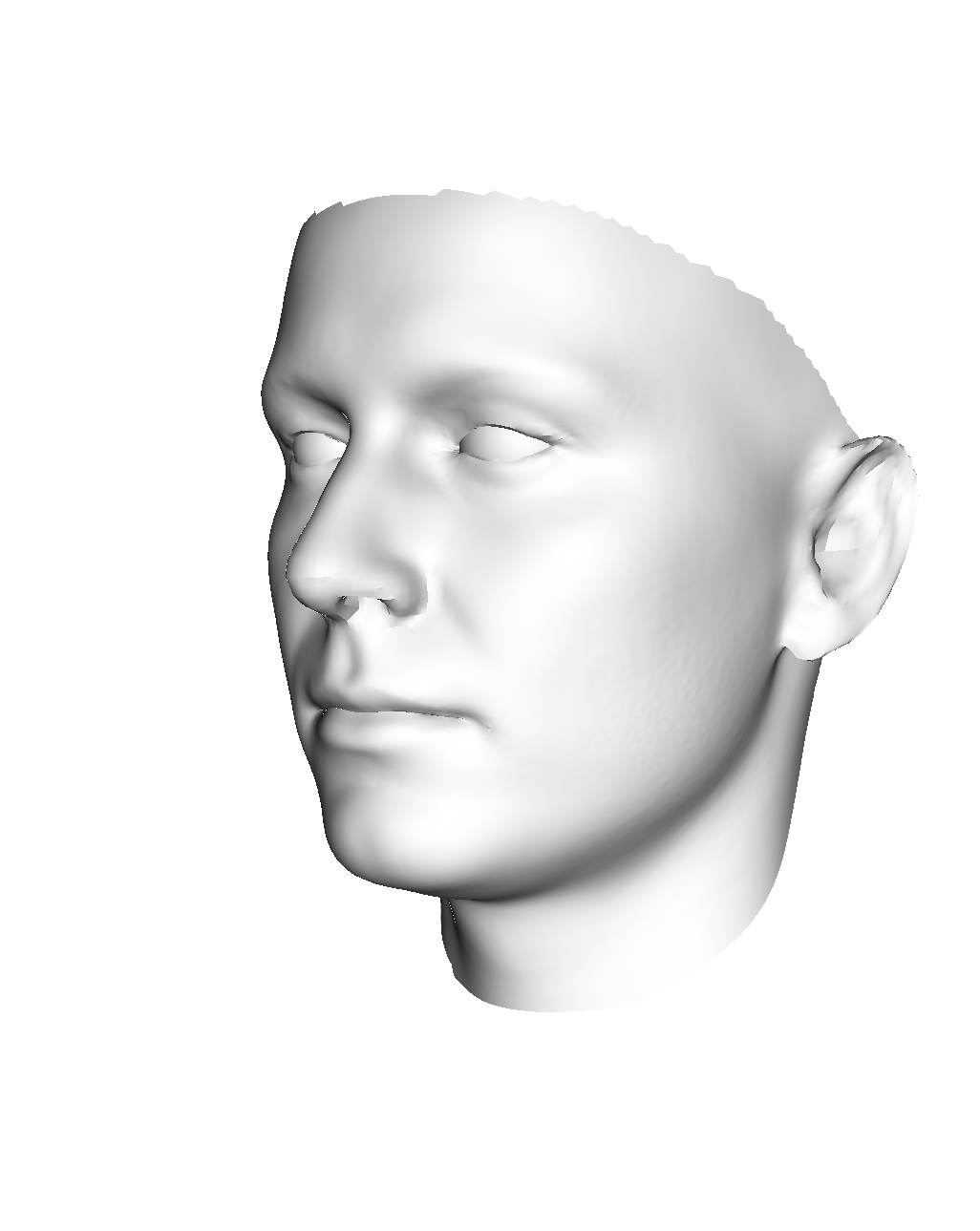}
\includegraphics[trim=2cm 2cm 2cm 2cm,width=0.3\columnwidth]{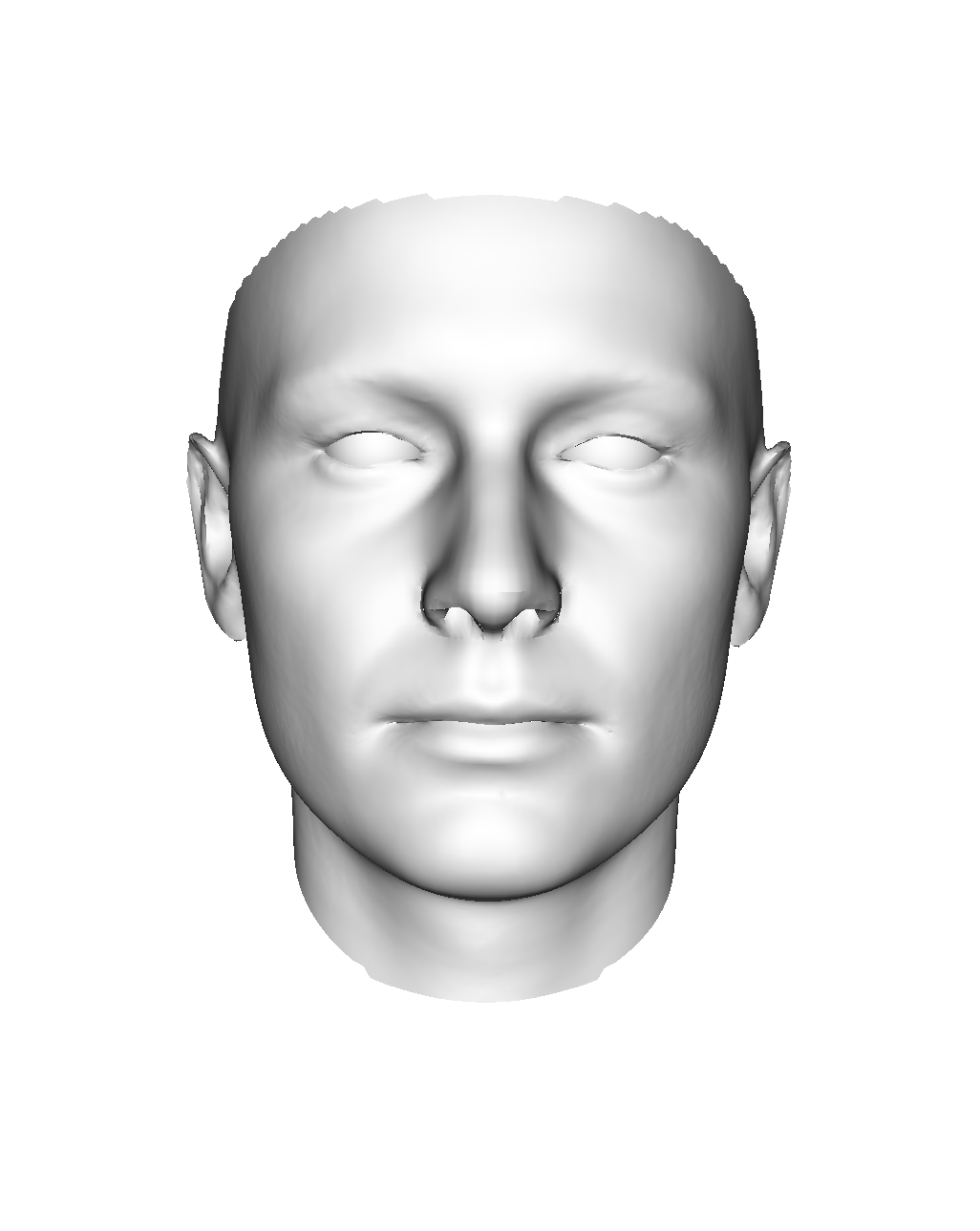}
\includegraphics[trim=2cm 2cm 2cm 2cm,width=0.28\columnwidth]{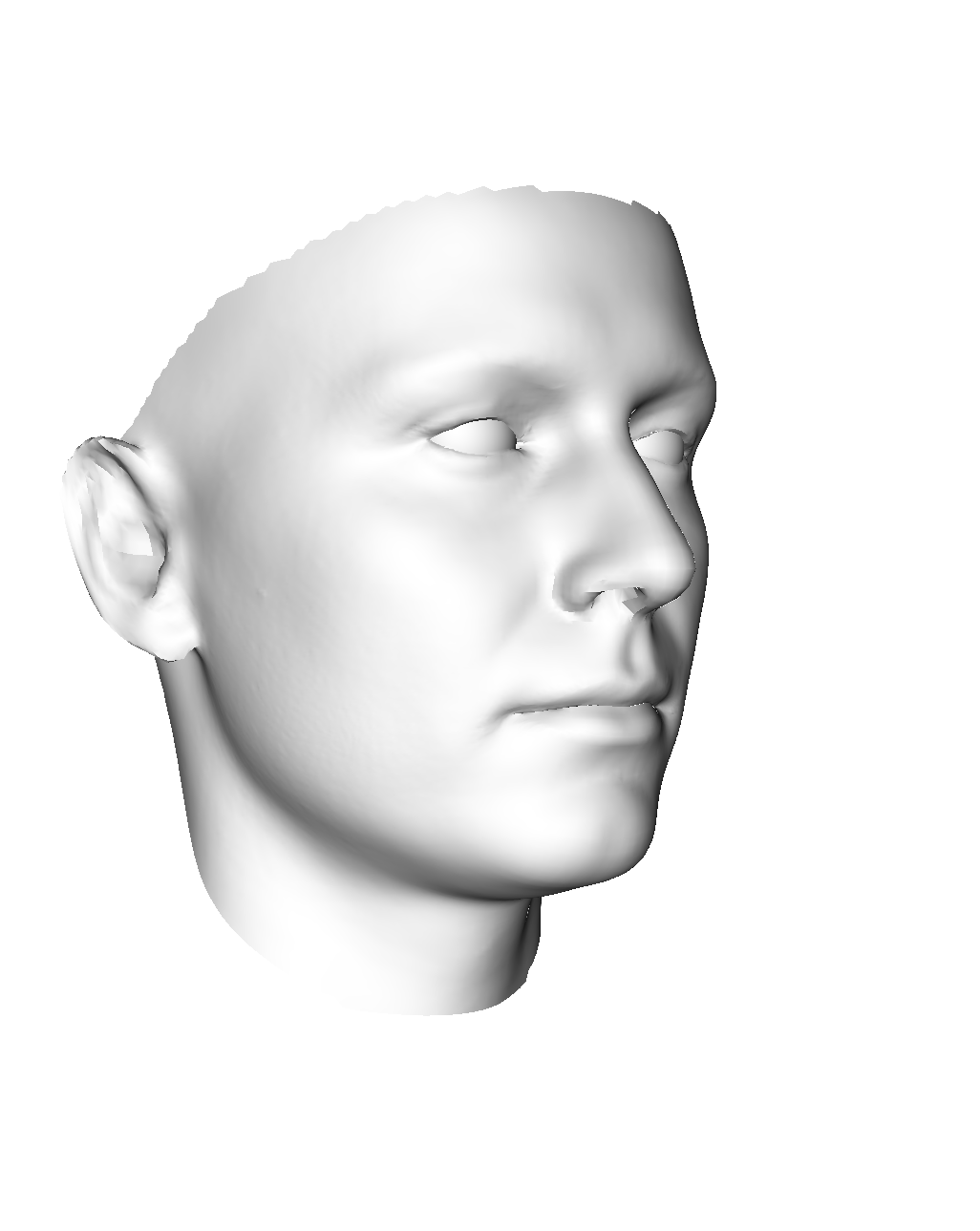}}
\caption{The 3D face surface used to illustrate the effect of different models.}
\label{fig:bfm-mean}
\end{figure}

To visualize the shape variations represented by a model, we define a GPMM on the face surface (see Figure~\ref{fig:bfm-mean}) and show the effect that randomly sampled deformation from this model have on this face surface.\footnote{This face is the average face of the
  publicly available Basel Face Model \cite{paysan20093d}.}  Using the face for visualizing shape variations has the 
advantage that we can judge how anatomically valid a given shape deformation is. 

\subsection{Models of smooth deformations}
\begin{figure}
\subfloat[$s = 100, \sigma = 100$ mm]{
  \includegraphics[trim=2cm 2cm 2cm 2cm,width=0.33\columnwidth]{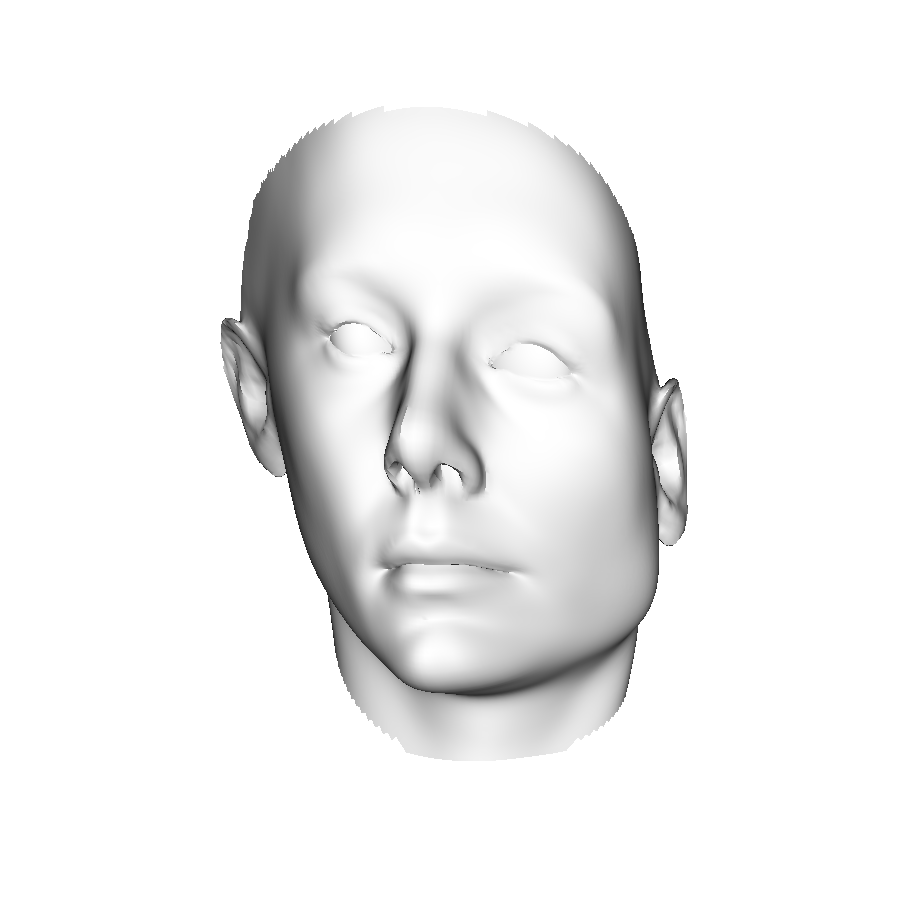}
  \includegraphics[trim=2cm 2cm 2cm 2cm,width=0.33\columnwidth]{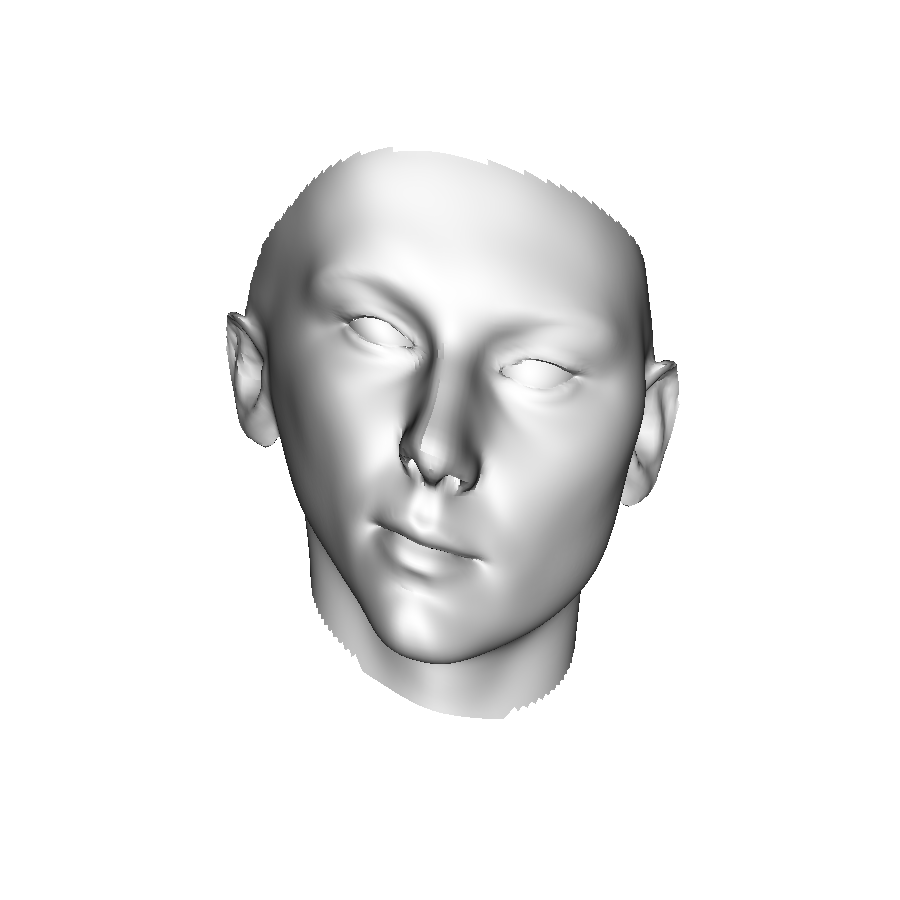}
  \includegraphics[trim=2cm 2cm 2cm 2cm,width=0.33\columnwidth]{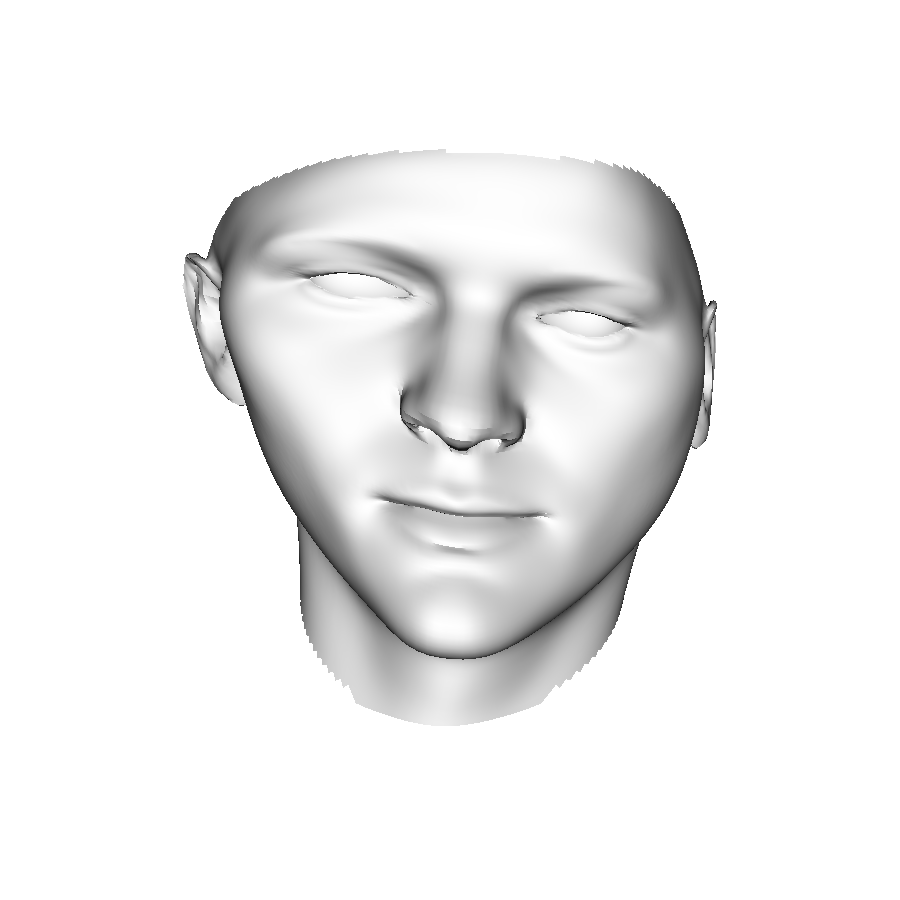} \label{fig:gaussian-samples-100-100}} 
\\
\subfloat[$s = 10, \sigma = 30$ mm]{
  \includegraphics[trim=2cm 2cm 2cm 2cm, width=0.33\columnwidth]{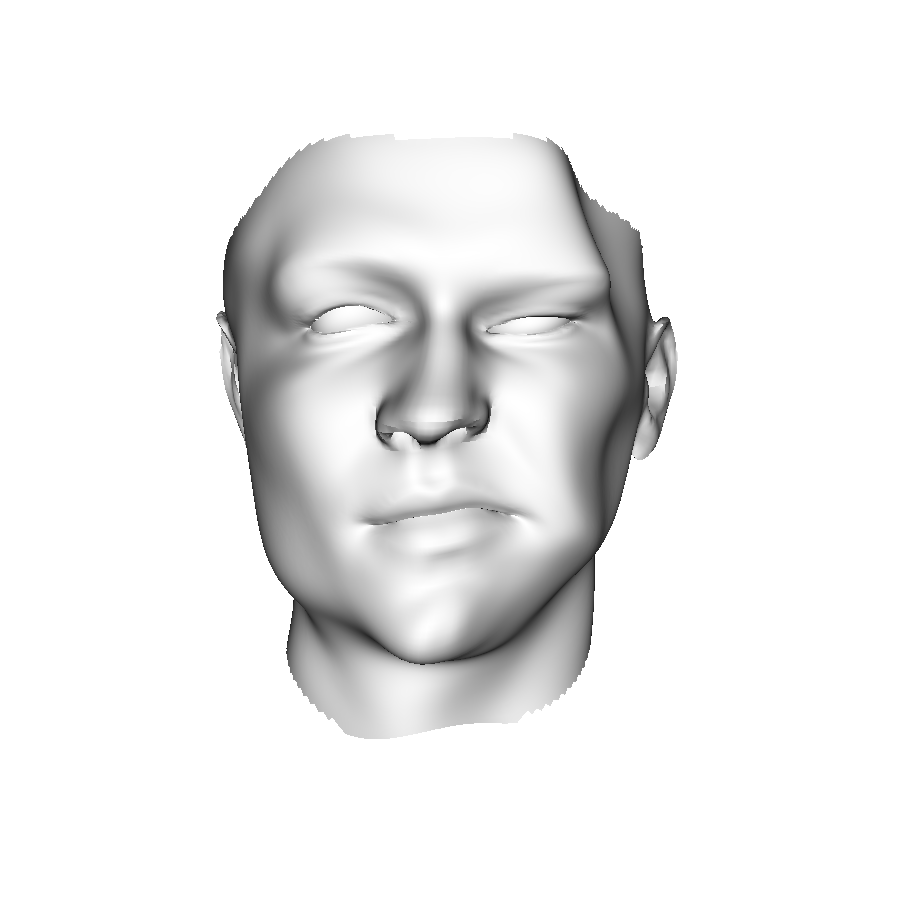}
  \includegraphics[trim=2cm 2cm 2cm 2cm,width=0.33\columnwidth]{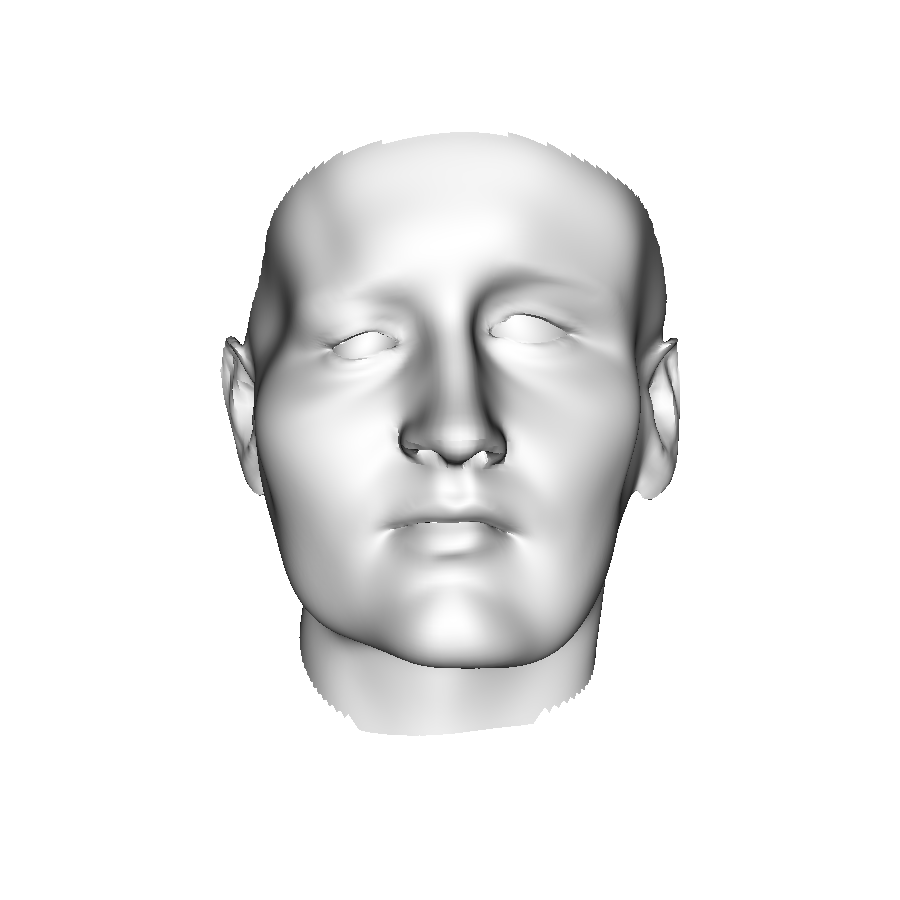}
  \includegraphics[trim=2cm 2cm 2cm 2cm,width=0.33\columnwidth]{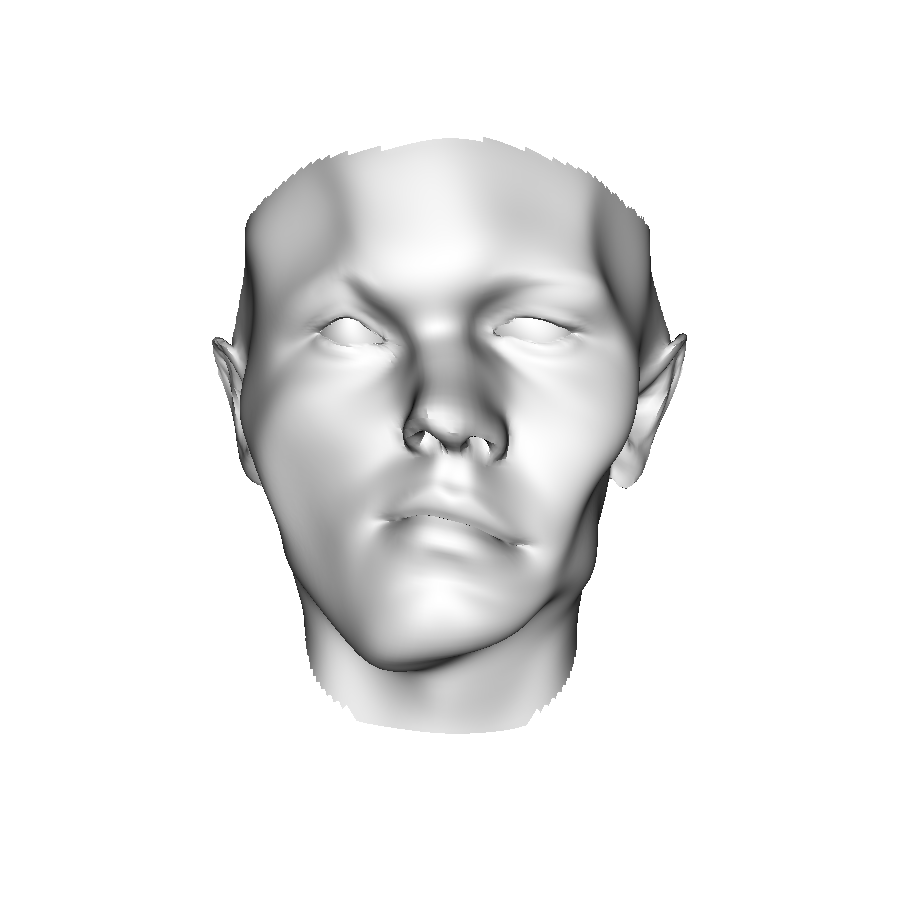}
}
\caption{Samples using a Gaussian kernel with scale factor $s$ and bandwidth  $\sigma$. Choose $\sigma$ large leads to smooth, global deformations
of the face, while choosing it small yields more local deformations.}
\label{fig:gaussian-samples}
\end{figure}

A simple Gaussian process model is a zero mean Gaussian process that
enforces smooth deformations. The assumption of a zero mean is
typically made in registration tasks. It implies that the reference
surface is a representative shape for the class of shapes which are
modelled, that is the shape is close to a (hypothetical) mean shape.  A
particularly simple kernel that enforces smoothness is the Gaussian
kernel defined by
\[
k_g(x,y) = \exp(-\norm{x -y }^2/\sigma^2),
\]
where $\sigma^2$ defines the range over which deformations are correlated. Hence the larger the values
of $\sigma$, the more smoothly varying the resulting deformations fields will be.
In order to use this scalar-valued kernel for registration, we define the matrix valued kernel
\[
k(x,y) = s \cdot I_{3\times3} k_g(x,y),
\]
where the identity matrix $I_{3\times 3}$ signifies that the $x,y,z$
component of the modelled vector field are independent.  The parameter
$s \in \R$ determines the variance (i.e.\ scale) of a deformation
vector.  Figure~\ref{fig:gaussian-samples} shows random samples from
the model for two different values of $\sigma$.

Besides Gaussian kernels, there are many different kernels that are
known to lead to smooth functions.  For registration purposes, spline
models, Elastic-Body Splines \cite{davis1995elastic} B-Splines \cite{kybic2003fast} or Thin Plate Splines
\cite{rohr2001landmark} are maybe the most commonly used. Another useful class of kernels that enforce smoothness is given by the Mat{\'e}rn
class of kernels (see e.g. \cite{rasmussen_gaussian_2006}, Chapter 4), which allows us to 
explicitly specify the degree of differentiabiliy of the model. 

\subsection{Statistical shape models} \label{sec:ssmkernel}
\begin{figure}
  \includegraphics[trim=2cm 2cm 2cm 2cm, width=0.327\columnwidth]{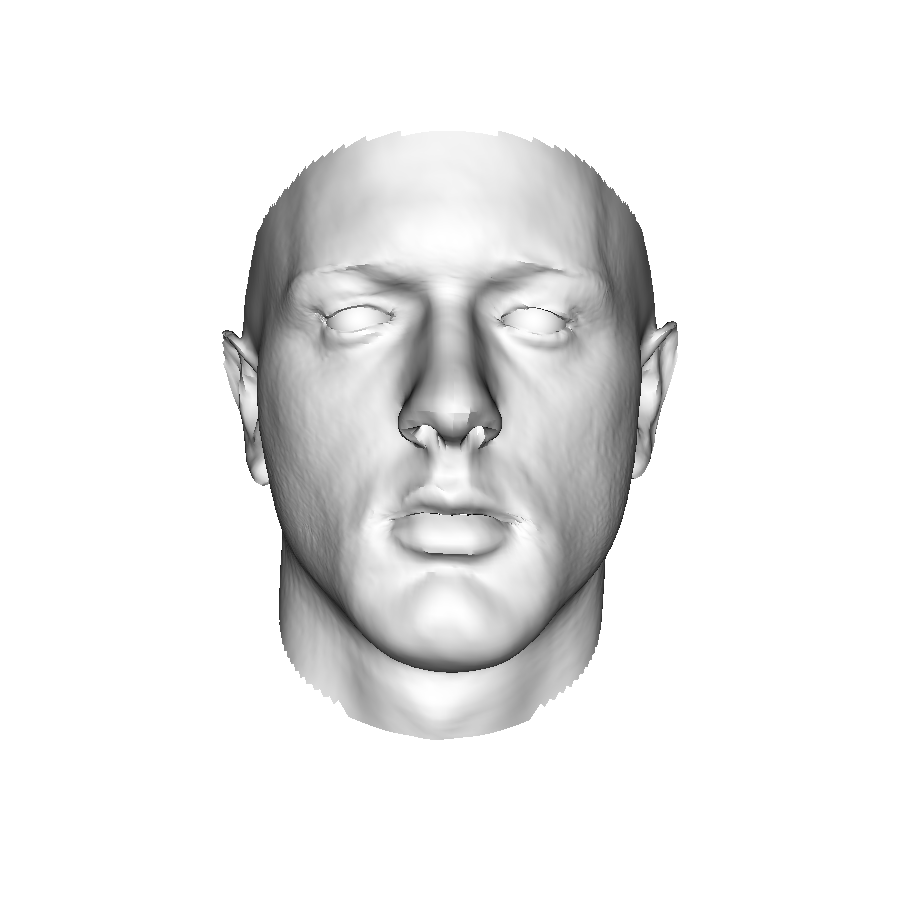}
  \includegraphics[trim=2cm 2cm 2cm 2cm,width=0.327\columnwidth]{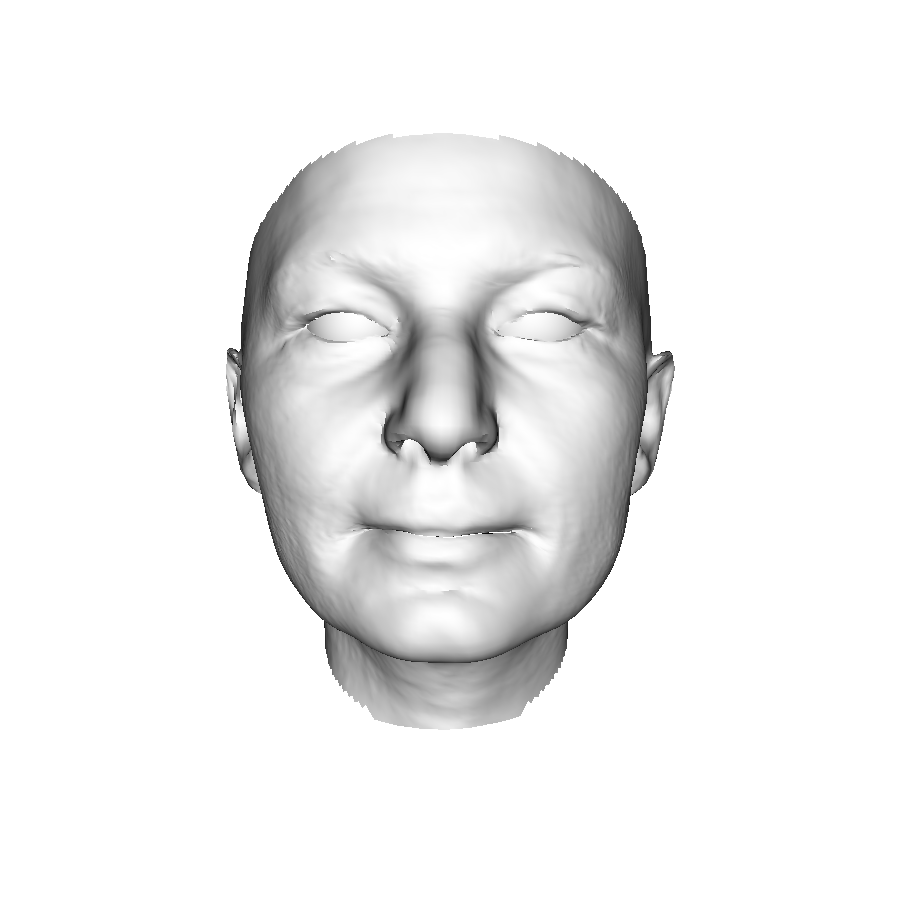}
  \includegraphics[trim=2cm 2cm 2cm 2cm,width=0.327\columnwidth]{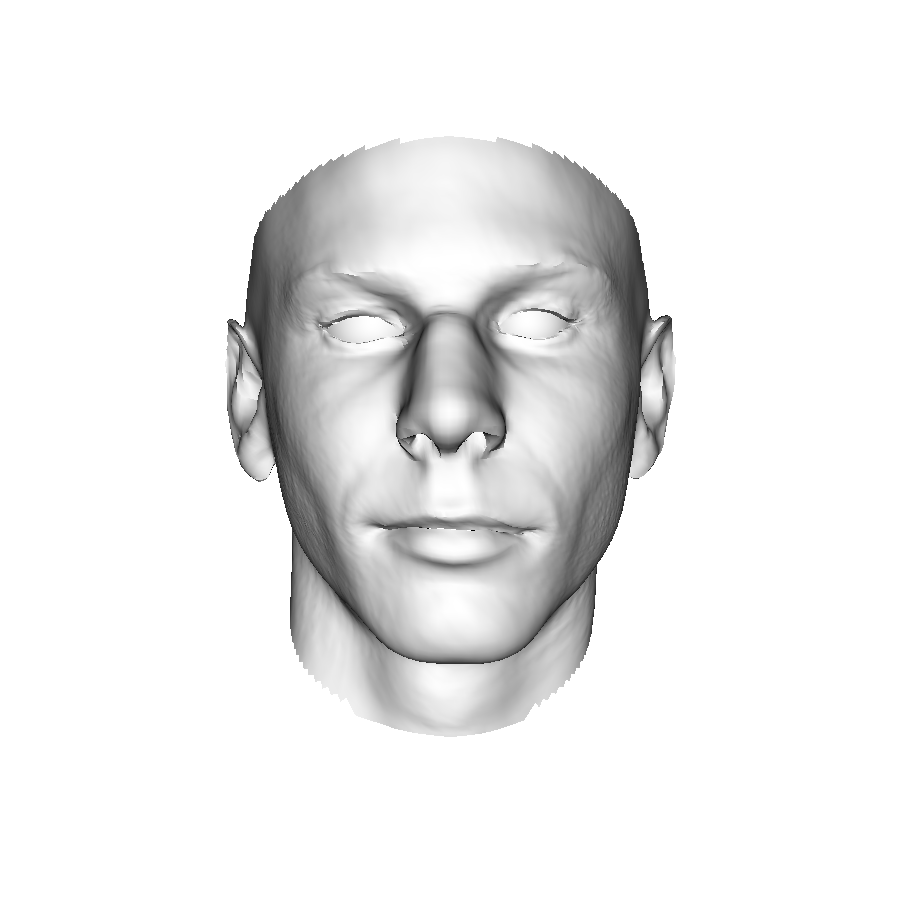}
\caption{Samples using a sample covariance kernel, which is learned from 200 training faces. All the random samples look like valid faces.}
\label{fig:ssm-samples}
\end{figure}

The smoothness kernels are generic, in the sense that they do not
incorporate prior knowledge about a shape.  This is reflected in
Figure~\ref{fig:gaussian-samples}, where the sampled faces do not
look like valid anatomical face shapes.  An ideal prior for the
registration of faces would only allow valid face shapes. This is the motivation behind SSMs \cite{cootes_active_1995,blanz_morphable_1999}. The
characteristic deformations are learned from a set of typical examples surfaces
$\Gamma_1,\ldots, \Gamma_n$.  More precisely, by
establishing correspondence, between a reference $\Gamma_R$ and each
of the training surfaces, we obtain a set of deformation fields
$\{u_1, \ldots, u_n\}, u_i : \Omega \to \R^d$, where $u_i(x)$ denotes
a deformation field that maps a point on the reference $x \in
\Gamma_R$ to its corresponding point $u_i(x)$ on the $i-$th training
surface.  A Gaussian process $\GP(\mu_{SM}, k_{SM})$ that
models these characteristic deformations is obtained by estimating the
empirical mean
\[
\mu_{SM}(x) = \frac{1}{n} \sum_{i=1}^n u_i(x)
\]
and covariance function
\begin{equation} \label{eq:smkernel}
 k_{SM}(x, y) = \frac{1}{n-1} \sum_{i=1}^n (u_i(x) - \mu_{SM}(x)) (u_i(y) - \mu_{SM}(y))^T.
\end{equation}
We refer to the kernel $k_{SM}$ as the \emph{sample covariance kernel} or empirical kernel. 
Samples from such a model are depicted in Figure~\ref{fig:ssm-samples}, 
where the variation was estimated from 200 face surfaces from the  Basel Face Model \cite{paysan20093d}. In contrast to the smoothness priors, all the sampled face surfaces represent anatomically plausible faces. The model that we obtain using this sample covariance
kernel is a continuous analogon to a PCA based shape model.

\subsection{Combining kernels}

The kernels that we have discussed so far already provide a large variety
of different choices for modeling prior assumptions. But the real
power of these models comes to bear if the ``simple`` kernels are combined
to define new kernels, making use of a relative rich algebra that kernels admit. 
In the following, we present basic combinations of kernels to give the reader a taste of what can be achieved. 
For a more thorough discussion of how positive definite kernels can be combined, we 
refer the reader to Shawe-Taylor et al. \cite{shawe2004kernel} (Chapter 3, Proposition 3.22). 

\subsubsection{Multiscale models}
\begin{figure}
  \includegraphics[trim=2cm 2cm 2cm 2cm,width=0.327\columnwidth]{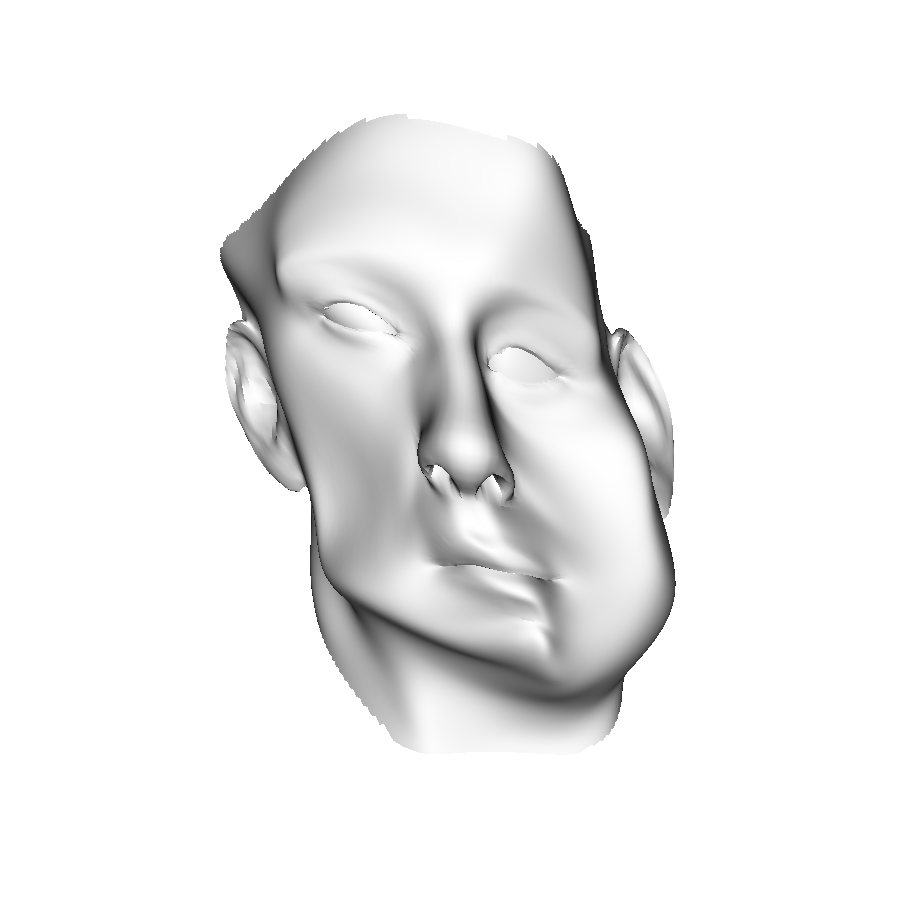}
  \includegraphics[trim=2cm 2cm 2cm 2cm,width=0.327\columnwidth]{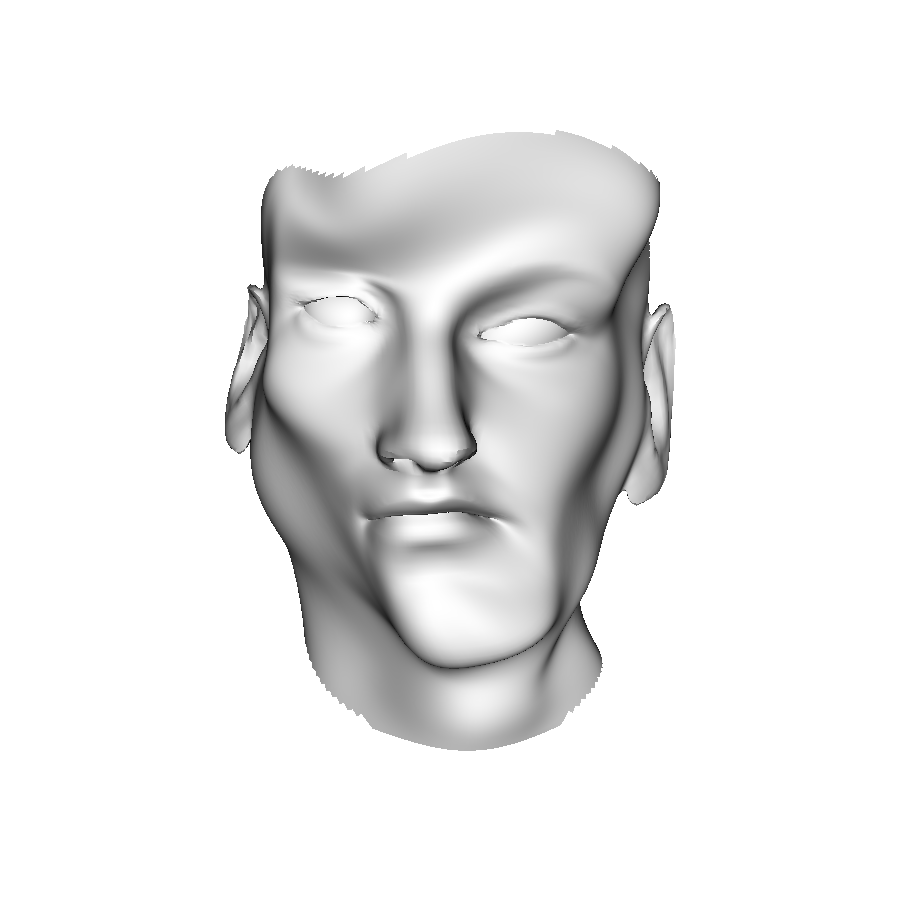}
  \includegraphics[trim=2cm 2cm 2cm 2cm,width=0.327\columnwidth]{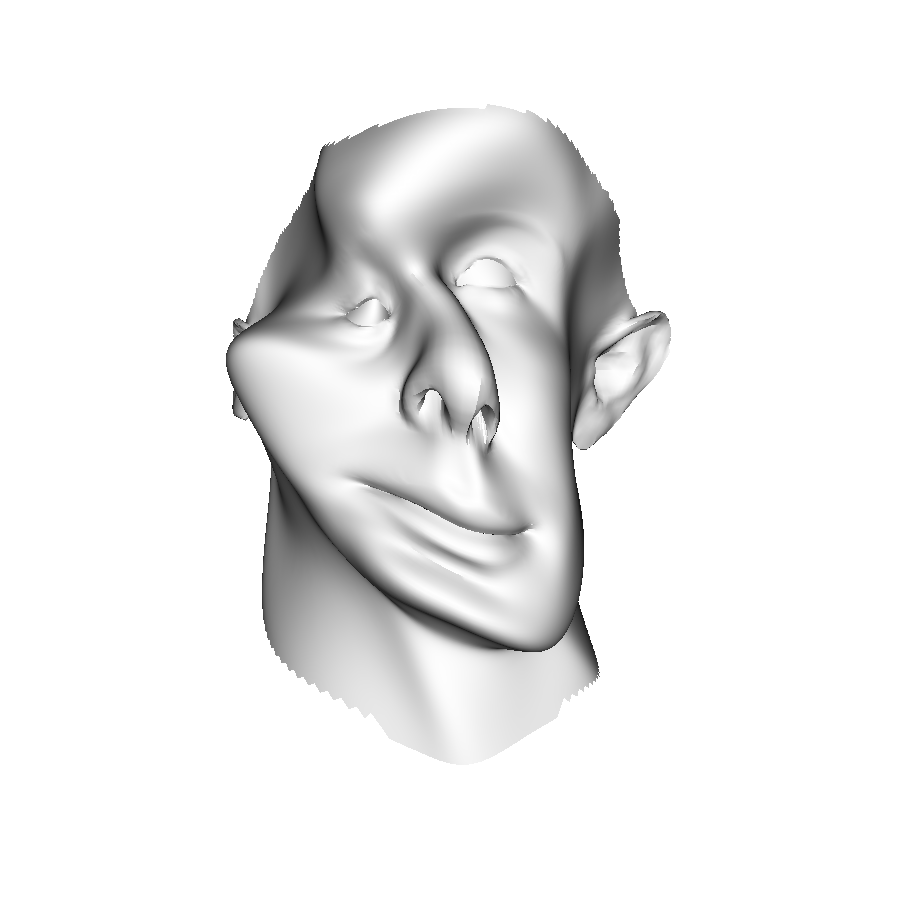} 
  \caption{Samples using a kernel defined on multiple scales. The random sample show large deformations which change the overall face shape, as well as local, detailed shape variations.}
  \label{fig:ms-samples}
\end{figure}

If $k_1, \ldots, k_n: \Omega \times \Omega \to \R^{d \times d}$ are positive definite kernels, then the linear combination
\[
k_S(x, x') = \sum_{i=1}^l \alpha_i k_i(x, x'), \; \alpha_i \in \R
\]
is positive definite as well. This provides a simple means of modeling deformations on multiple scale levels by summing
kernels that model smooth deformations with kernels for more local, detailed deformation. 
A particularly simple implementation of such a strategy is to sum up Gaussian kernels, with decreasing scale and bandwidth:
\[
k_{MS}(x,x') =  \sum_{i=1}^l \frac{s}{i} I_{3 \times 3} \exp(-\frac{\norm{x-x'}^2}{(\sigma/i)^2})
\]
where $s$ determines the base scale and $\sigma$ the smoothness and $l$ the number of levels. 
As shown in Figure~\ref{fig:ms-samples}, this simple approach already leads to a multiscale structure that models both large scale deformations as well as local details. This idea could be extended to obtain wavelet-like multi-resolution strategies, by choosing the kernels to be refinable (which is, for example true for the B-spline kernel). A more detailed discussion of such kernels is given in \cite{opfer_multiscale_2006,xu2009refinement}.

\subsubsection{Reducing the bias in statistical shape models} \label{sec:bias}
Another use case that makes use of the possibility to add two kernels is to explicitly model the bias of a statistical shape model. 
Due to the limited availability of training examples, statistical shape models are often not able to represent the full shape space accurately and thus introduce a bias
towards the training shapes into model-based methods \cite{le_pdm-enlor:_2013}. One possibility to avoid this problem is to provide an explicit bias model. 
Denote by $k_{SM} : \Omega \times \Omega \to \R^{d \times d}$ the sample covariance kernel and let $k_g : \Omega \times \Omega \to \R$ be a  Gaussian kernel
with bandwidth parameter $\sigma$. A simple model to reduce the bias would be to use a Gaussian kernel with a large bandwidth, i.e.\ we define
\[
k_{b}(x,x') = k_{SM}(x,x') + s I_{3\times 3}k_g(x,x'), 
\]
where the parameter $s$ defines the scale of the average error. This parameter could, for example, be estimated using crossvalidation. 
This simple model assumes that the error is spatially correlated, i.e.\ if a model cannot explain the structure at a certain point, its neighboring points
are likely to also show the same error. An example of how this strategy can reduce the bias in statistical shape models is given in our previous publication \cite{luthi2013unified}. 

\subsubsection{Localizing models} \label{sec:ssm-locality}
\begin{figure}
  \includegraphics[trim=2cm 2cm 2cm 2cm,width=0.327\columnwidth]{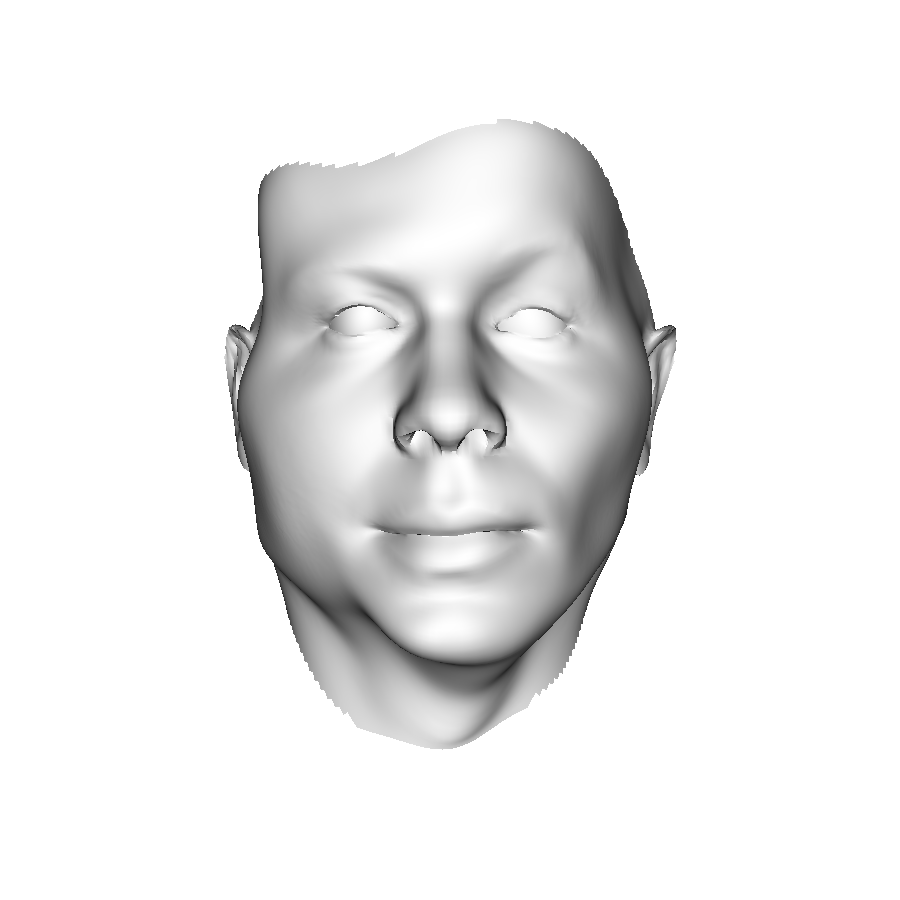}
  \includegraphics[trim=2cm 2cm 2cm 2cm,width=0.327\columnwidth]{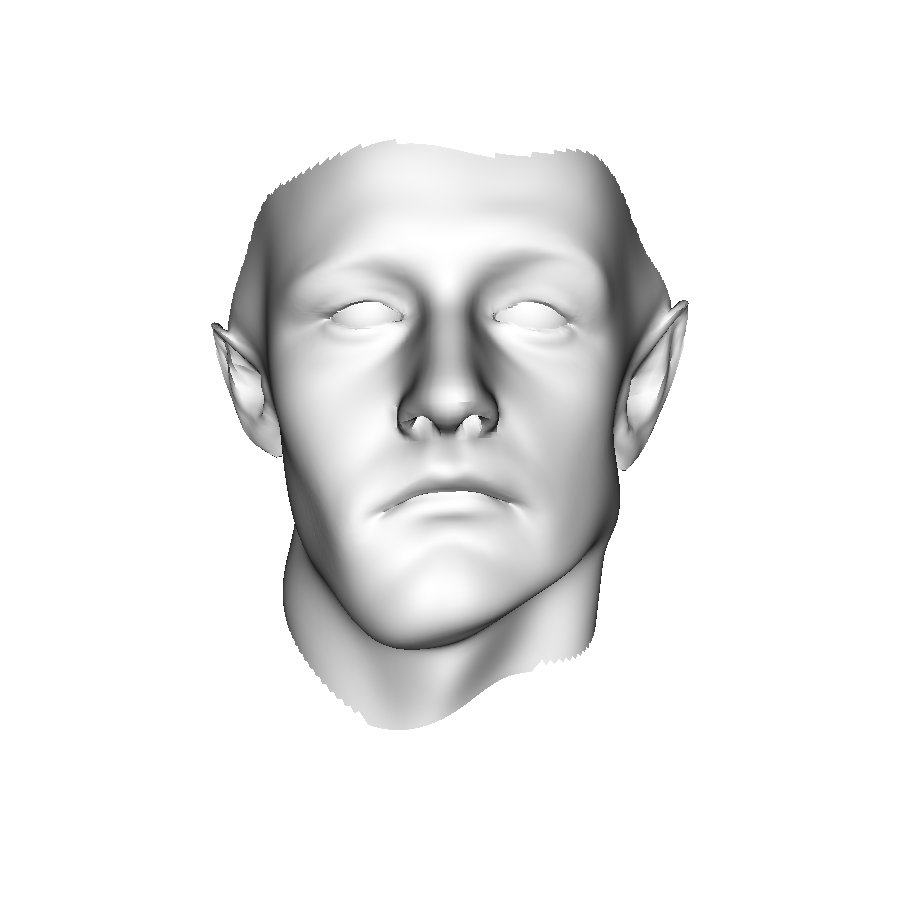}
  \includegraphics[trim=2cm 2cm 2cm 2cm,width=0.327\columnwidth]{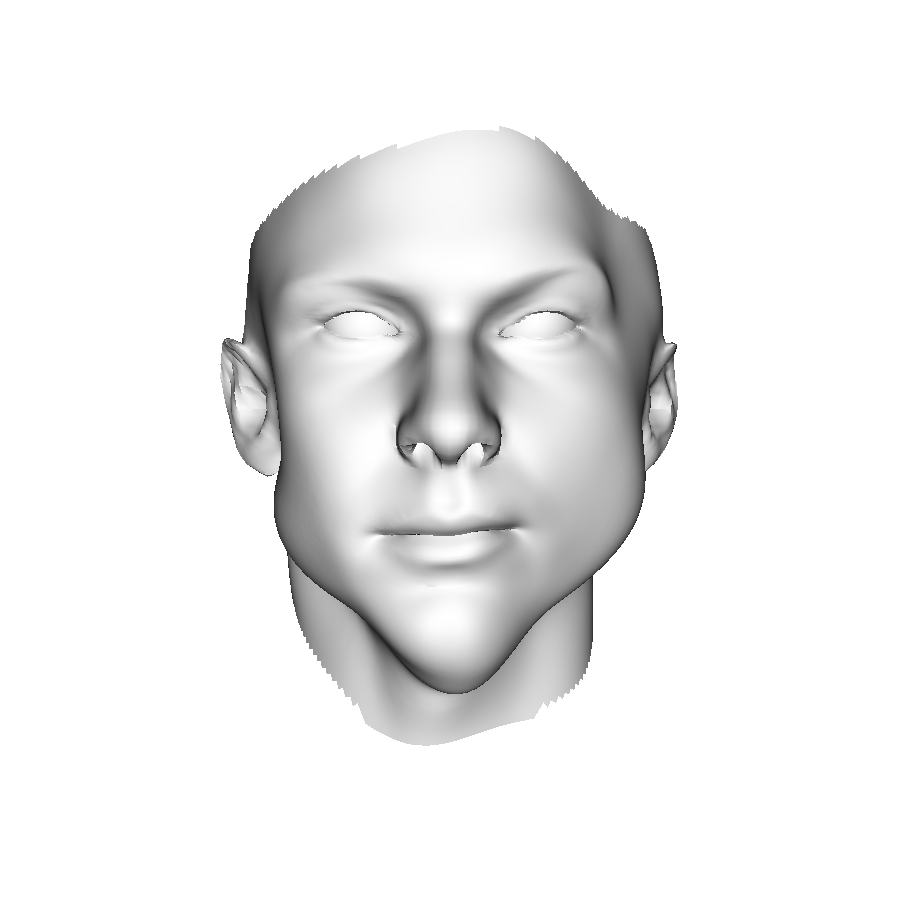} 
  \caption{Random samples form a localized statistical shape model. Whereas locally the variations look anatomically valid, there are no global correlations anymore, which 
    makes the model more flexible.}
  \label{fig:localized-samples}
\end{figure}

Another possibility to obtain more flexible models is to make models more local. 
Recalling that the kernel function $k(x,x')$ models the correlation 
between point $x$ and $x'$, we see that setting the correlation $k(x,x')$ to $0$ for $x\neq x'$ decouples the points and hence increases the flexibility of a model.
Such an effect can be achieved by a multiplication of two kernel functions, which again results in a positive definite kernel. A simple example of a local model is obtained
by multiplying a kernel with a Gaussian kernel. For example, by defining 
\[
k_l(x,x') = k_{SM}(x,x') \odot I_{3 \times 3} k_g(x,x')
\]
(where $\odot$ defines element wise multiplication), we obtain a localized version of a statistical shape model. 
Samples from such a model are shown in Figure~\ref{fig:localized-samples}. We observe that the samples locally look like valid faces, but globally, the kernel still allows for more flexible variations, 
which could not be described by the model, and which may not constitute an anatomically valid face. 

\subsubsection{Spatially varying models}
Taking these ideas one step further, we can also combine the two
approaches and sum up several localized models to obtain a
non-stationary kernel. The use of such kernels in the context of
registration has recently been proposed by Schmah et al.\
\cite{schmah2013left}, who showed how it can be used in a method for
spatially-varying registration.

\begin{figure}\center{
  \includegraphics[trim=2cm 2cm 2cm 2cm,width=0.327\columnwidth]{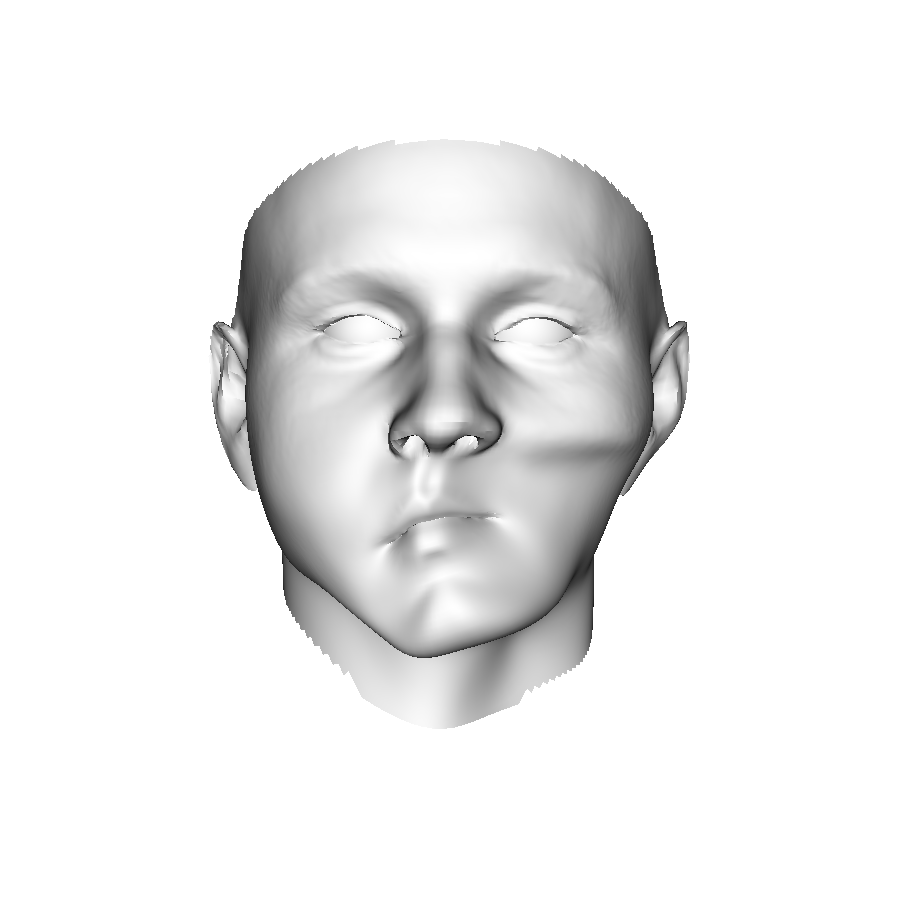}
  \includegraphics[trim=2cm 2cm 2cm 2cm,width=0.327\columnwidth]{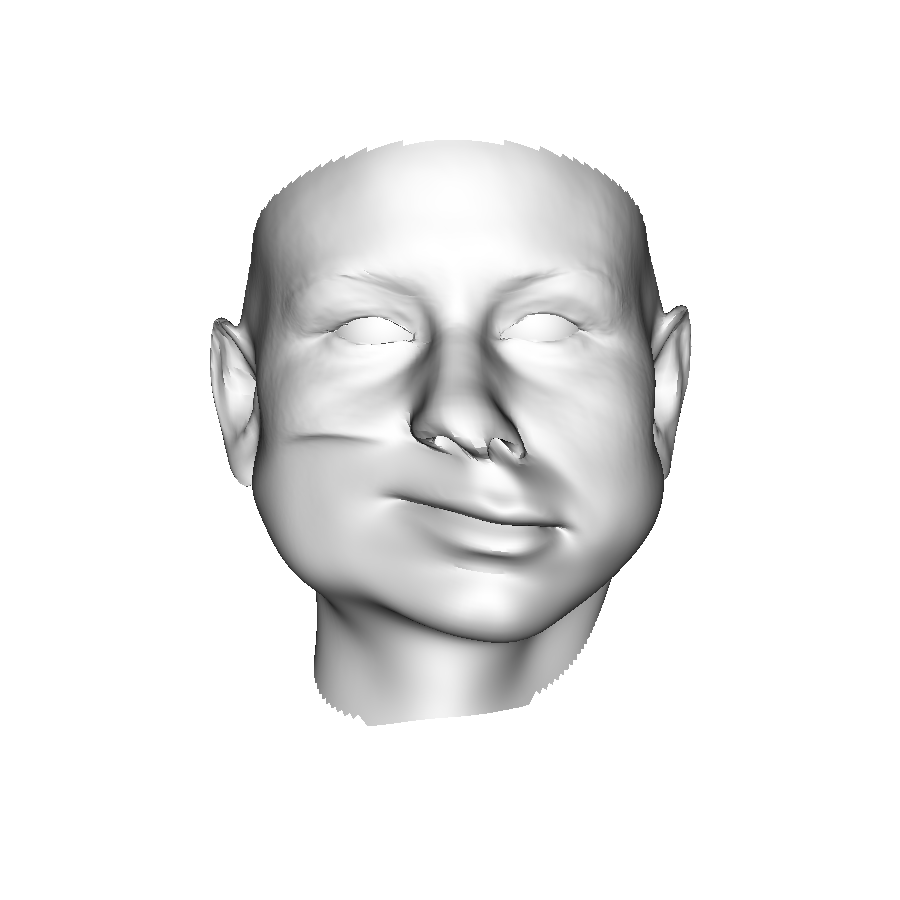}
  \includegraphics[trim=2cm 2cm 2cm 2cm,width=0.327\columnwidth]{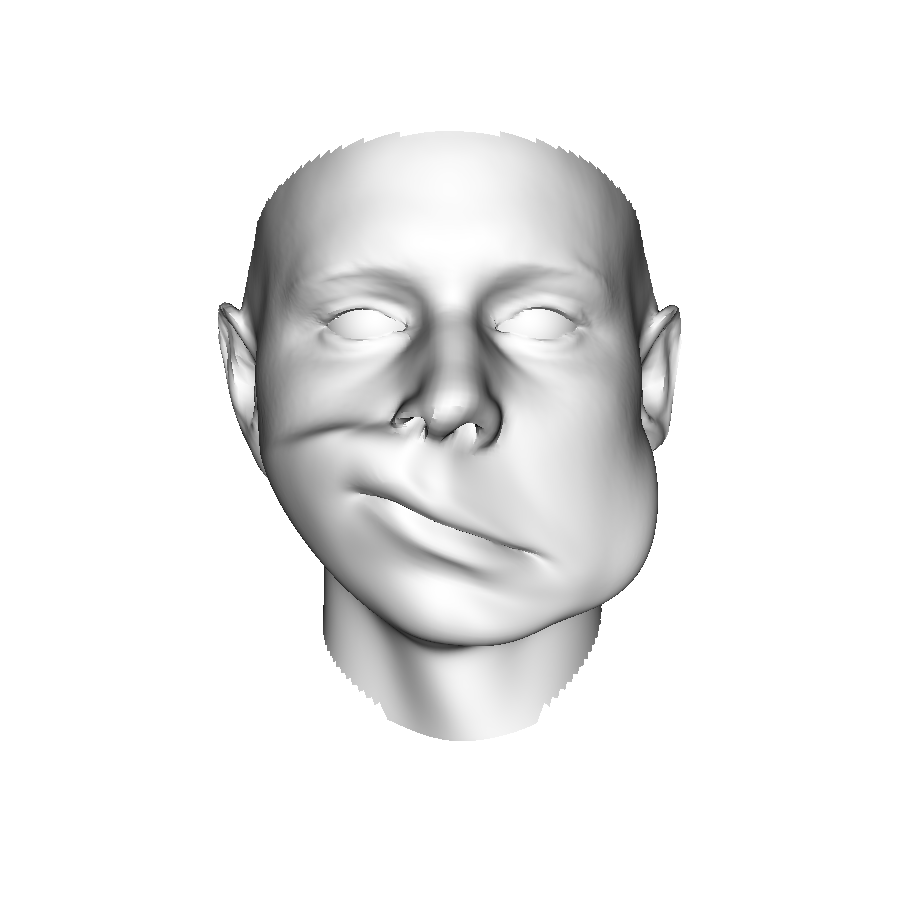}}
  \caption{Random samples form a spatially-varying model. The variation in the upper part of the face is model using a sample covariance kernel while the lower part uses a Gaussian kernel.}
  \label{fig:spatially-varying}
\end{figure}

Let $\Omega =
(\Omega_1, \ldots, \Omega_n)$ be a partition of the domain $\Omega$
into several regions $\Omega_i$ and associate to each region a kernel $k^i$ with the desired characteristics.  
We define a weight functions $w_i: \Omega \to [0,1]$, such that $\sum_i w_i(x) = 1$. The
weight function $w_i$ is chosen such that $w_i(x) = 1$ for the region
$\Omega_i$.  For any real valued function $f$ the kernel
$k(x,x') = f(x)f(x')$ is positive definite, and we can define a ``localization`` kernel 
\[
k_l^i(x,x') = w_i(x)w_i(x'). 
\]
The final spatially-varying model $\GP(0,k_{sv})$ using this kernel is defined by
\[
k_{sv}(x,x') = \sum_{i} k_l^i(x,x') k^i(x,x').
\]
Figure~\ref{fig:spatially-varying} shows samples from a model where the upper
part of the face is modelled using a statistical model, while the lower
part (below the nose) undergoes arbitrary smooth deformations.

\subsection{Posterior models}\label{sec:fixed-deformations}
\begin{figure}
\subfloat[$s = 10, \sigma = 30mm$]{
  \includegraphics[trim=2cm 2cm 2cm 2cm, width=0.33\columnwidth]{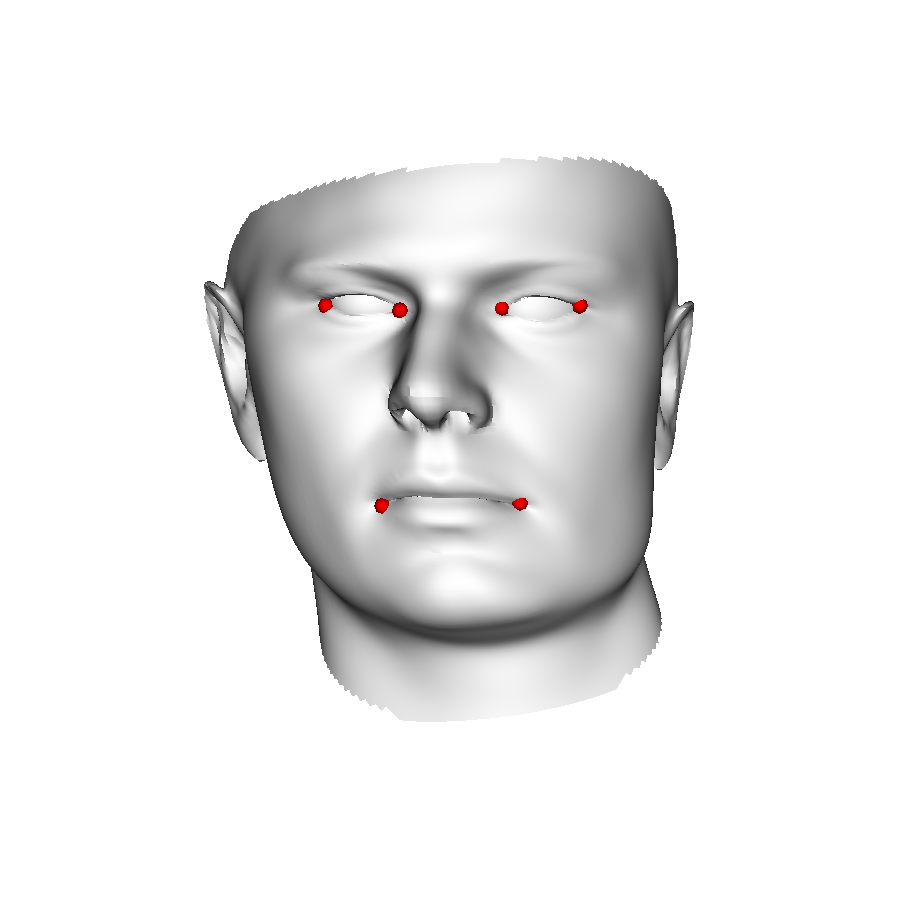}
  \includegraphics[trim=2cm 2cm 2cm 2cm,width=0.33\columnwidth]{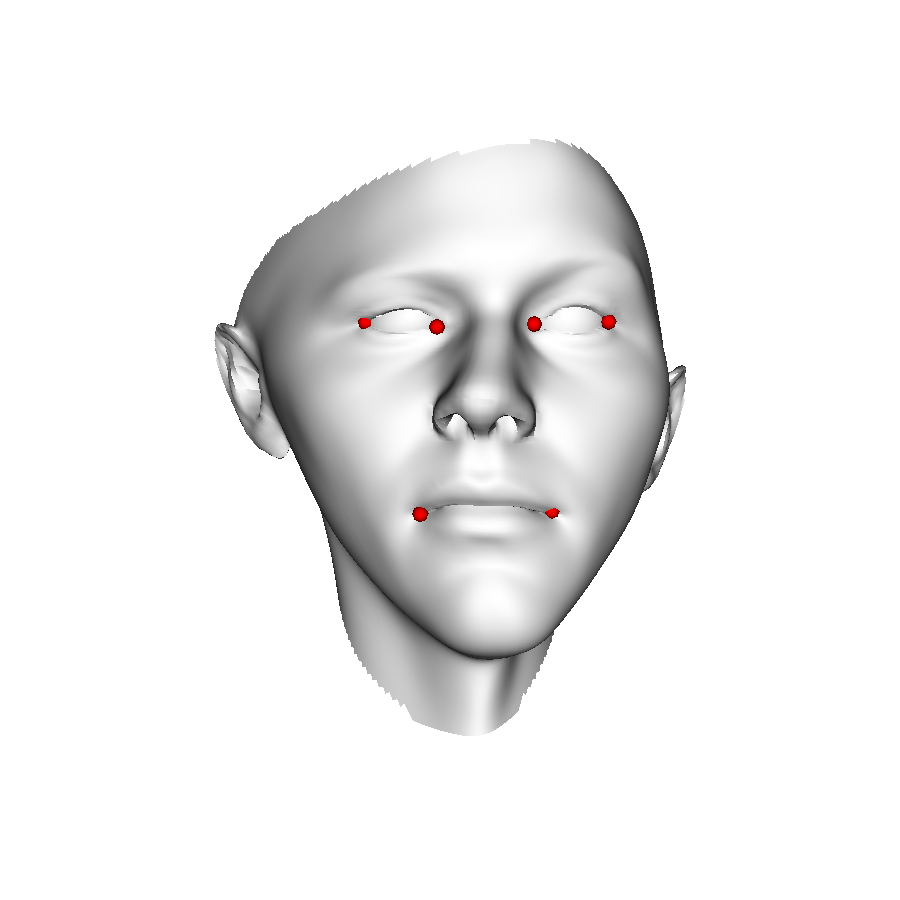}
  \includegraphics[trim=2cm 2cm 2cm 2cm,width=0.33\columnwidth]{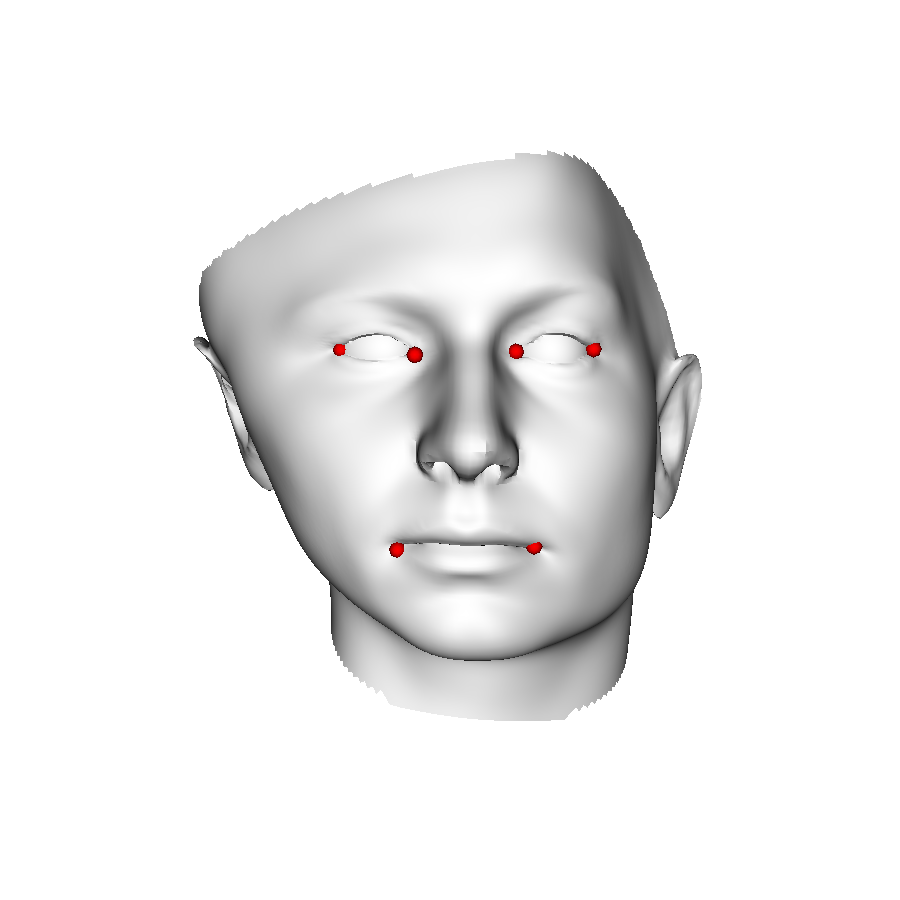}
}
\caption{Random samples from a posterior model, which has been obtained by taking the Gaussian process model shown in Figure~\protect\ref{fig:gaussian-samples-100-100}, and applying Gaussian process regression to keep the points shown in red fixed.}
\label{fig:gp-posterior}
\end{figure}

Modeling by combining different kernels amounts to specifying our prior
assumptions by modeling how points correlate. In many applications we
have not only information about the correlations, but know for certain
points exactly how they should be mapped.  Assume for instance, that a
user has clicked a number of landmark points on a reference shape
${L_R}=\{l_R^1, \ldots, l_R^n)\}$ together with the matching points on
a target surface ${L_T}=\{l_T^1, \ldots, l_T^n)\}$.  These landmarks
provide us with known deformation at the matching points, i.e.\
 \[
 {L} = \{(l_R^1, l_T^1 - l_R^1), \ldots, (l_R^n, l_T^n - l_R^n)\} =: \{(l_R^1, \hat{u}^1), \ldots, (l_R^n, \hat{u}^n)\}. 
 \]
Assuming further that the deformations $\hat{u}^i$ are subject to Gaussian noise $\epsilon \sim I_{3 \times 3} \sigma$, 
we can use Gaussian process regression to compute from a given Gaussian process $GP(\mu, k)$ a new Gaussian process, known as the posterior process or posterior model, $\GP(\mu_p, k_p)$ (cf.\ \cite{rasmussen_gaussian_2006}, Chapter 2). Its mean $\mu_p$ and covariance $k_p$ are known in closed form and given by
 \begin{align}  \label{eq:gp-vec-post}
   \mu_p(x) &= \mu(x) + K_X(x)^T (K_{XX} + \sigma^2\mathcal{I})^{-1}\hat{U}\\
   k_p(x,x') &= k(x, x') - 
   K_X(x)^T(K_{XX}+\sigma^2\mathcal{I})^{-1}K_X(x').
 \end{align}
 Here, we defined $K_X(x) = (k(x, x_i))_{i=1}^n \in \R^{3n \times 3}$,
 $K_{XX} = \left(k(x_i, x_j)\right)_{i,j=1}^n \in \R^{3n \times 3n}$
 and $\hat{U} = (\hat{u}^1 - \mu(x), \ldots, \hat{u}^n - \mu(x))^T \in \R^{3n}$. Note
 that $K_X$ and $K_{XX}$ consist of sub-matrices of size $3 \times 3$.
 The posterior is again a Gaussian process, and hence is itself a (data-dependent) GPMM, which 
 we can use anywhere we would use a standard GPMM. 
 Figure~\ref{fig:gp-posterior} shows random
 samples from such  prior, where the points shown in red are fixed by setting 
 $(\hat{u}^i=(0, 0, 0)^T), i=1, \ldots, n)$. 




\section{Registration using Gaussian Process Morphable Models } \label{sec:registration}
In this section we show how we can use GPMMs as
prior models for surface and image registration. The idea is that we
define a model for the variability of a given object $O_R \subset \R^d$ and fit
this model to a target image or surface $O_T \subset \R^d$. Our main
assumption is that we can identify for each
point $x_R \in O_R$, a corresponding
point $x_T \in O_T$ of the target object $O_T$.  
More formally, it is
assumed that there exists a deformation $u : \Omega \to \R^d$, such
that
\[
O_T =  \{x + u(x) | x \in O_R \}.
\] 
The goal of the registration problem is to
recover the deformation field $u$, that relate the two objects.

To this end, we formulate the problem as a MAP estimate:
\begin{equation}\label{eq:gp-functional}
\argmax_u p(u) p(O_T | O_R, u),
\end{equation}
where $p(u) \sim \GP(\mu, k)$ is a Gaussian process prior over the
admissible deformation fields and $p(O_T|O_R, u) =
\frac{1}{Z}\exp(\eta^{-1}\mathcal{D}[O_R, O_T, u])$, $\mathcal{D}$ is
a metric  that measures the similarity of the objects $O_T, O_R$. Here $\eta \in \R$ is a weighting parameter and $Z$ a normalization constant.  In order to find the MAP solution, we reformulate 
the registration problem as an energy minimization problem. 
It is well known (see e.g. \cite{wahba_spline_1990} for
details) that the  solution to the MAP problem
\eqref{eq:gp-functional} is attainable by minimizing in the RKHS
$\mathcal{F}_k$ defined by $k$ :
\begin{equation}\label{eq:rkhs-formulation1}
  \argmin_{u \in \mathcal{F}_k}   \mathcal{D}[O_R, O_T, u] + \eta \norm{u}_{k}^2,
\end{equation}
where $\norm{\cdot}_{k}$ denotes the RKHS norm. Using the low-rank approximation
(Equation~\eqref{eq:low-rank}) we can restate the problem in the
parametric form
\begin{equation}\label{eq:rkhs-formulation2}
  \argmin_{\alpha_1, \ldots, \alpha_r}    \mathcal{D}[O_R, O_T, \mu + \sum_{i=1}^r \alpha_i \sqrt{\lambda}_i\phi_i]  + \eta \sum_{i=1}^r \alpha_i^2
\end{equation}
The final registration \eqref{eq:rkhs-formulation2} is highly
appealing. All the assumptions are represented by the eigenfunctions
$\phi_i, i =1 \ldots, r$, which in turn are determined by the Gaussian
process model. Thus, we have split the registration problem into three
separate problems:
\begin{enumerate}
\item \textbf{Modelling:} Specify a model for the deformations by
  defining a Gaussian process model for the deformations $u \sim
  \GP(\mu, k)$ \label{enum:gp}
\item \textbf{Approximation:} Approximate the model by replacing in parametric form $\tilde{u} = \mu + \sum_{i=1}^r \alpha_i \sqrt{\lambda_i} \phi_i \sim \GP(\mu, \tilde{k})$, in 
terms of its eigendecomposition.  \label{enum:ev}
\item \textbf{Fitting:} Fit the model to the data by minimizing the
  optimization problem
  \eqref{eq:rkhs-formulation2}. \label{enum:fitting}
\end{enumerate}
The separation of the modelling and the fitting step is most
important, as it allows us to treat the conceptual work of modelling
our prior assumptions independently from the search of a good algorithm to actually
perform the registration. Indeed, in this paper we will use
the same, simple fitting approach for both surface and image fitting, which we detail in the following.

\subsection{Model fitting for  surface and image registration} \label{sec:surface-and-image-registration}
To turn the conceptual problem \eqref{eq:rkhs-formulation2} into a practical one,  we need to specify the representations of the reference and target object $O_R, O_T$ and define a distance measure $\mathcal{D}$ between them. 

We start with the case where the object $O_R, O_T$
correspond to surfaces $\Gamma_R, \Gamma_T \subset \R^3$. A simple
measure $\mathcal{D}$ is the mean squared distance from the reference
to the closest target point, i.e.\
\[
D[\Gamma_R, \Gamma_T, u] = \int_{\Gamma_R}(CP_{\Gamma_T}(x + u(x)))^2
\, dx,
\] 
where $\text{CP}_{\Gamma_T}$ is the distance function defined by
\[
\text{CP}_{\Gamma_T}(x) = \norm{x - \argmin_{x' \in \Gamma_T} \norm{x-x'}}.
\] 
Hence, the registration problem \eqref{eq:rkhs-formulation2} for surface registration (with $u = \sum_{i=1}^r\alpha_i\sqrt{\lambda}_i\phi_i$) becomes
\begin{equation}
\alpha^* = \argmin_{\alpha} \int_{\Gamma_r} \text{CP}_{\Gamma_T}(x + \sum_{i=1}^r \alpha_i \sqrt{\lambda}_i \phi_i(x)) \, dx +  \eta \sum_{i=1}^r \alpha_i^2.
\end{equation}
Note that for surface registration, we are only interested in
deformations defined on $\Gamma_R$. It is therefore sufficient to
compute the Nystr\"om approximtion using only points sampled from the
reference $\Gamma_R$.

The second important case is image registration. Let $I_R, I_T :
\Omega \to \R$ be two images defined on the image domain $\Omega$. 
In
this case, we usually choose $\mathcal{D}$ such that it integrates some
function of the image intensities over the two images (see e.g.\
\cite{sotiras2013deformable} for an overview of different similarity
measures). In the simplest case, we can use the squared distance of
the intensities. The image registration problem becomes:
\begin{equation} \label{eq:image-registration}
\alpha^* = \argmin_{\alpha} \int_{\Omega} [I_R(x) - I_T(x + \sum_{i=1}^r \alpha_i \sqrt{\lambda}_i\phi_i(x))]^2 \, dx +  \eta \sum_{i=1}^r \alpha_i^2.
\end{equation}

Note that to be well defined, the Gaussian process needs to be defined
on the full image domain $\Omega$. Therefore, we sample points from
the full image domain $\Omega$ to compute the Nystr\"om
approximation. As evaluating the eigenfunction $\phi$ can be
computationally expensive (Cf. Equation~\ref{eq:nystrom-phi}) we
propose to use a stochastic gradient descent method to perform the
actual optimization \cite{klein_evaluation_2007}.

Besides these straight-forward algorithms for surface and image
registration, we can also directly make use of any algorithm that is
designed to work with PCA-based shape models. This is possible because our
model \eqref{eq:low-rank} is of the same form as a PCA-model, with the
only difference that we have continuously defined basis function. We
can recover a standard model by discretizing the basis functions for a
given set of points. A popular example of such an algorithm is Active Shape Model fitting \cite{cootes_active_1995}. We will see an application of it in Section~\ref{sec:results}.

\subsection{Hybrid registration}

Independently of whether we do surface or image registration, we can easily obtain a hybrid registration scheme by including
landmarks directly into the model. Recall from Section~\ref{sec:fixed-deformations} that using Gaussian process regression, we 
can obtain for any GPMM a corresponding posterior model that is  guaranteed to match a set of deformations between landmark points. 
When we use such a model for registration, this leads to a hybrid registration schemes that combines landmark and shape or intensity
 information. Compared to previously proposed method for hybrid registration
 \cite{johnson_consistent_2002,fischer_combination_2003,biesdorf_hybrid_2009},
 this model incorporates the landmark constraint directly as into the
 prior, and thus does not require any change in the actual algorithm.


\section{Results} \label{sec:results}

In this section we illustrate the use of GPMMs for the application of
model-based segmentation of forearm bones from CT images.  We start by
showing how to build an application specific prior model of the ulna
bone using analytically defined kernels. We use this model to perform
surface registration in order to establish correspondence between a
set of ulna-surfaces, and thus to be able to build a statistical shape
model. In a second experiment we use this model to perform Active Shape Model fitting
and show how increasing the model's flexibility using a GPMM improves the results. Finally, we also show an application of
GPMMs for image registration.

\subsection{Experimental setup}
\begin{figure}
 \subfloat[CT Image Slice]{
  \includegraphics[width=0.5\columnwidth]{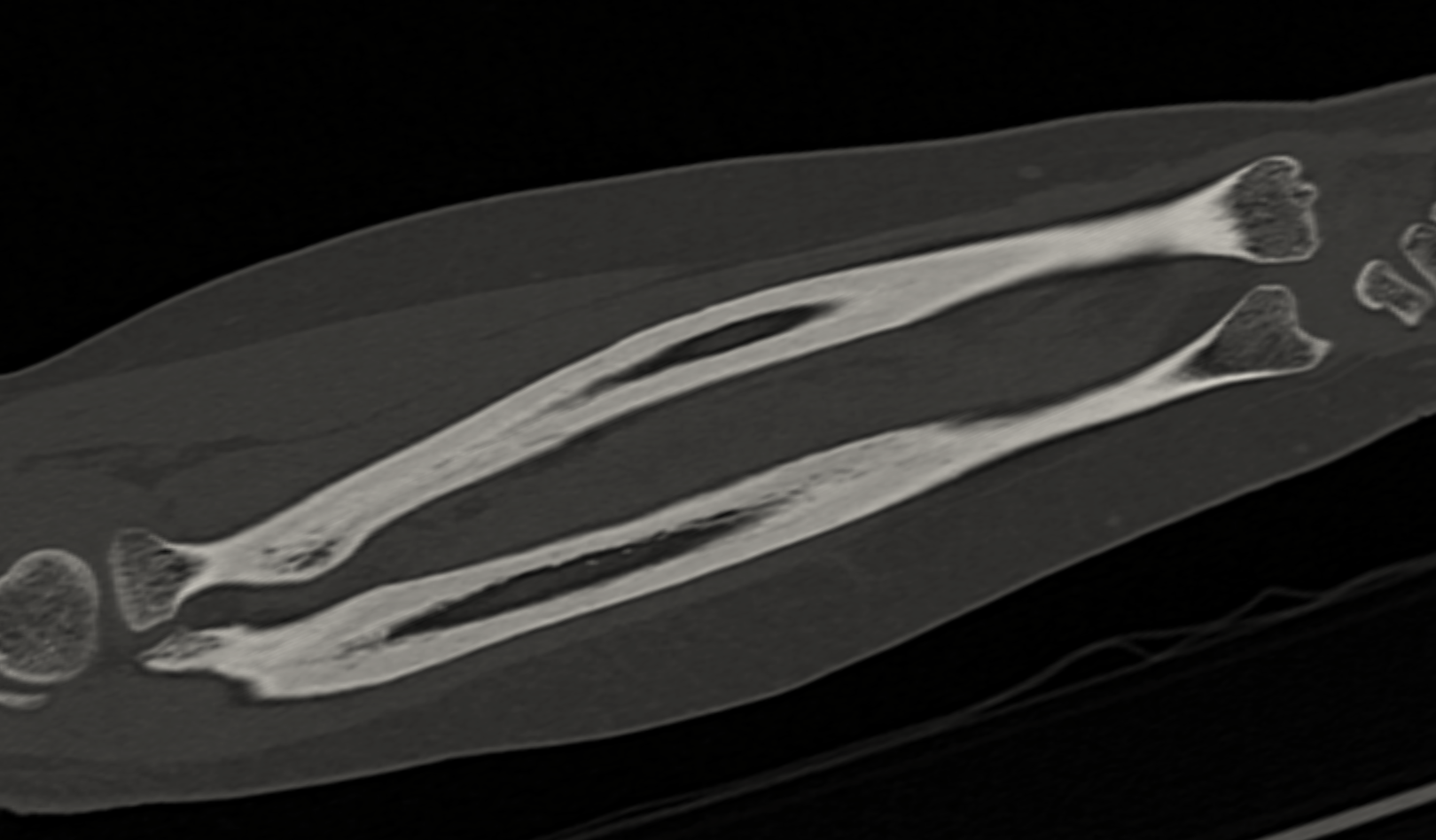}
}
 \subfloat[Surface]{
  \includegraphics[height=0.5\columnwidth, angle=90]{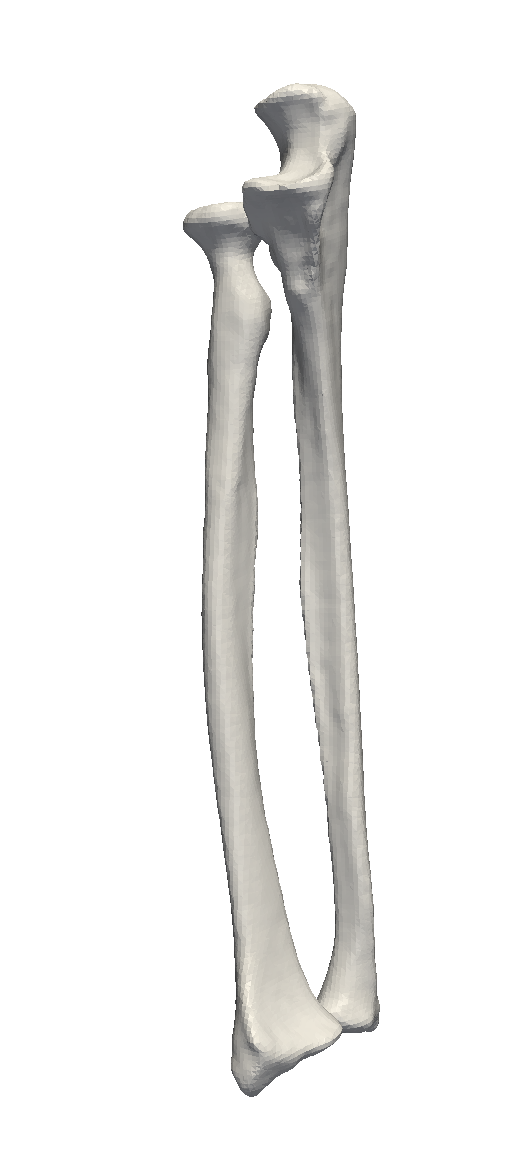}
}
\caption{A slice through a CT image of the forearm (left) and the extracted bone surface from a ground-truth segmentation.}
\label{fig:datasets}
\end{figure}

Our data consists of 36 segmented images of the right forearm bones (ulna and radius). For 27 of these bones we have the original CT image.
Using the 36 given segmentations, we extracted the ulna surface using the marching cubes algorithm \cite{lorensen1987marching}.
We chose an arbitrary data-set as a reference and defined on each ulna surface 4 landmark points (two on the proximal, two on the distal part of the ulna), which we used to rigidly aligned the original images, the segmentation as well as the extracted surfaces to the reference image \cite{umeyama1991least}. Figure~\ref{fig:datasets} shows a typical CT image, and the forearm bones.

We integrated GPMMs in the open source shape modelling software libraries scalismo \cite{scalismo} and statismo \cite{luthi_statismo-framework_2012}. We used scalismo for model-building, surface registration and Active shape model fitting. For performing the image registration experiments, we used statismo, together with the Elastix toolbox for non-rigid image registration \cite{klein2010elastix}.

\subsection{Building prior models}
\begin{figure}
  \subfloat[Gauss]{
  \includegraphics[trim=2cm 2cm 2cm 2cm,width=0.13\columnwidth]{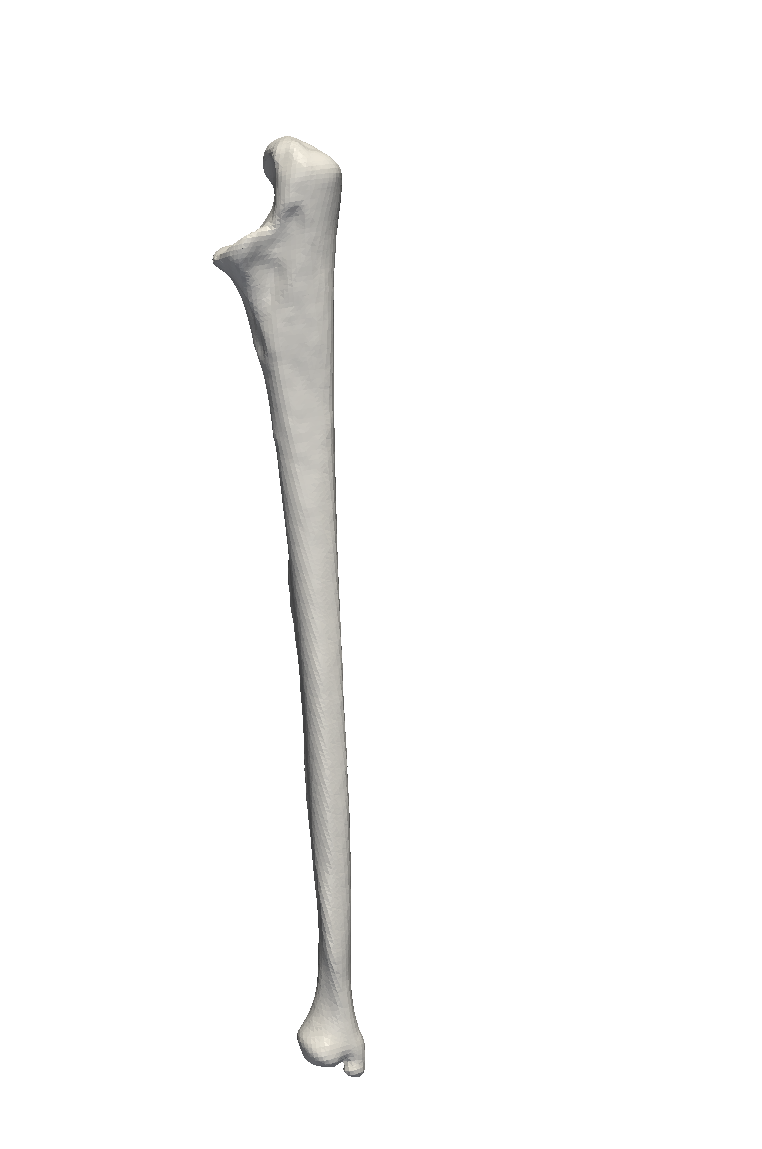}
  \includegraphics[trim=2cm 2cm 2cm 2cm,width=0.13\columnwidth]{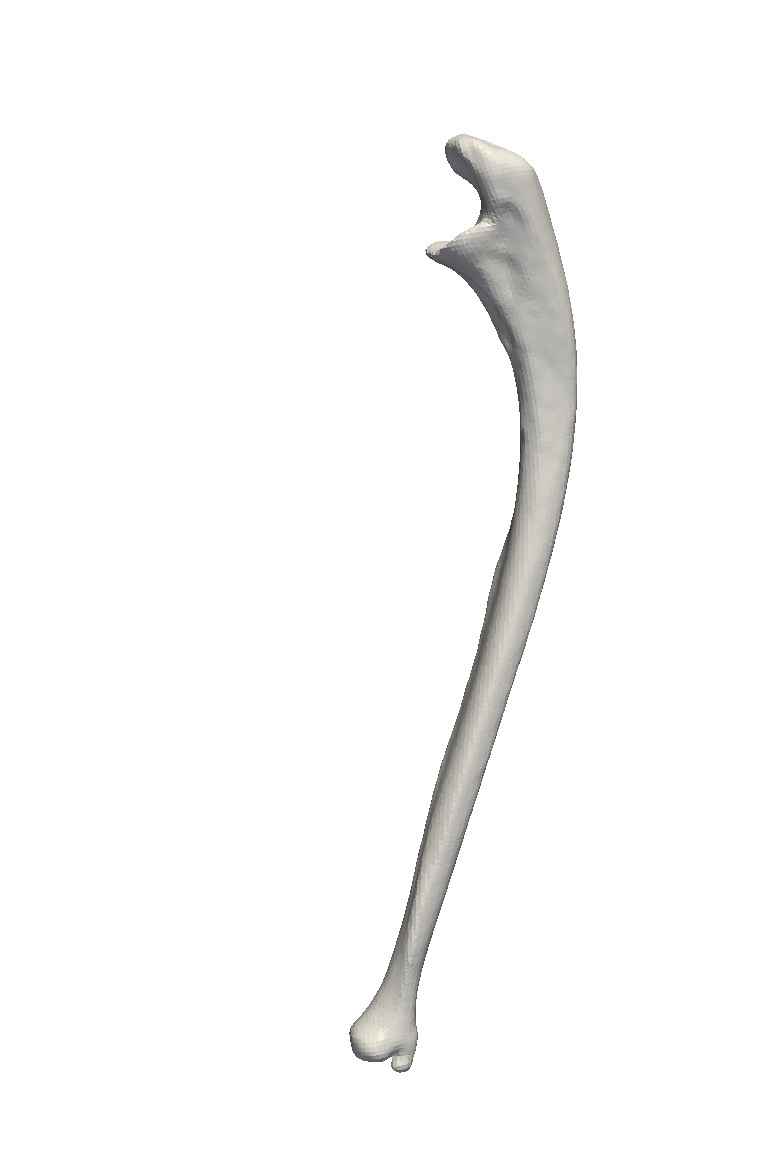}
}
  \hspace{0.8cm}
  \subfloat[Multiscale (isotropic)]{
  \includegraphics[trim=2cm 2cm 2cm 2cm,width=0.13\columnwidth]{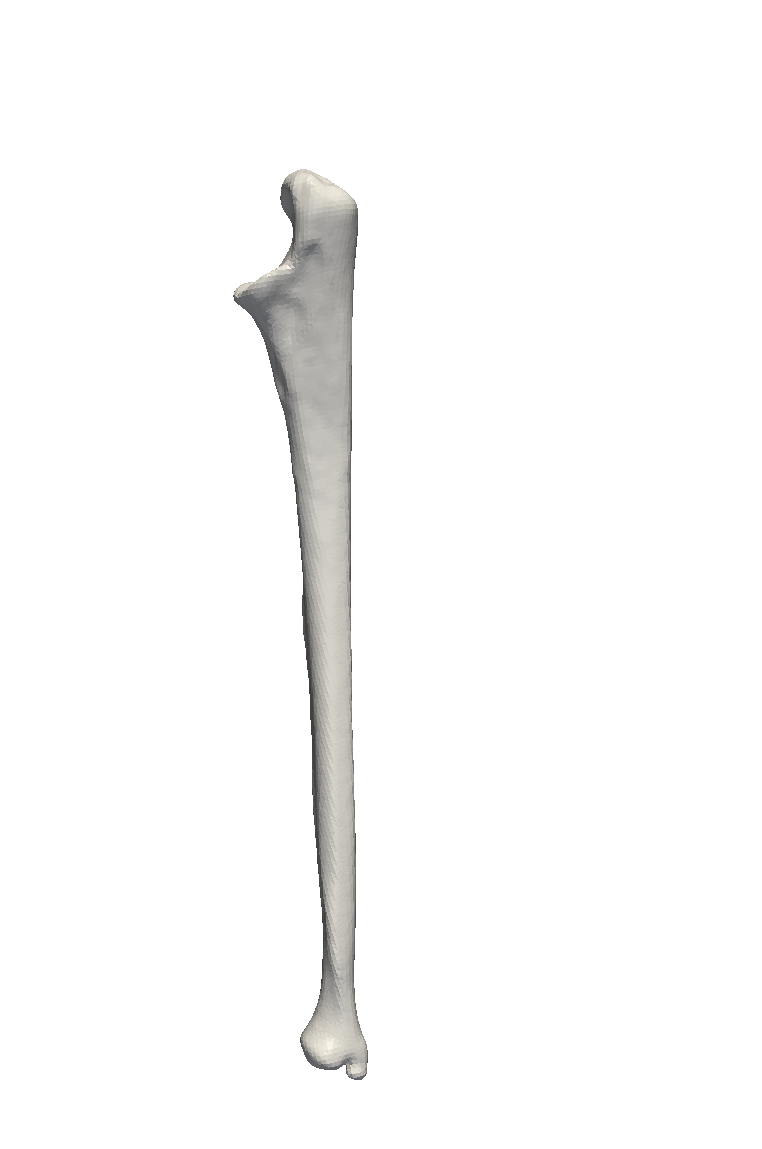}
  \includegraphics[trim=2cm 2cm 2cm 2cm, width=0.13\columnwidth]{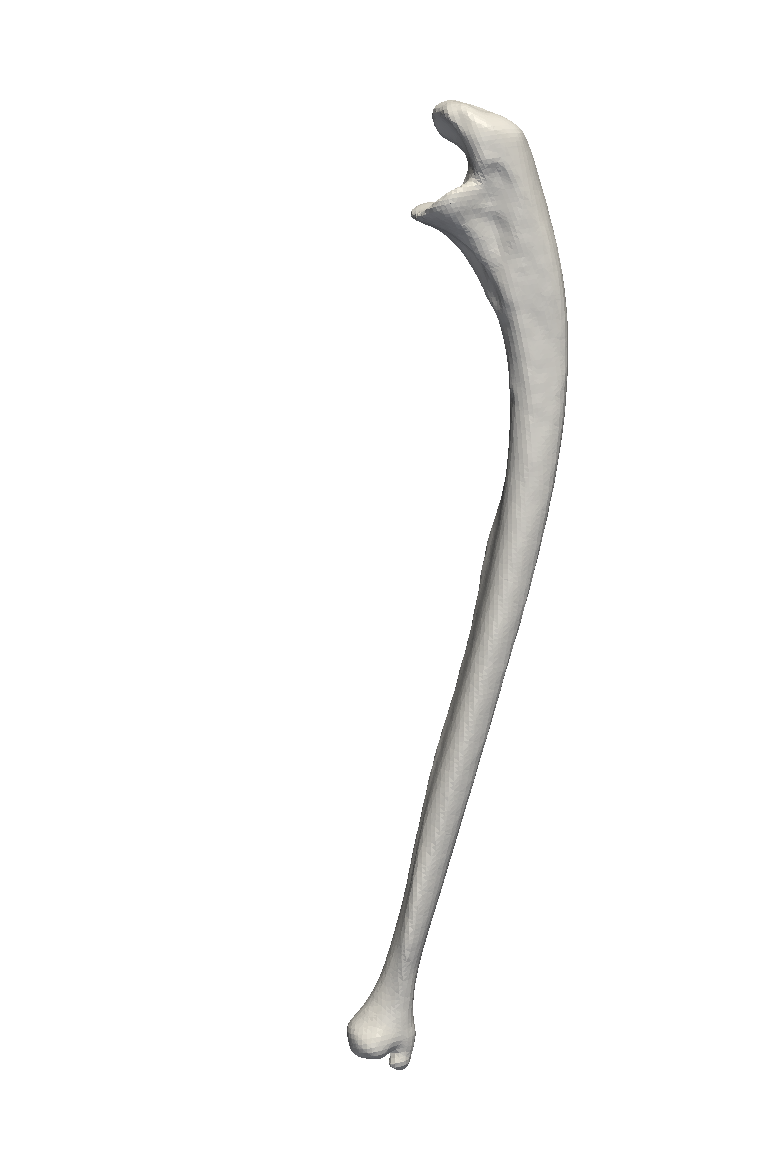}
}
    \hspace{0.8cm} \subfloat[Multiscale (anisotropic)]{
  \includegraphics[trim=2cm 2cm 2cm 2cm, width=0.13\columnwidth]{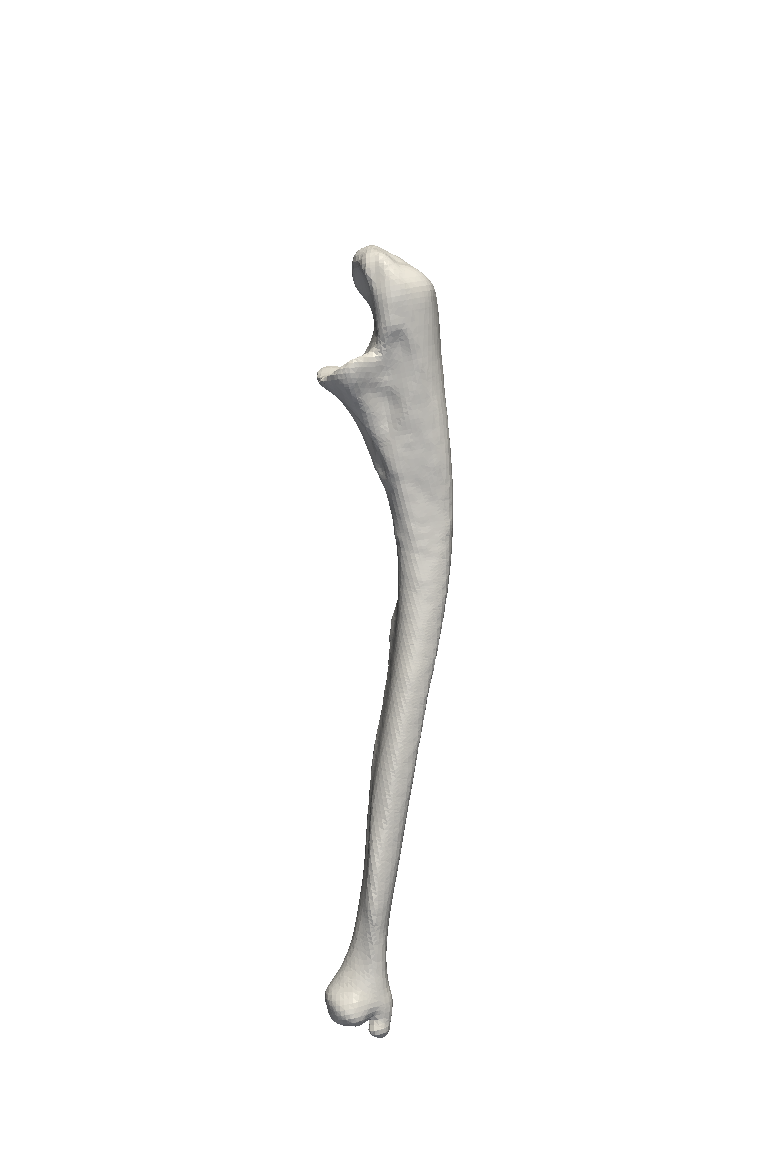}
  \includegraphics[trim=2cm 2cm 2cm 2cm,width=0.13\columnwidth]{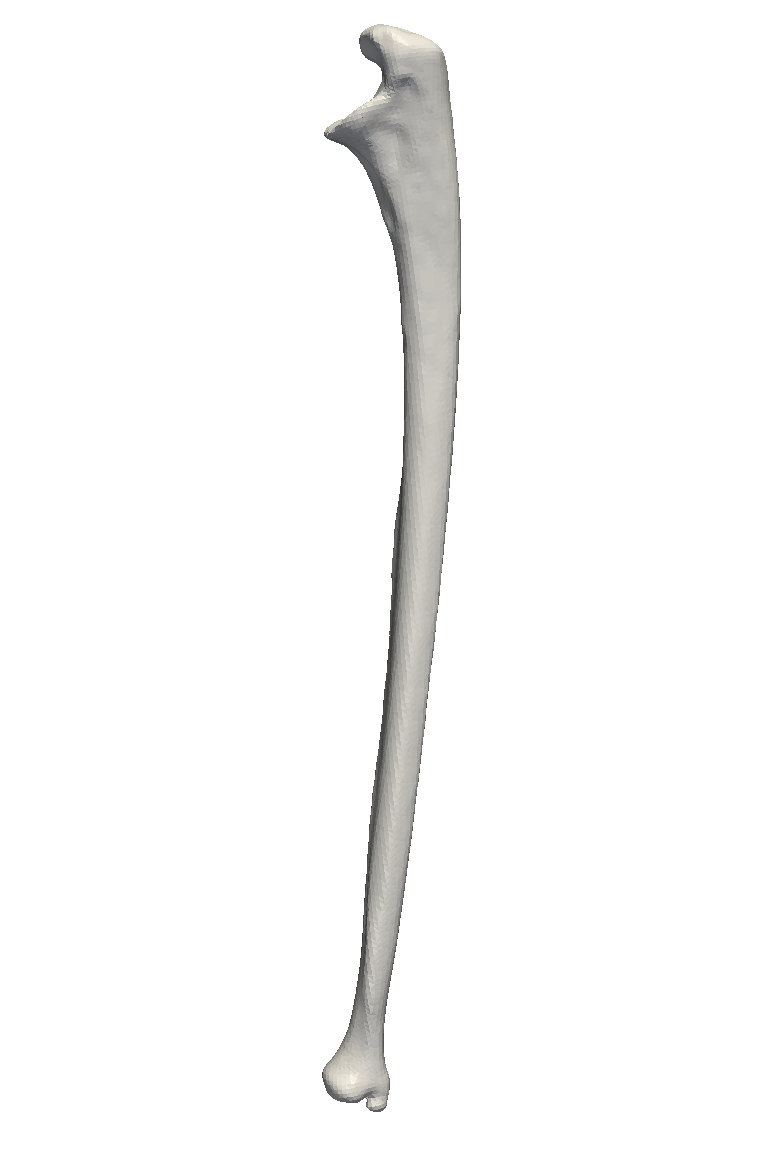}} 
\caption{The effect of varying the first two modes of variation for each model.}
\label{fig:ulna-prior}
\end{figure}

The first step in any application of GPMMs is building a suitable
model. In the first two examples, we concentrate on the ulna. We know from prior experience
that the deformations are smooth. We capture this by building our models using a Gaussian kernel
\[k_{g}^{(s, \sigma)}(x,x')=s\mathcal{I}_{3 \times3} \exp(-\norm{x-x'}^2/\sigma^2),
\]
where $s$ determines the scale of the deformations and $\sigma$ the smoothness.
The simplest model we build is an isotropic Gaussian model defined using only a single kernel $k_g^{(100, 100)}(x,x')$. The next, more complex model is an (isotropic) multi-scale model that models deformations on different scale levels:
\[k_{ms}(x,x') = \sum_{i=1}^3 k_g^{(100/i, 100/i)}(x,x')\]
In the third model, we include the prior knowledge that for the long bones, the dominant shape variation corresponds to the length of the bone. We capture this in our model by defining the anisotropic covariance function
 \[
k_{ams}(x,x')=RS k_{g}^{(150, 100)}(x,x')S^TR^T + k_{g}^{(50, 50)}(x,x') + k_{g}^{(30 , 30)}(x,x'),
\] where $R \in \R^{3\times 3}$ is the matrix of the main principal axis of the
reference and $S=\diag(1, 0.1, 0.1) \in \R^{3\times3}$ is a scaling
matrix\footnote{Using the rules given in \cite{shawe2004kernel}, (Chapter 3, Proposition 3.22), it is easy to show that this kernel is positive definite)}. Multiplying with the matrix $SR$ has the effect that the scale of the deformations in the
direction of the main principal axis (i.e.\ the length axis) is amplified 10 times compared to 
the deformations in the other space directions.   We compute for each model the low-rank approximation, where we choose the number of basis functions such that 99 \% of the total variance of the model is approximated. Figure~\ref{fig:ulna-prior} shows the first mode of variation of the three models. We observe that for the anisotropic model, the main variation is almost a pure scale variation in the length axis, while in the other models it goes along with a bending of the bone. 

Following Styner et al.\ we evaluate these three models using the standard criteria generalization, specificity and compactness \cite{styner2003evaluation}. Generalization refers to the model's ability to accurately represent all valid instances of the modelled class. We will discuss it in the next subsection. Specificity refers to the model's ability to only represent valid instances of the modelled bone. It is evaluated by randomly sampling instances of the model and then determining their distance to the closest example of a set of anatomically normal training examples.  Compactness, is the accumulated variance for a fixed number of components. This reflects the fact that if two models have the same generalization ability, we would prefer the one with less variance. Table~\ref{tab:specificity-and-compactness} summarizes the specificity and compactness for these models. We evaluated both measures once  consider only the first component, and once with the full model. We see the anisotropic model is more specific and more compact than the other models, which means that it should lead to more robust results in practical applications. 

\begin{table}

\begin{tabular}{l|cc|cc}
  Model & \multicolumn{2}{l}{Specificity} & \multicolumn{2}{l}{Compactness} \\  \hline 
  & 1st PC & Full model  & 1st PC & Full model \\ \hline
  Gauss & 2.6 & 5.8 & 50.6 & 299.1  \\
  Isotropic Multiscale & 2.3 &6.1 & 51.1 & 317.0 \\
  Anisotropic Multiscale & 1.9 & 2.9 & 51.1 & 137.1 \\
\end{tabular}
\caption{The specificity and compactness values computed for each of the three models. The lower the specificity and compactness the better.}
\label{tab:specificity-and-compactness}
\end{table}

\subsection{Surface registration}
\begin{figure}
  \includegraphics[width=\textwidth,height=5cm]{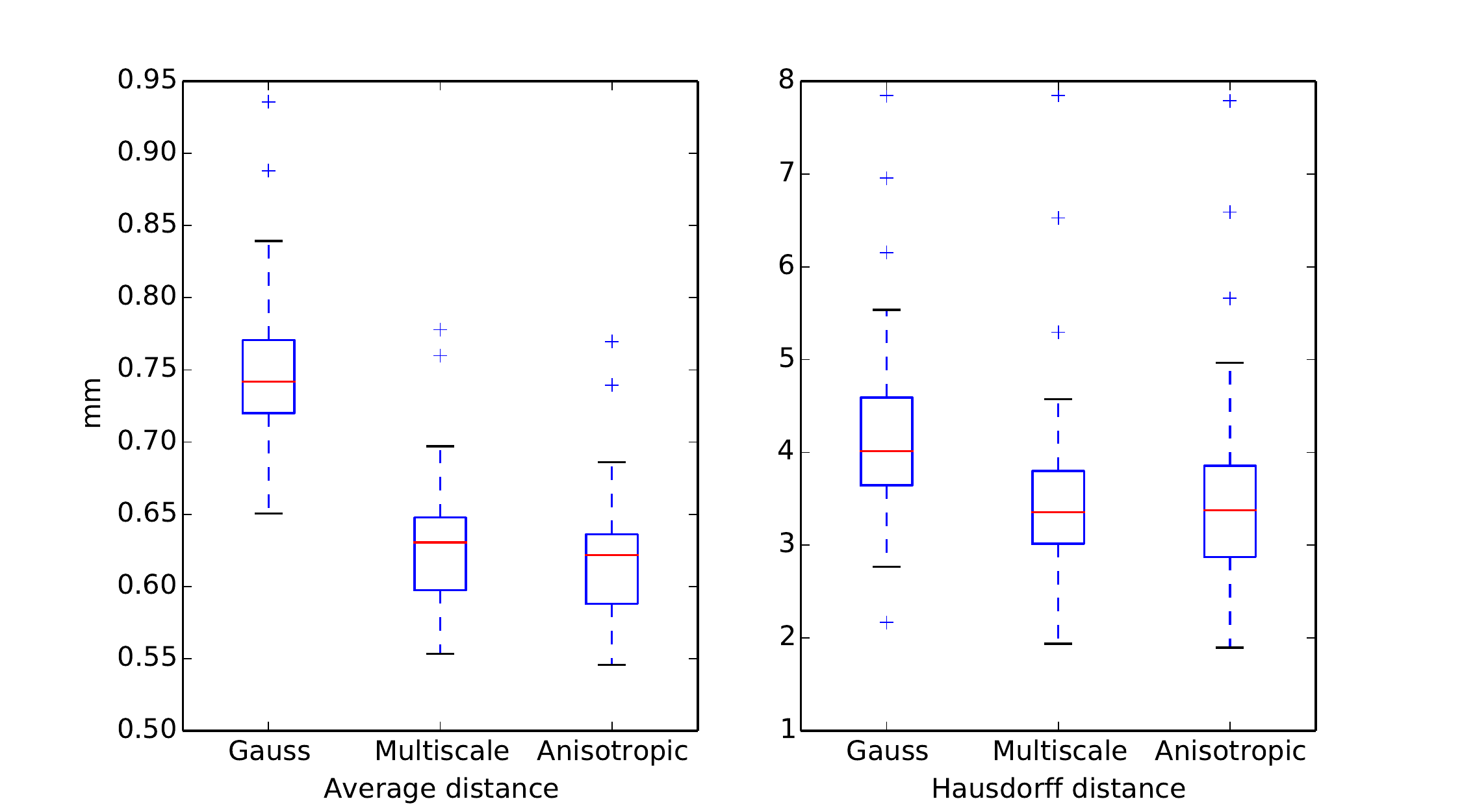}
\caption{Generalization ability measured by fitting the different models to all ulna surfaces.}
\label{fig:generalization}
\end{figure}

To evaluate the generalization ability, we need to determine how well
the model can represent valid target shapes, by fitting the model to typical 
shape surfaces. To fit the model, we use the surface registration algorithm
presented in
Section~\ref{sec:surface-and-image-registration}. Figure~\ref{fig:generalization}
shows a boxplot with the generalization results. We also see that the
multiscale and the anisotropic model lead to similar results, but both
outperform model where only a simple Gaussian kernel was used. Since the anisotropic model can fit the models with the same accuracy than the multiscale model, despite being much more compact, means that it is clearly better targeted to the given application. We we will see in the last experiment, this is a big advantage in more complicated registration tasks, such as image to image registration. 

\subsection{Generalized Active Shape Model fitting}
The well known Active Shape Modelling approach \cite{cootes_active_1995} can be interpreted as 
a special case of Gaussian process registration as introduced in Section~\ref{sec:registration},  where the model is a classical SSM (i.e.\ the sample mean and covariance kernel \eqref{eq:smkernel} are used) and an iterative algorithm is used to fit the model to the image. Active Shape Model fitting is a very
successful technique for model-based segmentation. Its main drawback
is that the solution is restricted to lie in the span of the
underlying statistical shape model, which might not be flexible enough
to accurately represent the shape.  In our case, where we have only 36
datasets of the ulna available, we expect this to be a
major problem.

\begin{figure}[htb]
\includegraphics[width=\textwidth,height=5cm]{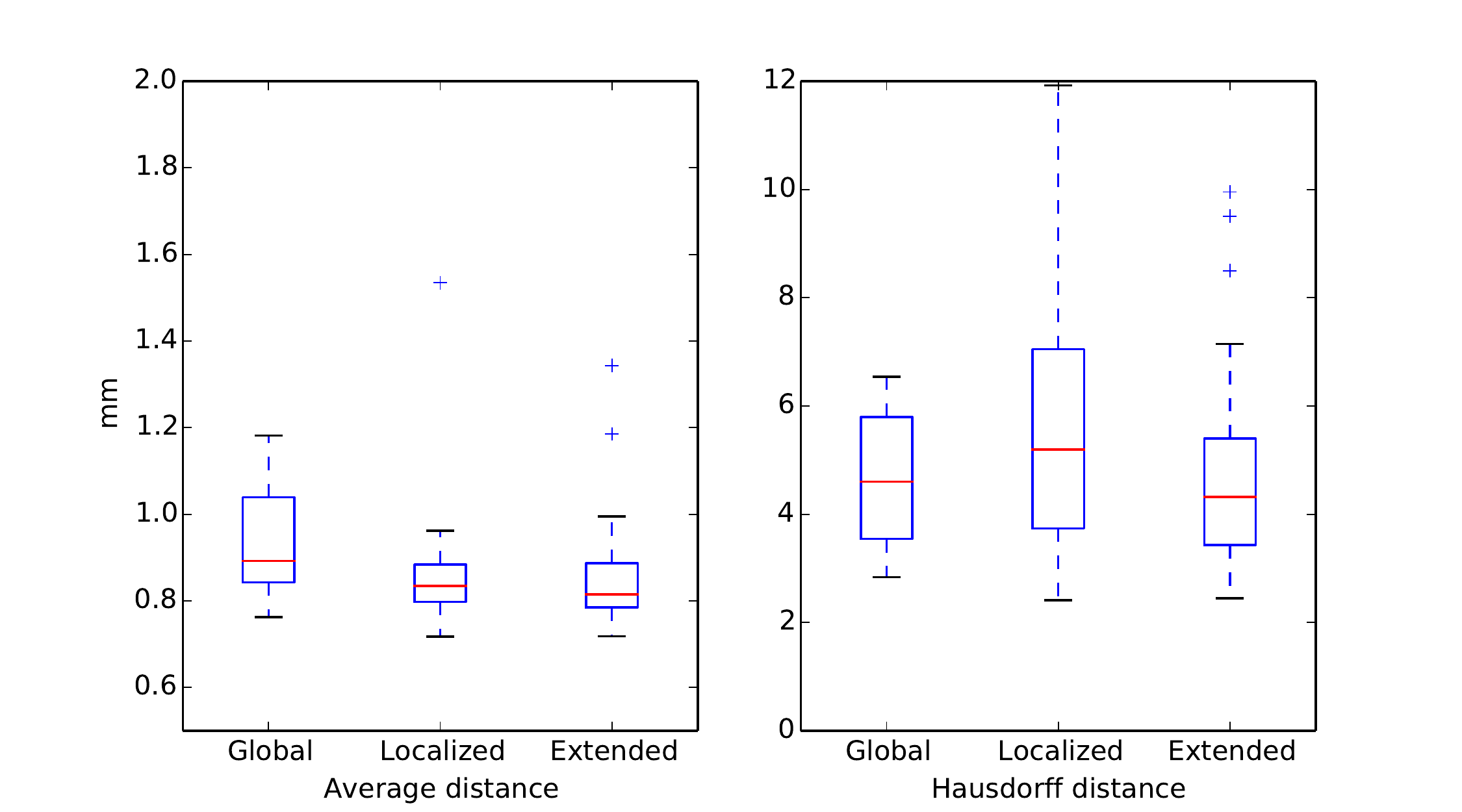}
\caption{Accuracy of the Active Shape Model fitting algorithm for three different models.}
\label{fig:asm}
\end{figure}

\begin{figure}[htb]
\includegraphics[width=\textwidth,height=5cm]{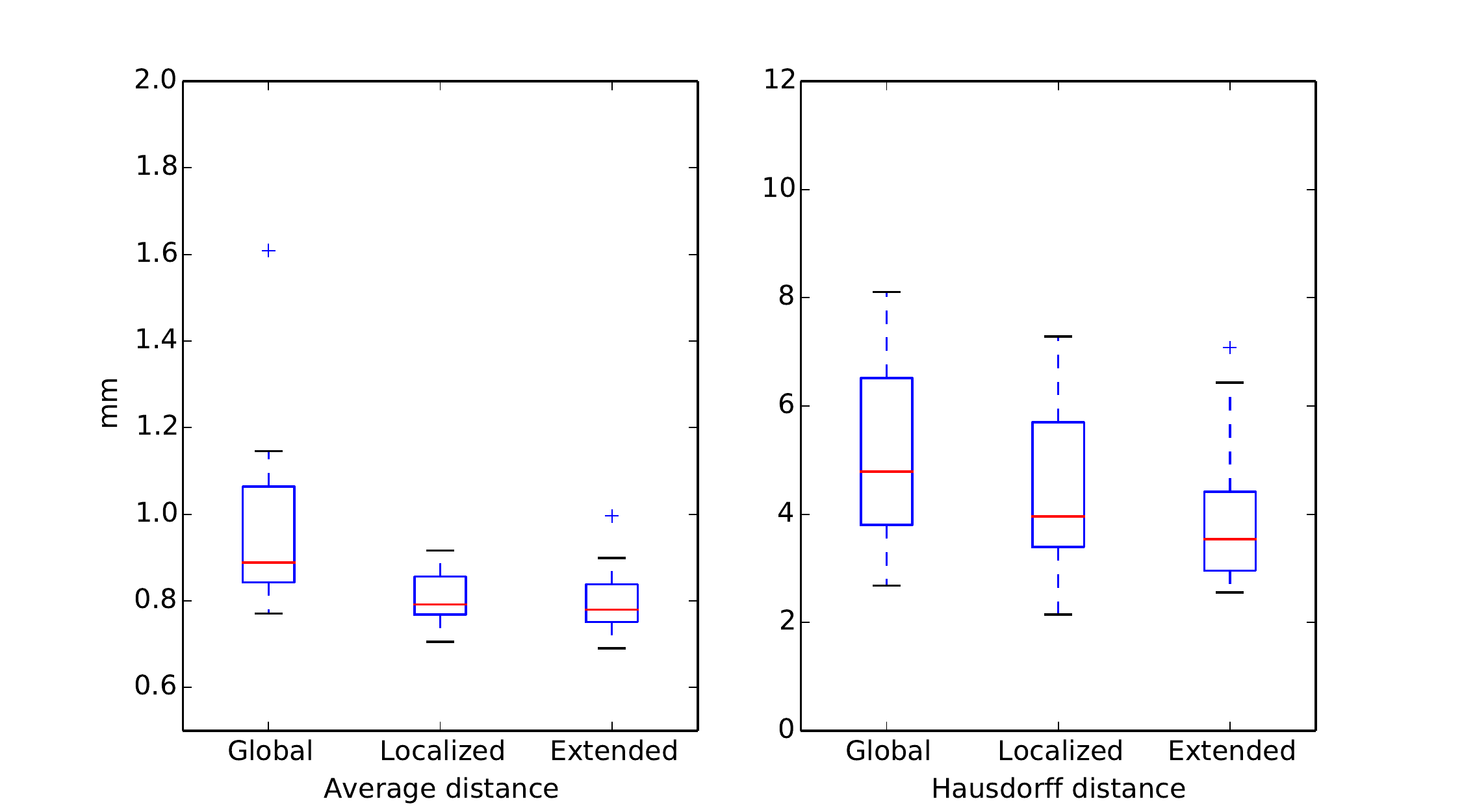}
\caption{Accuracy of the Active Shape Model fitting algorithm for three different models, when 4 landmarks at the proximal and distal ends were used to make the models more robust.}
\label{fig:asm-with-lm}
\end{figure}

To build an Active Shape Model, we use the fitting results obtained in the previous section together with the original CT images as training data. Besides a standard ASM, we use the techniques for enlarging the flexibility of shape models discussed in Section~\ref{sec:modeling-with-kernels}, to build also an extended model with additive smooth deformations (cf.\ Section~\ref{sec:bias}), and a ``localized`` model (cf.\ Section~\ref{sec:ssm-locality}). In
the first case, we use a Gaussian kernel $k_g^{(3, 100)}$ to model the unexplained part. Also for localization we choose a Gaussian kernel $k_g^{(1, 100)}$, but this time with scale 1, in order no to change the variance of the original model. In both cases, we approximate the first 100 eigenfunctions.  Figure~\ref{fig:asm} shows the corresponding
fitting result from a leave-one-out experiment. We see that both the
extended and the localized model improve the results compared to the
standard Active Shape Model. We can also observe that by adding flexibility, the model
becomes less robust and the number of outliers (i.e. bad fitting results) increases. 
We can remedy that effect by incorporating landmark constraints on the proximal and distal ends, by computing a posterior model (see
Section~\ref{sec:fixed-deformations}). This has the effect of fixing the proximal and distal ends and prevents the model from moving away too far from the correct solution.  Figure~\ref{fig:asm-with-lm} shows that this has the desired effect and the combination of including landmarks and increasing the model flexibility leads to clearly improved results.

\subsection{Image to image registration}
\begin{figure}
\includegraphics[width=\textwidth,height=6cm]{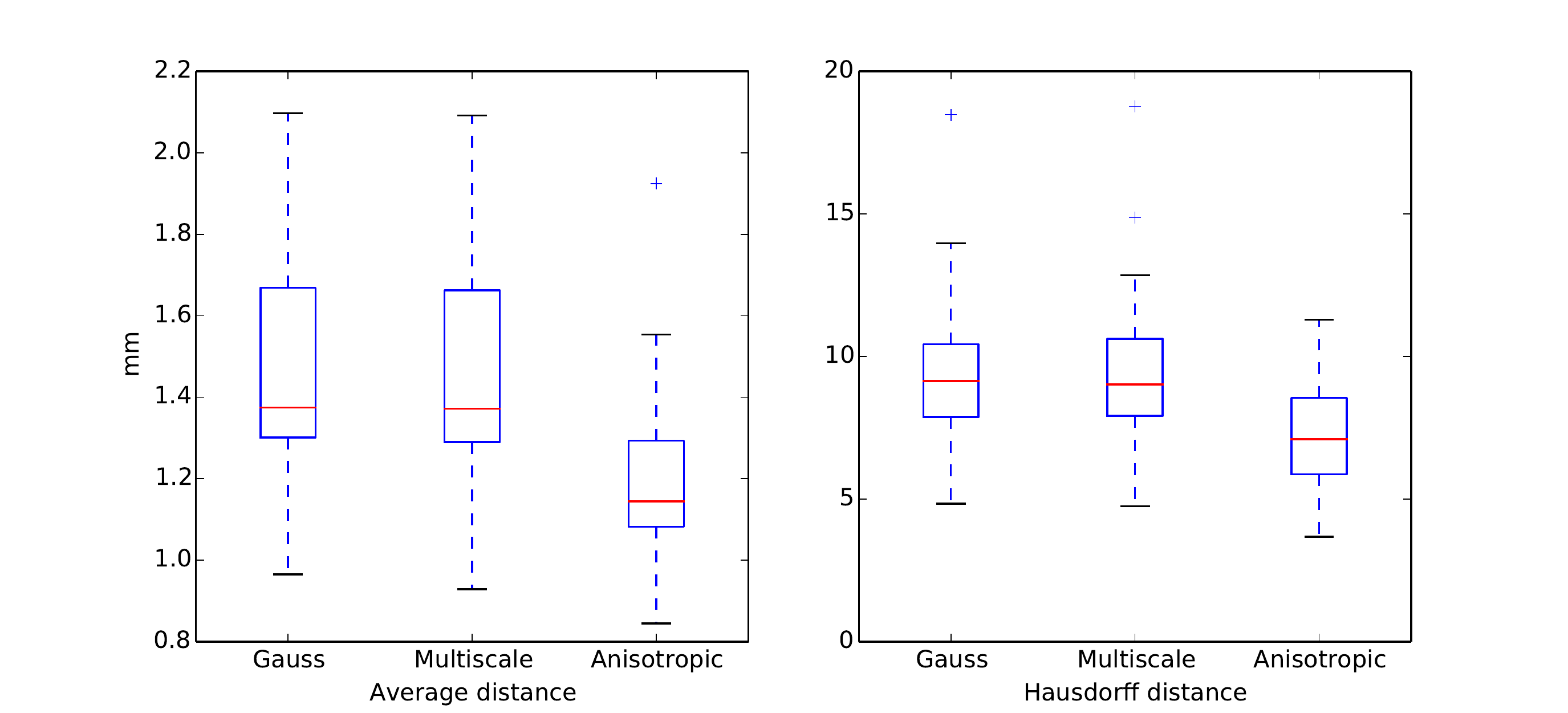}
\caption{Accuracy of image to image registration results performed with different models, compared on a ground-truth segmentation of the bones.}
\label{fig:image-reg-results}
\end{figure}

\begin{figure}
\includegraphics[width=\textwidth,height=6cm]{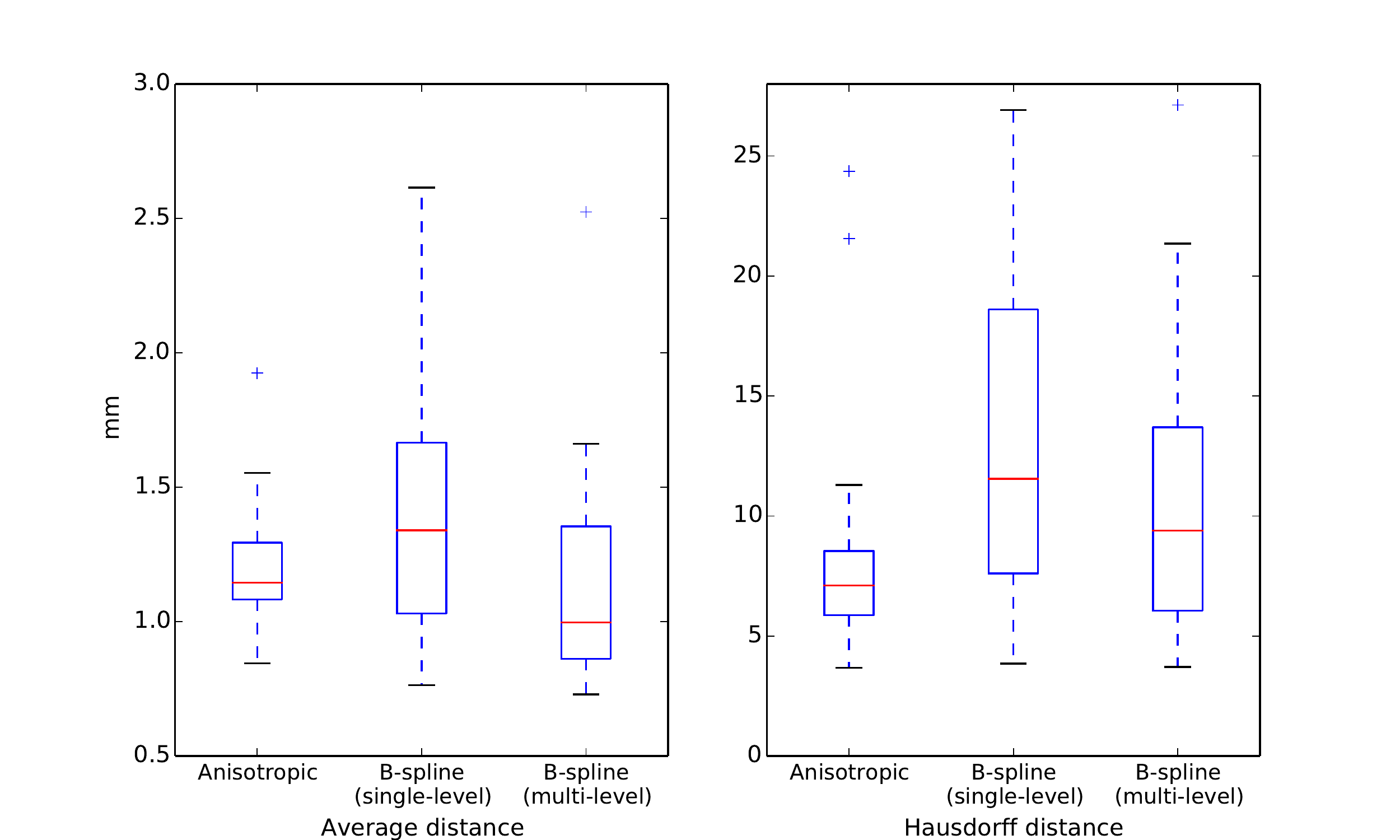}
\caption{Accuracy of image to image registration results performed with our best model, compared to a single-level and multi-level B-spline registration method implemented in Elastix.}
\label{fig:image-reg-results-bspline}
\end{figure}

In a last experiment we show that our model can also be used to perform 3D image to image registration, using the full forearm CT images. We choose one image as a reference and build a GPMM on the full image domain. In the application of GPMMs to image registration, we have to be careful about the image borders, as the basis functions are global, and hence values at the boundary might strongly influence values in the interior. 
We therefore mask the images, and optimize only on the bounding box of the bones. We use a simple mean squares metric and a stochastic gradient descent algorithm to optimize the registration functional (see Equation~\eqref{eq:image-registration}).
To evaluate the method, we warp the ground-truth segmentation of the forearm bones with the resulting deformation field and determine the distance between the corresponding surfaces.
Figure~\ref{fig:image-reg-results} shows the results for the same three models as used in the first experiment. In this example, where the optimization task is much more difficult, we see that the anisotropic model, which is much more targeted to the application, has clear advantages. 
We also compared our method to a standard B-Spline registration method \cite{rueckert_automatic_2001}, which is the
standard registration method used in Elastix. First, we use a B-spline that is only defined on a single scale level. As expected, since B-Splines are not application-specific, the result are less robust and the accuracy is worse on average (see Figure~\ref{fig:image-reg-results}). In its standard setting, Elastix uses a multi-resolution approach, where it refines the B-Spline grid, in every resolution level. This corresponds roughly to our multiscale approach, but with the important difference that new scale levels are added for each resolution level. This strategy makes the approach much more stable and, thanks to the convenient numerical properties of B-Splines, allows for arbitrarily fine deformations. As shown in Figure~\ref{fig:image-reg-results-bspline} in this multi-resolution setting, the B-Spline registration yields more accurate results on average than our method, but, as expected, is less robust.  
It is interesting to compare the two strategies in more detail. While our model has 500 parameters, the final result of the B-Spline registration has 37926 parameters. Thanks to the convenient numerical properties of B-Splines, more could be added if to increase the model's flexibility even further. This explains why the B-Spline approach can yield more accurate solutions than GPMMs. With GPMMs, the number of parameters is limited by the number of eigenfunctions we can accurately approximate (see Appendix~\ref{sec:low-rank} for a more detailed discussion). If the image domain is large compared to the scale of the features we need to match, this quickly becomes a limitation. To achieve a similar accuracy as the B-Spline approach, one possibility would be to perform a hierarchical decomposition of the domain and to define more flexible models for each of the smaller subdomains.



\section{Conclusion}
We have presented Gaussian Process Morphable Models, a generalization
of PCA-based statistical shape models. GPMMs extends standard SSMs
in two ways: First GPMMs are defined by a Gaussian process,
which makes them inherently continuous and do not force an early
discretization. More importantly, rather than only estimating the
covariances from example datasets, GPMMs can be specified using
arbitrary positive definite kernels. This makes it possible to build
complex shape priors, even in the case where we do not have many
example dataset to learn a statistical shape model. Similar to
an SSM, a GPMM is a low-dimensional, parametric model. It can be brought 
into the exact same mathematical form as an SSM by discretizing the domain on which the model is defined. Hence our generalized shape models can be used in
any algorithm that uses a standard shape model.  To make our method easily accessible, 
we have made the full implementation available as part of the open source 
framework statismo \cite{luthi_statismo-framework_2012}  and
scalismo \cite{scalismo}. 

Our experiments have confirmed that GPMMs are ideally suited for
modeling prior shape knowlege in registration problems.  As all prior
assumptions about shape deformations are encoded as part of the GPMM,
our approach achieve a clear separation between the modelling and
optimization. This separation makes it possible to use the same
numerical methods with many different priors. Furthermore, as a GPMM is generative, we can assess the
validity of our prior assumptions by sampling from the model. We have shown how the same registration
method can be adapted to a wide variety of different applications by
simply changing the prior model. Indeed, Gaussian processes
give us a very rich modelling language to define this prior, leading
to registration methods that can combine learned with generic shape
deformations or are spatially
varying. From a practitioners point of view, the straight-forward integration of landmarks may also be a valuable contribution, since it enables to develop efficient interactive registration schemes.

The most important assumption behind
our models is that the shape variations can be well approximated using
only a moderate number of leading basis functions. As shape
deformations between objects of the same class are usually smooth and
hence the deformations between neighboring points highly correlated,
this assumption is usually satisfied. Furthermore for most anatomical
shapes, fine detailed deformations only occur in parts of the
shape. GPMMs give us the modelling power to model these fine
deformations only where they are needed. Our method reaches its limitations, 
when very fine deformations need to be modelled over a large domain, as it is sometimes required  in image registration.  In this case the approximation scheme becomes inefficient and the approximations inaccurate. An interesting extension for future work would be to devise a hierarchical, multi-resolution approach, which would partition the domain in order and perform separate approximation on smaller sub-domain. In this way, the modelling power of GPMMs could be exploited to model good priors for image registration, while still offering all the flexibility of classical image registration approaches.

We hope with this work to bridge the gap between the so far distinct world of
classical shape modelling, where all the modelled shape variations are
a linear combination of the training shapes, and the word of
registration, where usually oversimplistic smoothness priors are
used. We believe that it is the middle ground between these two
extremes, where shape modelling can do most for helping to devise
robust and practical applications.


\newpage


%

\appendix

\section{Accuracy of the low-rank approximation} \label{sec:low-rank}

The success of our method strongly depends on how well the truncated KL-Expansion
\begin{equation}
	\tilde{u}(x) \sim \mu(x) + \sum_{i=1}^r \alpha_i \sqrt{\lambda}_i \phi_i(x), \, \alpha_i \in \N(0,1),
\end{equation}
approximates the full Gaussian process $u \sim GP(\mu, k)$  (cf. Section~\ref{sec:gpmms}). 
The final approximation error depends on one hand on the error we make by approximating the Gaussian process using the $r$ leading eigenfunctions only, and on the other hand also on how accurately we can compute these eigenfunctions using the Nystr\"om method. 

\subsection{An analytic example}
\begin{figure}
\subfloat[]{\includegraphics[width=0.5\columnwidth]{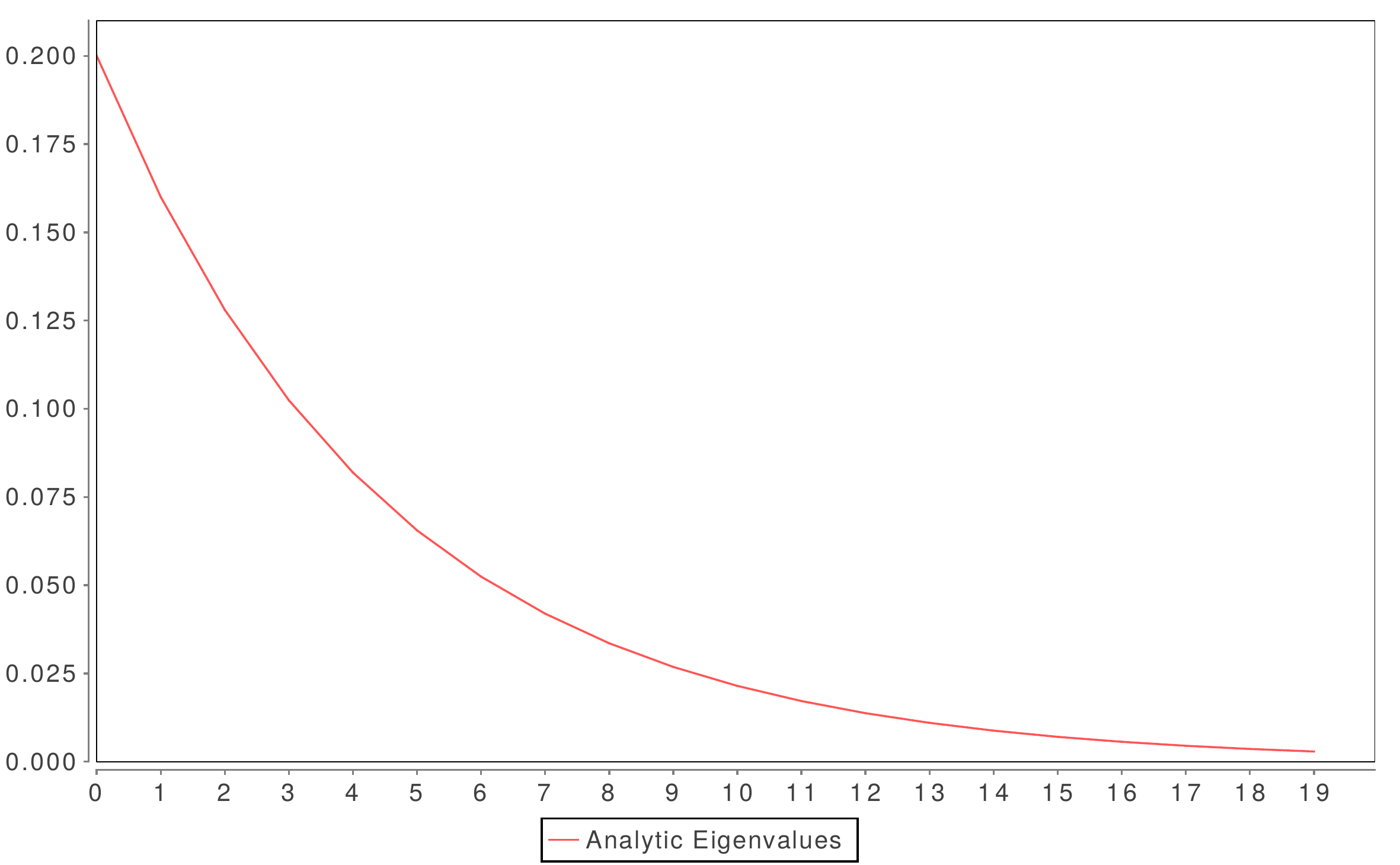}\includegraphics[width=0.5\columnwidth]{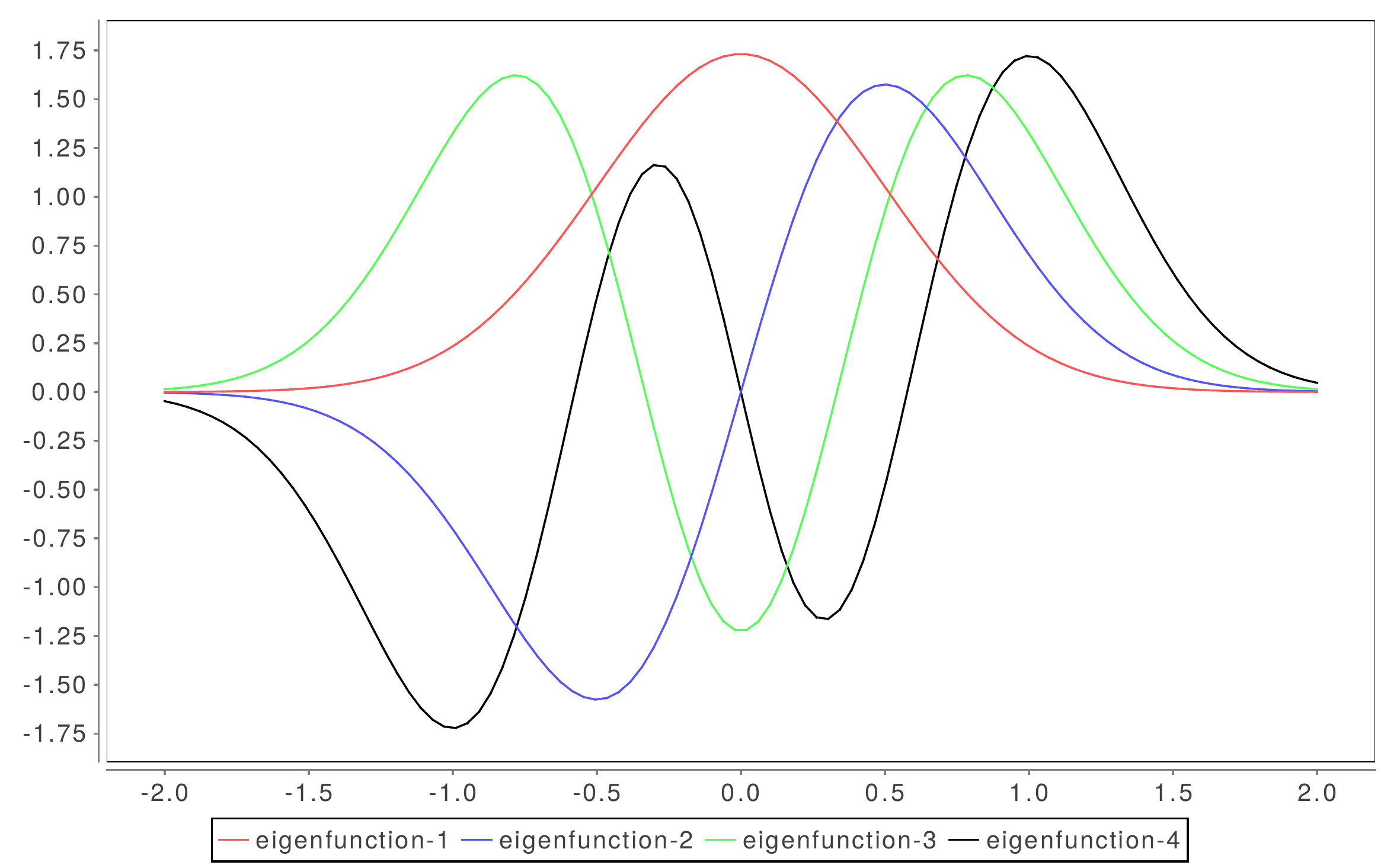}} \\
\subfloat[]{\includegraphics[width=0.5\columnwidth]{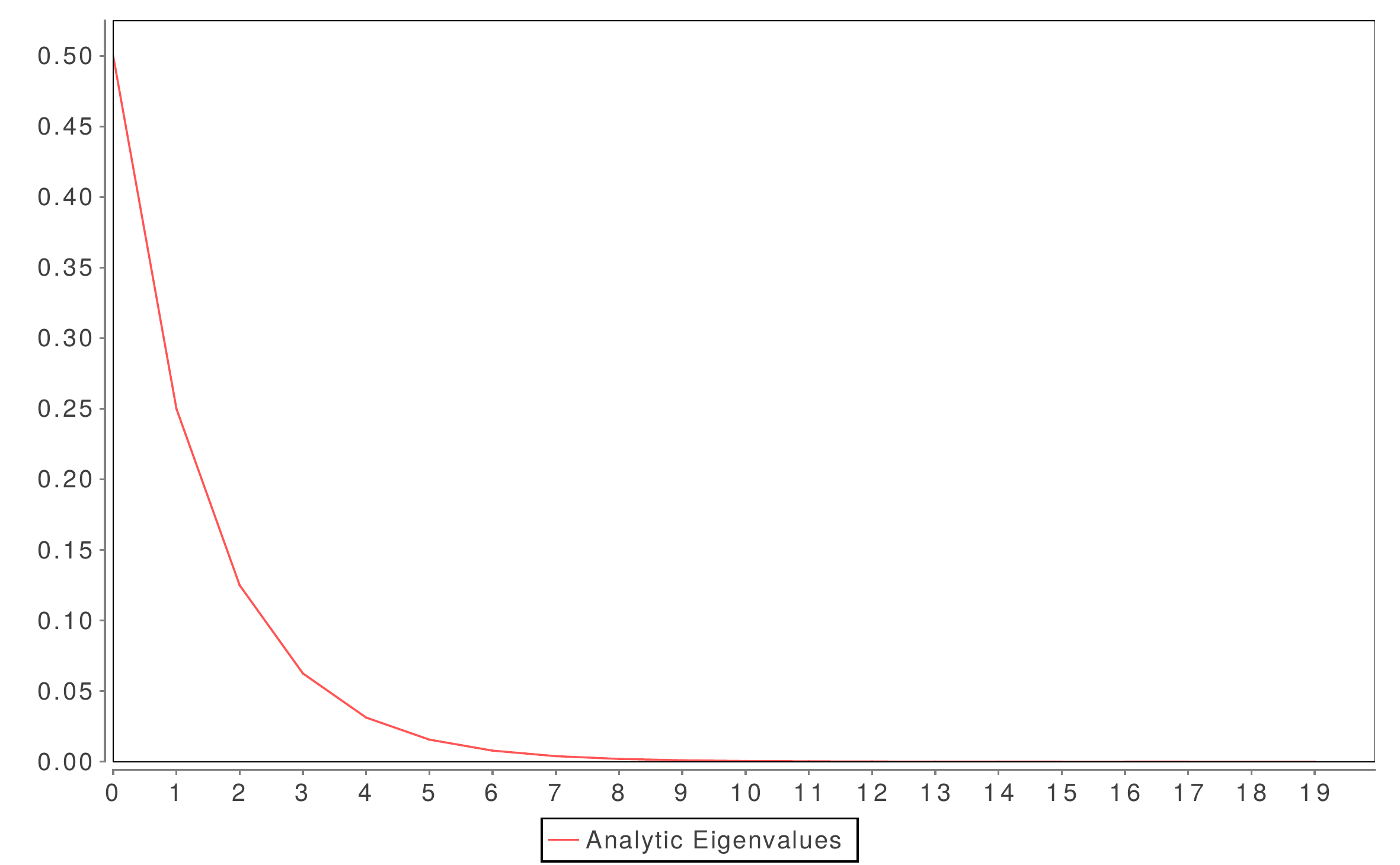}\includegraphics[width=0.5\columnwidth]{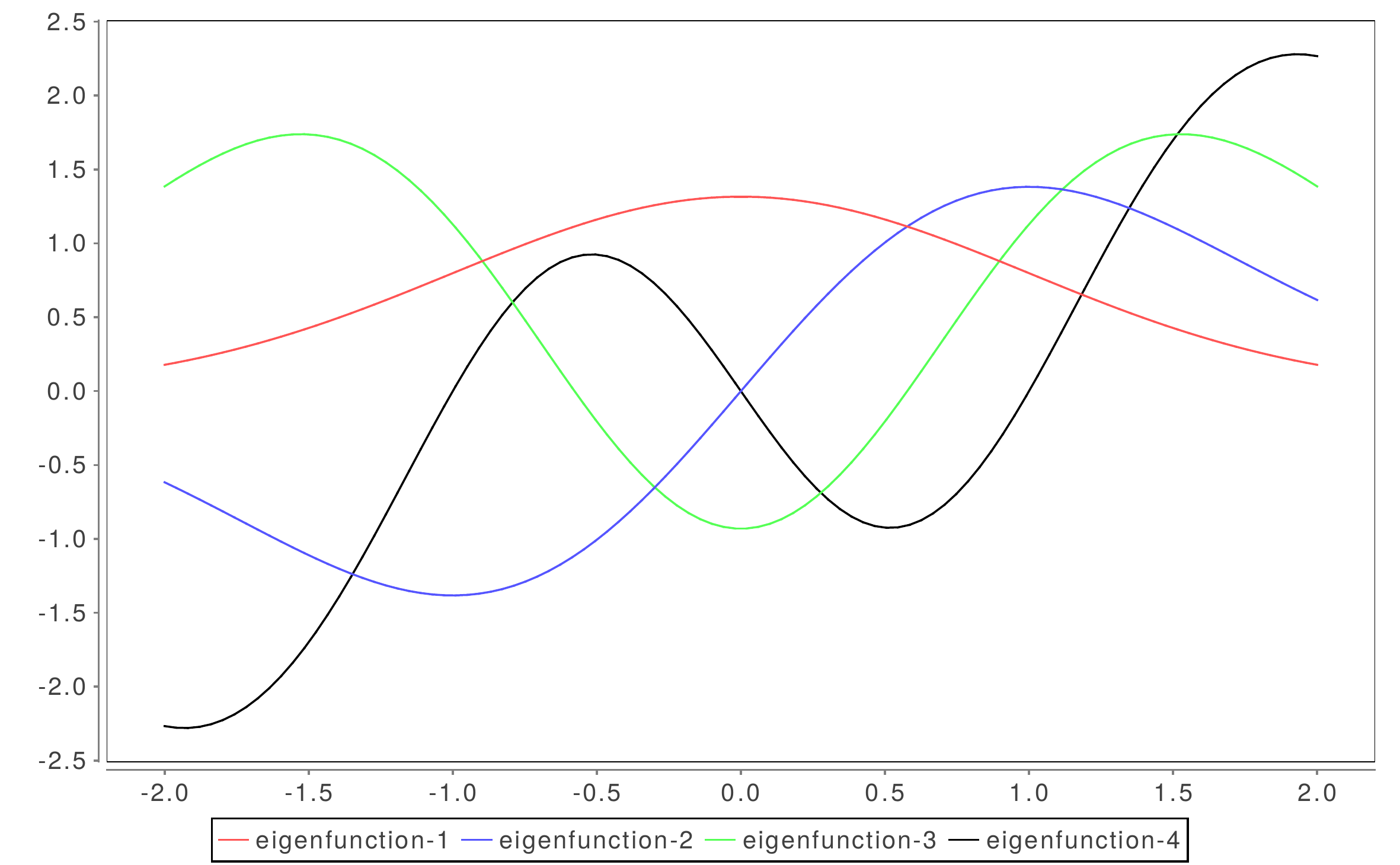}} \\
\subfloat[]{\includegraphics[width=0.5\columnwidth]{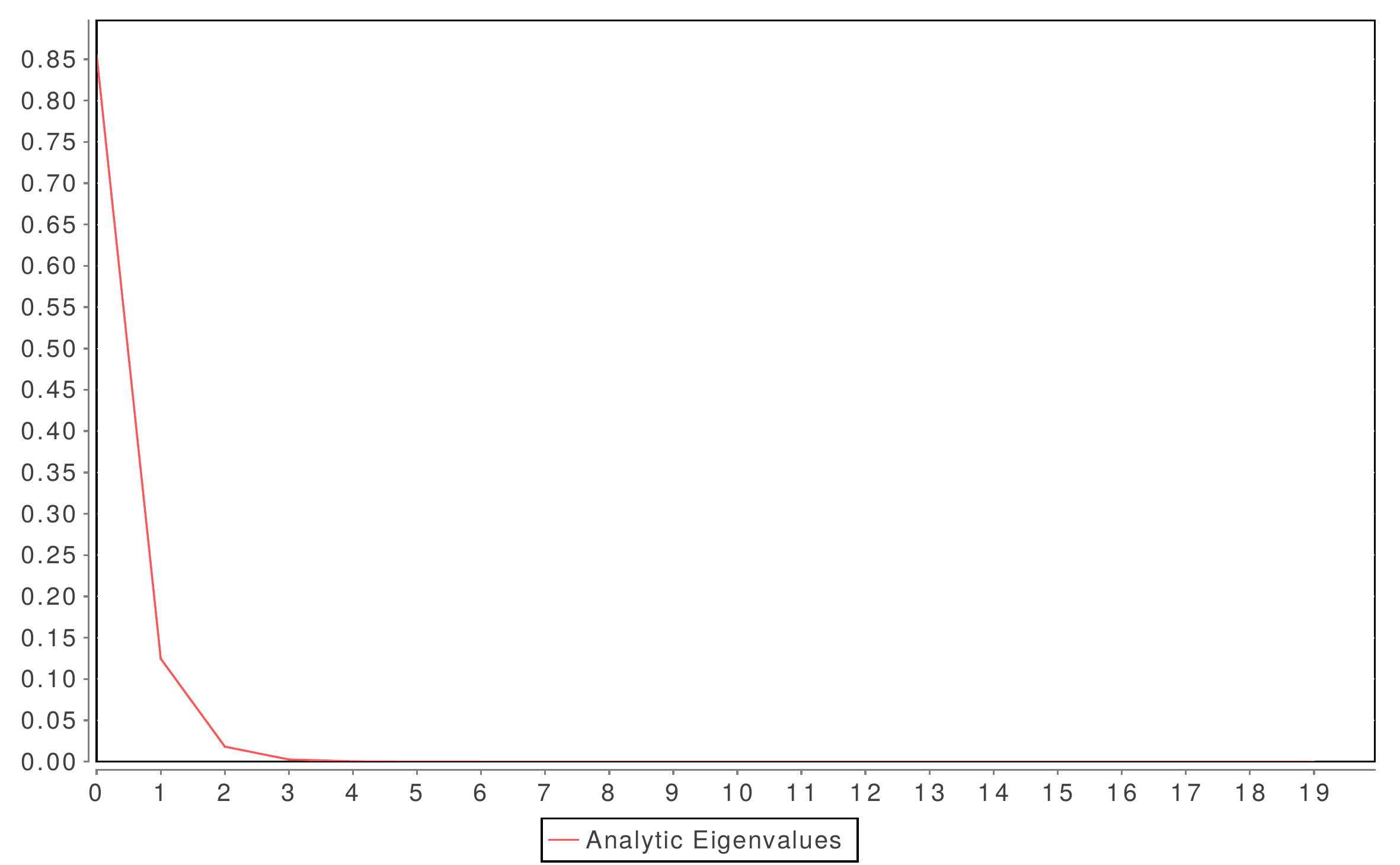}\includegraphics[width=0.5\columnwidth]{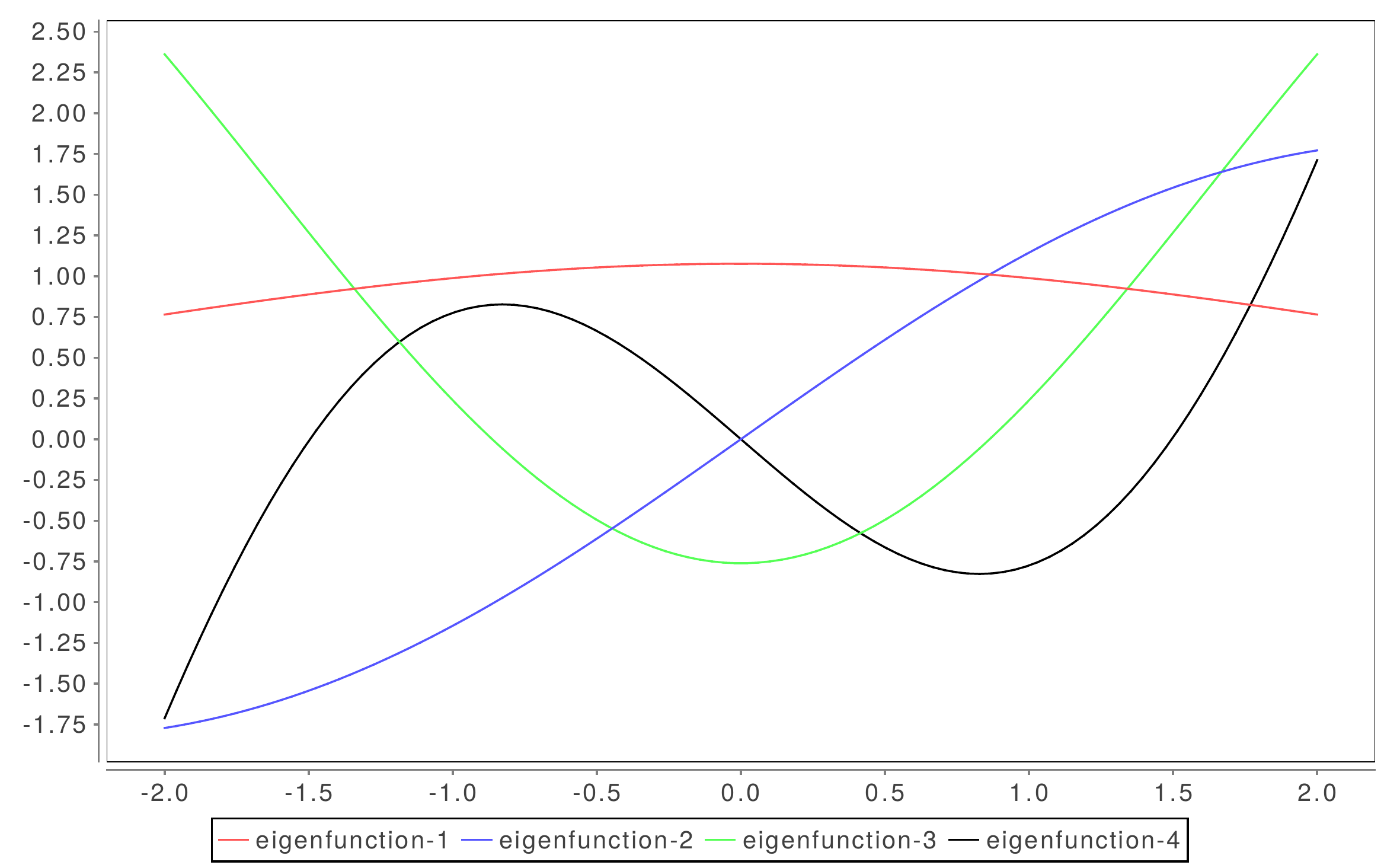}}
\caption{The first eigenfunctions and the eigenvalues spectrum for $\sigma=0.1$, $\sigma=1.0$ and $\sigma=10.0$.}
\label{fig:analytic-eigenvalues}\end{figure}

\begin{figure}
\subfloat[]{\includegraphics[width=0.5\columnwidth]{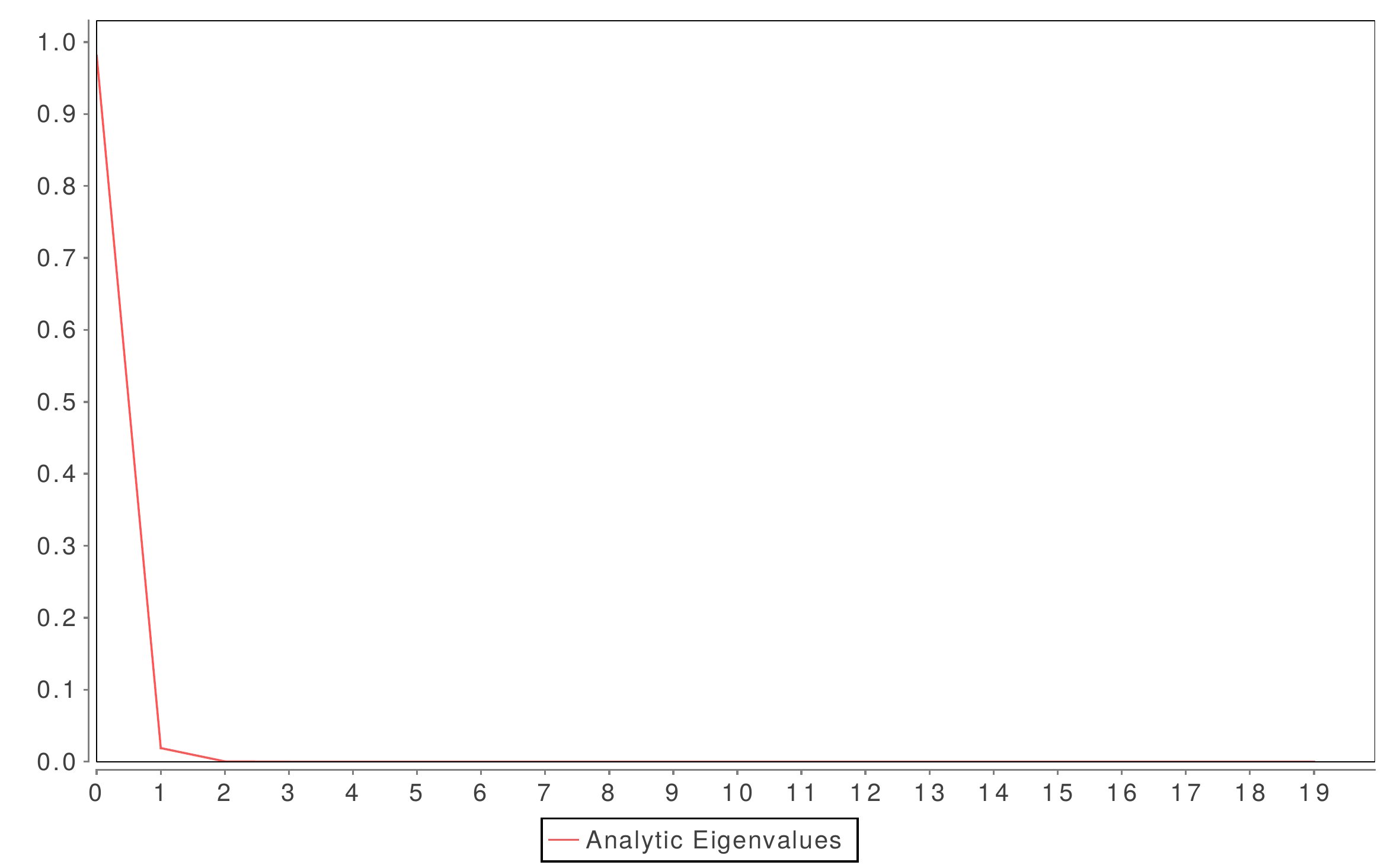}\includegraphics[width=0.5\columnwidth]{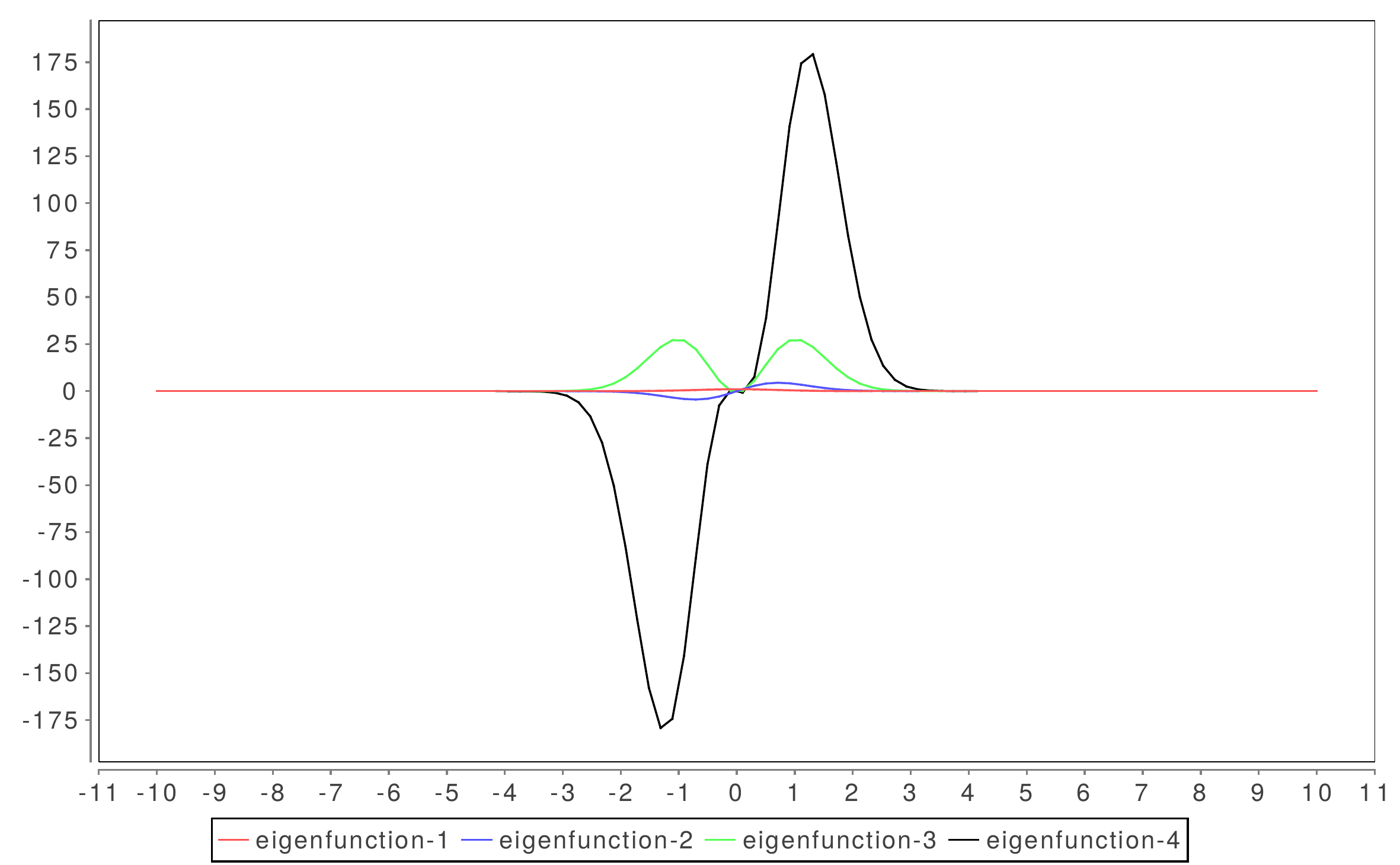}} \\
\subfloat[]{\includegraphics[width=0.5\columnwidth]{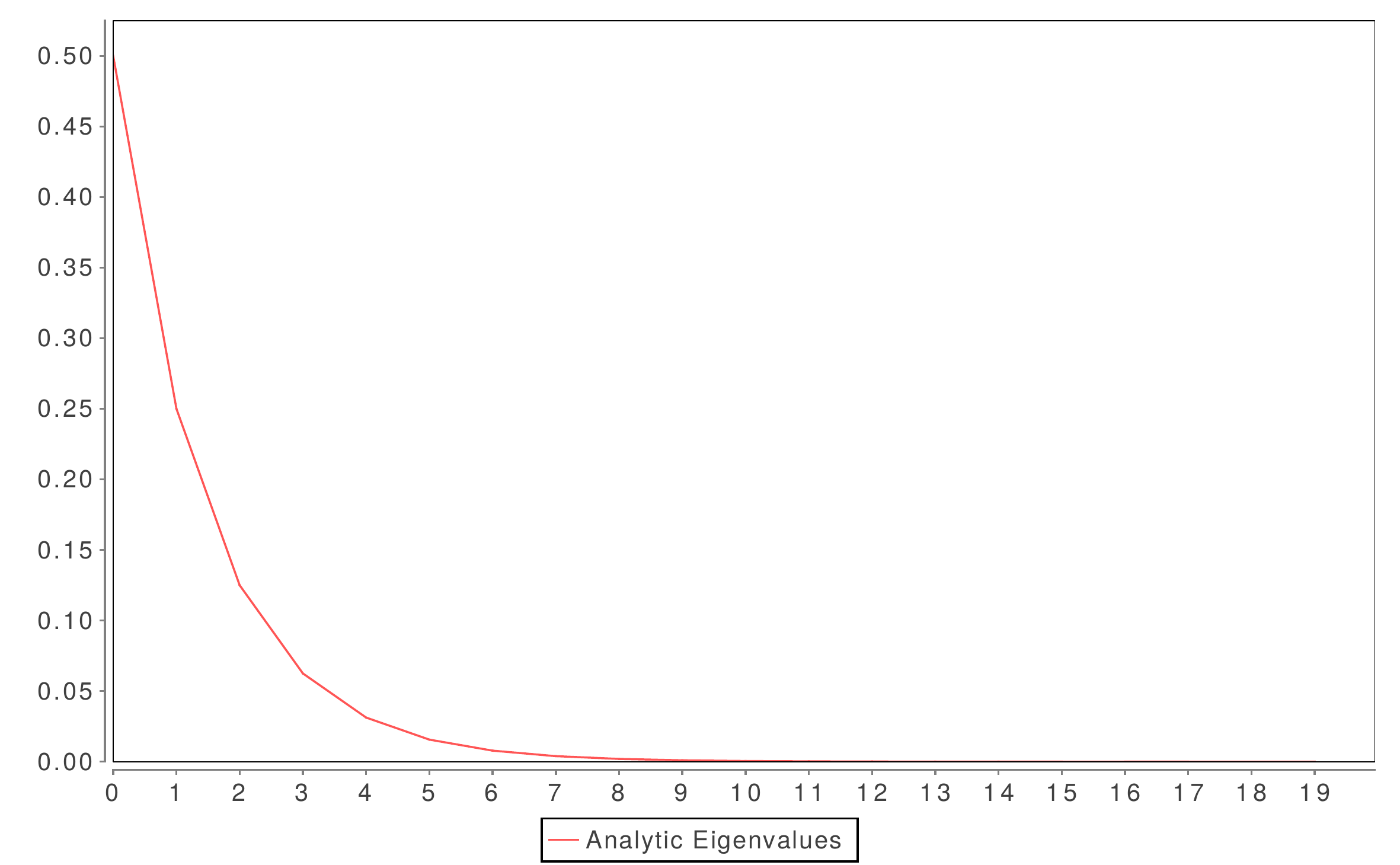}\includegraphics[width=0.5\columnwidth]{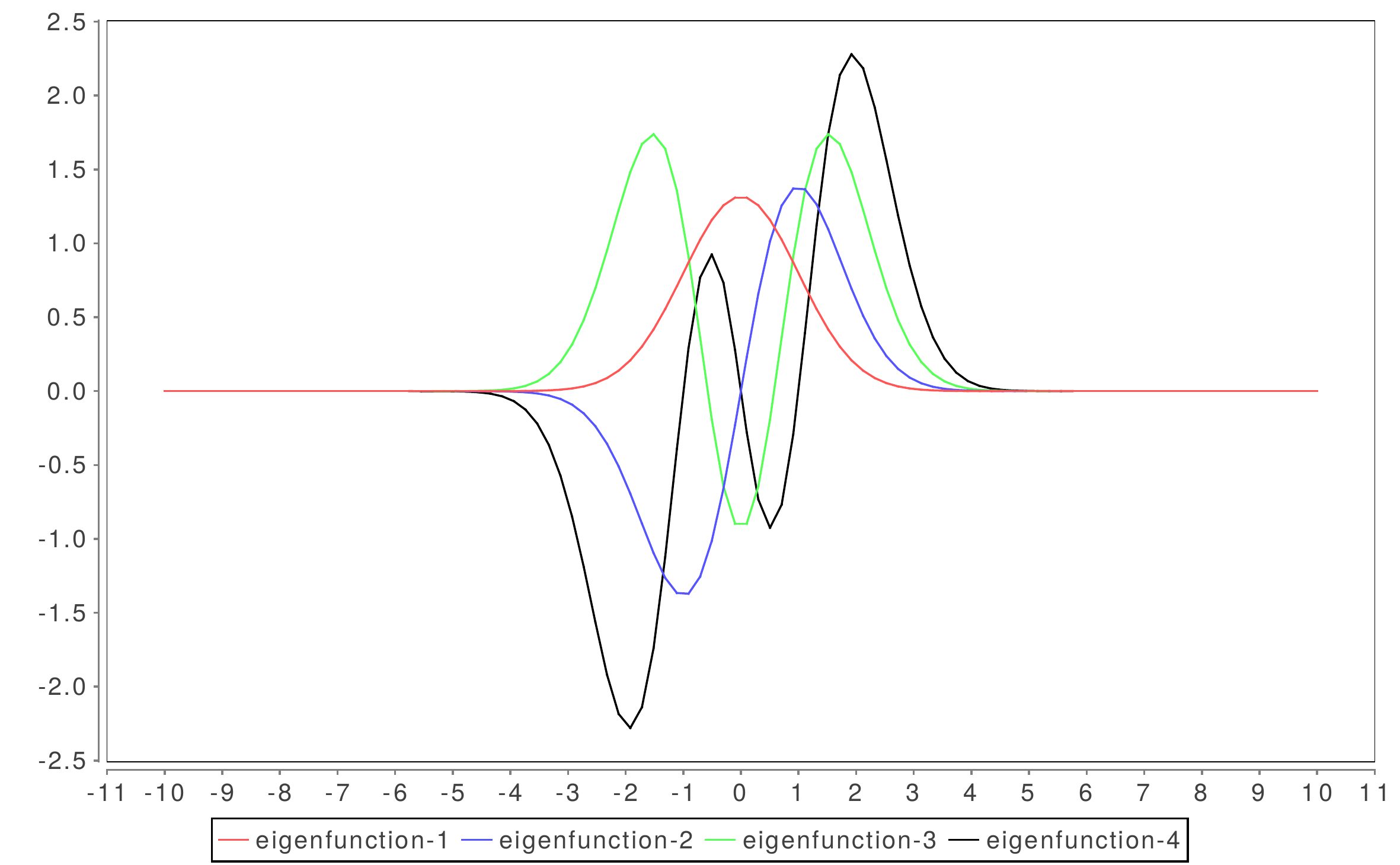}} \\
\subfloat[]{\includegraphics[width=0.5\columnwidth]{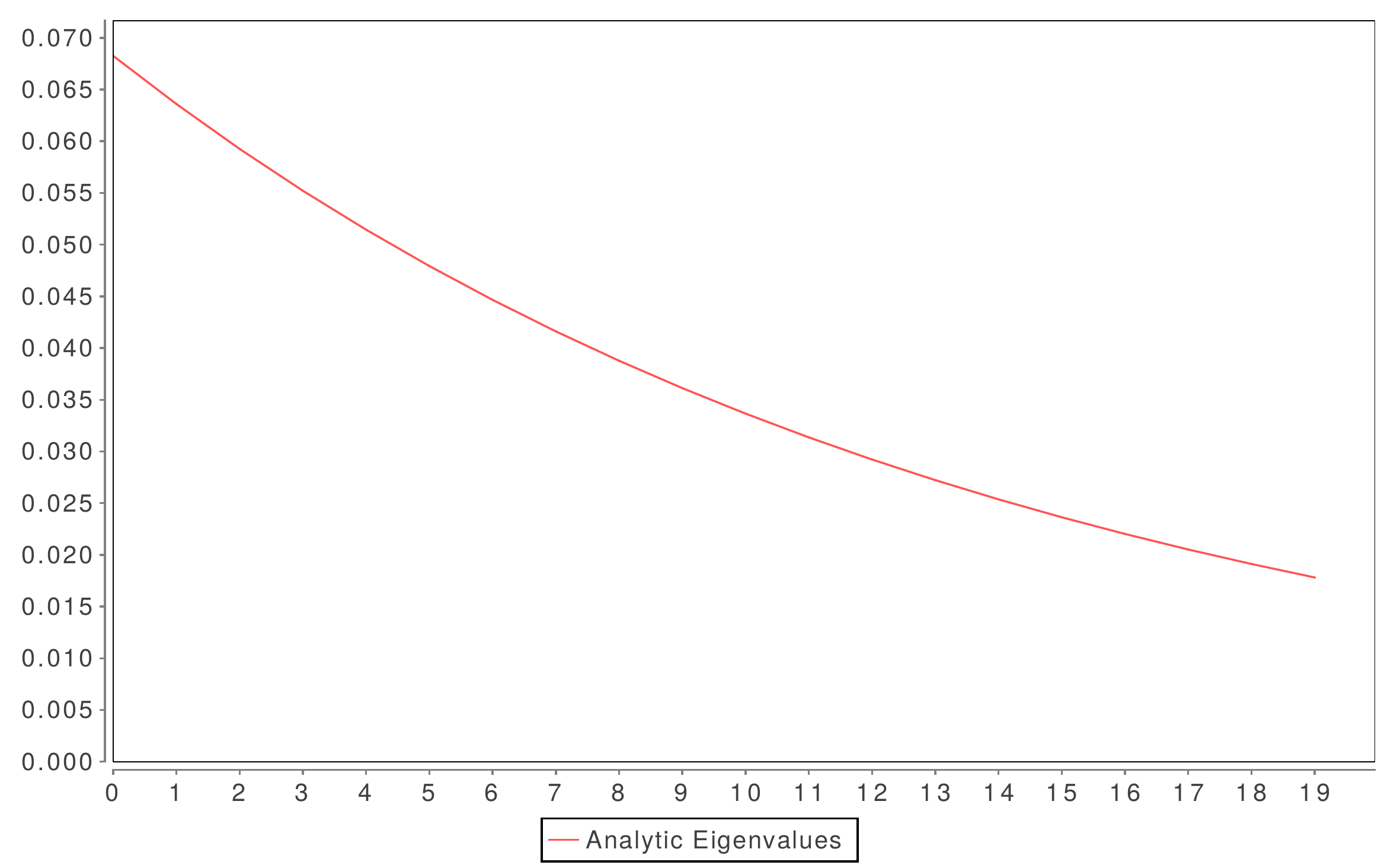}\includegraphics[width=0.5\columnwidth]{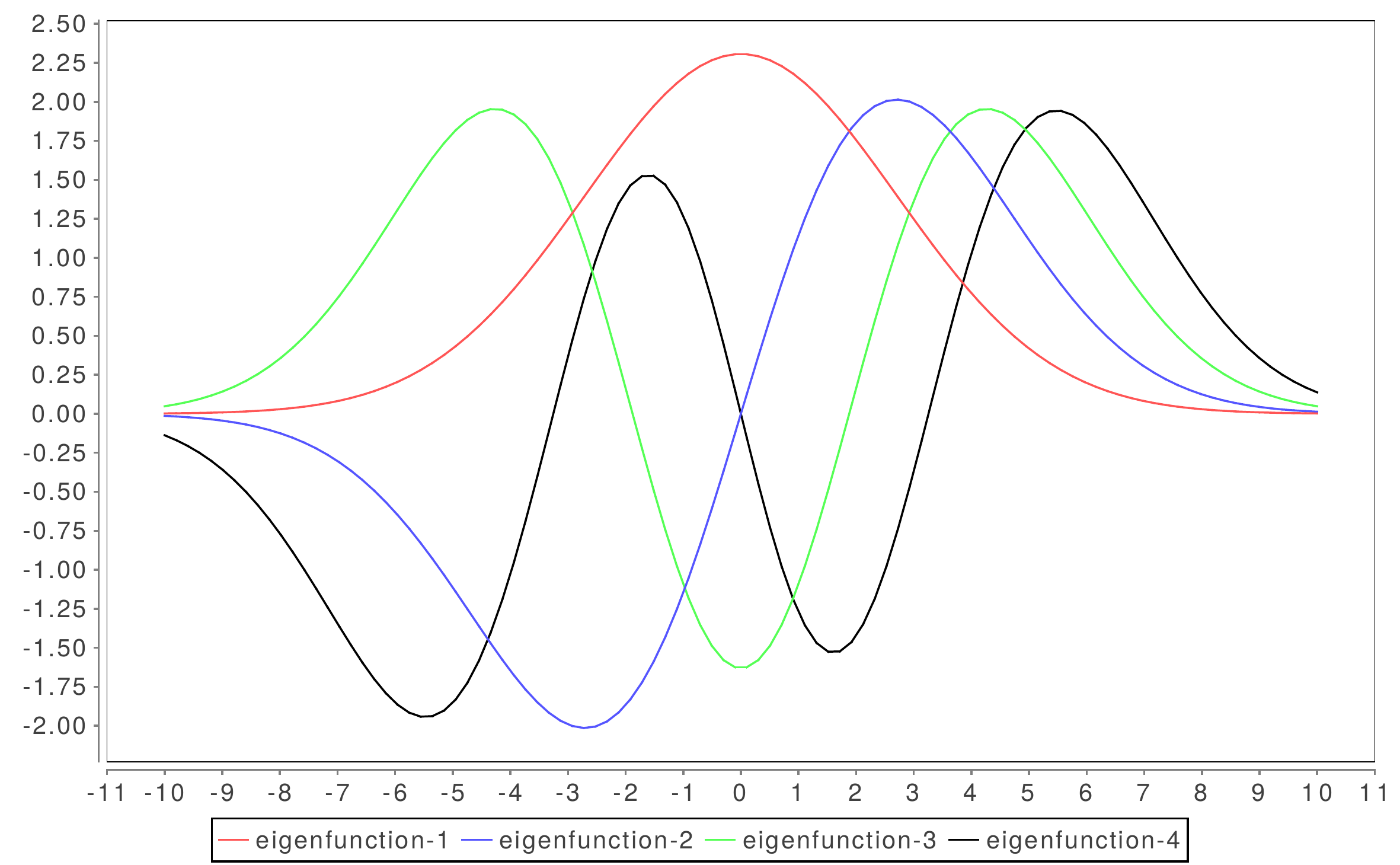}}
\caption{The first eigenfunctions and the eigenvalues spectrum for different sizes of the support  $s^2=0.1$, $s^2=1.0$ and $s^2=10.0$.}
\label{fig:analytic-eigenvalues-support}
\end{figure}

Before studying these approximations, we illustrate the trade-offs that we face 
on the example of the Gaussian kernel
\[
k(x,y)  = \exp(-\norm{x-y}^2/\sigma^2),
\]
for which the eigenfunctions are known. 
The example is given in Zhu et al.\ \cite{zhu_gaussian_1998}, and the discussion is
loosely based on a similar investigation by Williams et al. \cite{williams_effect_2000}.  
The eigenvalues
$\lambda_i$ and eigenfunctions $\phi_i$ for $i=1, \ldots, r$ of the
integral operator $\mathcal{T}_k$ associated to $k$ (cf. Equation~\eqref{eq:integral-operator}), with
respect to the distribution $\rho(x) \sim N(0, s^2)$ are given by,
\begin{align}
\lambda_i &= \sqrt{\frac{\pi}{A}}B^i  \\
\phi_i(x) &= \exp(-(c-a)x^2)H_i(\sqrt{2c}x) 
\end{align}
where
$H_i$ is the $i-$th order Hermite polynomial, and
\begin{align}
&a^{-1} = 4\sigma^2, \; b^{-1} = s^2, \; c=\sqrt{a^2 + 2 ab}, \; \\
& A = a + b + c, \; B = b/A.
\end{align}
Figure~\ref{fig:analytic-eigenvalues} shows a plot of the
first few eigenfunctions. We observe that the eigenfunctions are
global, even though the kernels are highly localized. We also observe
that the larger the bandwidth $\sigma$ of the Gaussian kernel, the lower
the frequency of the leading eigenfunctions. Furthermore, the spectrum
decreases more rapidly. Hence, the larger we chose $\sigma$ (i.e.\ the more
smooth the sample functions are) the fewer basis functions we need
to accurately represent the full Gaussian process.
A similar effect can be observed if we change the support of the domain over which 
the kernel is defined. In our example, we can simulate an increasing support by 
changing the variance $s^2$ of the probability measure $\rho(x)$.
As Figure~\ref{fig:analytic-eigenvalues-support}
confirms, the larger we choose $s^2$, the faster the eigenvalues decay. 
We see that it is the ratio between the support of the domain and the smoothness, which determines
how many basis functions we need to achieve a good approximation. 
While we have only shown it for the Gaussian kernel, it is intuitively
clear that similar observations hold for any kernel that enforces
smoothness. This can be seen by noting that the more smoothness is enforced by the kernel, 
the more will the points of the domain vary together. Hence, it is possible to capture
more of the variance using only a few basis function.

\subsection{Approximation accuracy of the eigenfunction computations}
In virtually all practical applications, it is not possible to obtain
an analytic expression of the eigenfunctions. Fortunately, it is often
possible to obtain good approximations by means of numerical
procedures, such as the Nystr\"om approximation that we presented in
Section~\ref{sec:eigendecomposition}. Bounds for the quality of this
approximation error have recently been given by Rosasco et
al. \cite{rosasco_learning_2010}. We repeat here the two main results,
which give us an insight on on the main influence factors for the
approximation quality, as a function of the number of nystrom points
$n$. 
\begin{theorem}[Rosasco et al.\ \cite{rosasco_learning_2010}] \label{thm:bound-eigenvalues}
Let $\mathcal{T}_k$ denote the integral operation associated to $k$ and $K$ be the kernel matrix defined by 
$k_{ij}=k(x_i, x_j)$. Further, let $\kappa=\sup_{x \in \Omega} k(x, x)$. 
There exist an extended enumeration $(\lambda_i)_{i \ge 1}$ of discrete eigenvalues for $\mathcal{T}_k$ and an extended enumeration 
$(\hat{\lambda}_i)_{i \ge 1}$ of discrete eigenvalues of $K$ such that 
\[
\sum_{i \ge 1} (\lambda_i - \hat{\lambda}_i)^2 \le \frac{8\kappa^2\tau}{n}
\] 
with confidence greater than $1-2\exp{\tau}$.
In particular 
\[
\sup_{i \ge 1} \abs{\lambda_i - \hat{\lambda}_i} \le \frac{2 \sqrt{2}\kappa\sqrt{\tau}}{\sqrt{n}}.
\]
\end{theorem}
\begin{theorem}[Rosasco et al.\ \cite{rosasco_learning_2010}] \label{thm:bound-eigenfunctions}
  Given an integer $N$, let $m$ be the sum of the multiplicities of the first $N$ distinct eigenvalues of $\mathcal{T}_K$, so that
  \[
  \lambda_1 \ge \lambda_2 \ge \ldots, \ge \lambda_m \ge \lambda_{m+1},
  \]
  and $P_N$ be the orthogonal projection from $L^2(\Omega, \rho)$ onto the span of the corresponding eigenfunctions. 
  Let $r$ be the rank of $K$, and $\hat{u}_1, \ldots, \hat{u}_k$ the eigenvectors corresponding to the nonzero eigenvalues of $K$ in a decreasing order. 
  Denote by $\hat{v}^1, \ldots, \hat{v}^k \subset L^2(\Omega, \rho)$ the corresponding Nystr\"om extension. 
  Given $\tau > 0$ if the number $n$ of examples satisfies
  \[
  n > \frac{128 \kappa^2 \tau}{(\lambda_m - \lambda_{m+1})^2}
  \]
  then
  \[
  \sum_{j=1}^m \norm{(I-P_N)\hat{v_j}}_\rho^2 + \sum_{j=m+1}^r \norm{P_N \hat{v}_j}_\rho^2 \le \frac{32 \kappa^2 \tau}{(\lambda_m - \lambda_{m+1})^2n},
  \]
  with probability greater than $1 - 2\exp{-\tau}$.
\end{theorem}
While these bounds are too loose to use them in practical
applications, they have a number of important implications. First, we
note that these theorems hold for an arbitrary distribution. Hence,
independently of whether we sample the points for the Nystr\"om
approximation from the surface, or in a volume, we know that
convergence is always guaranteed if the number of points $n$ is
sufficiently large. Second, the bound on the eigenvalues
(\ref{thm:bound-eigenvalues}) does not depend on the smoothness induced by the kernel, nor
does it depend on the support of the domain. Hence, for a given $n$ we
have a guaranteed accuracy independent of the smoothness induced by
the kernel. Third, the main factor that influences the projection
error (Theorem~\ref{thm:bound-eigenfunctions}) is how fast the
eigenvalues $\lambda_m - \lambda_{m+1}$ decay. As more smooth
functions imply faster decay, this implies that we need fewer points
for the approximation when the kernel induces more smoothness.
\begin{figure}
\subfloat[Spectrum $(n = 200)$]{\includegraphics[width=0.5\columnwidth]{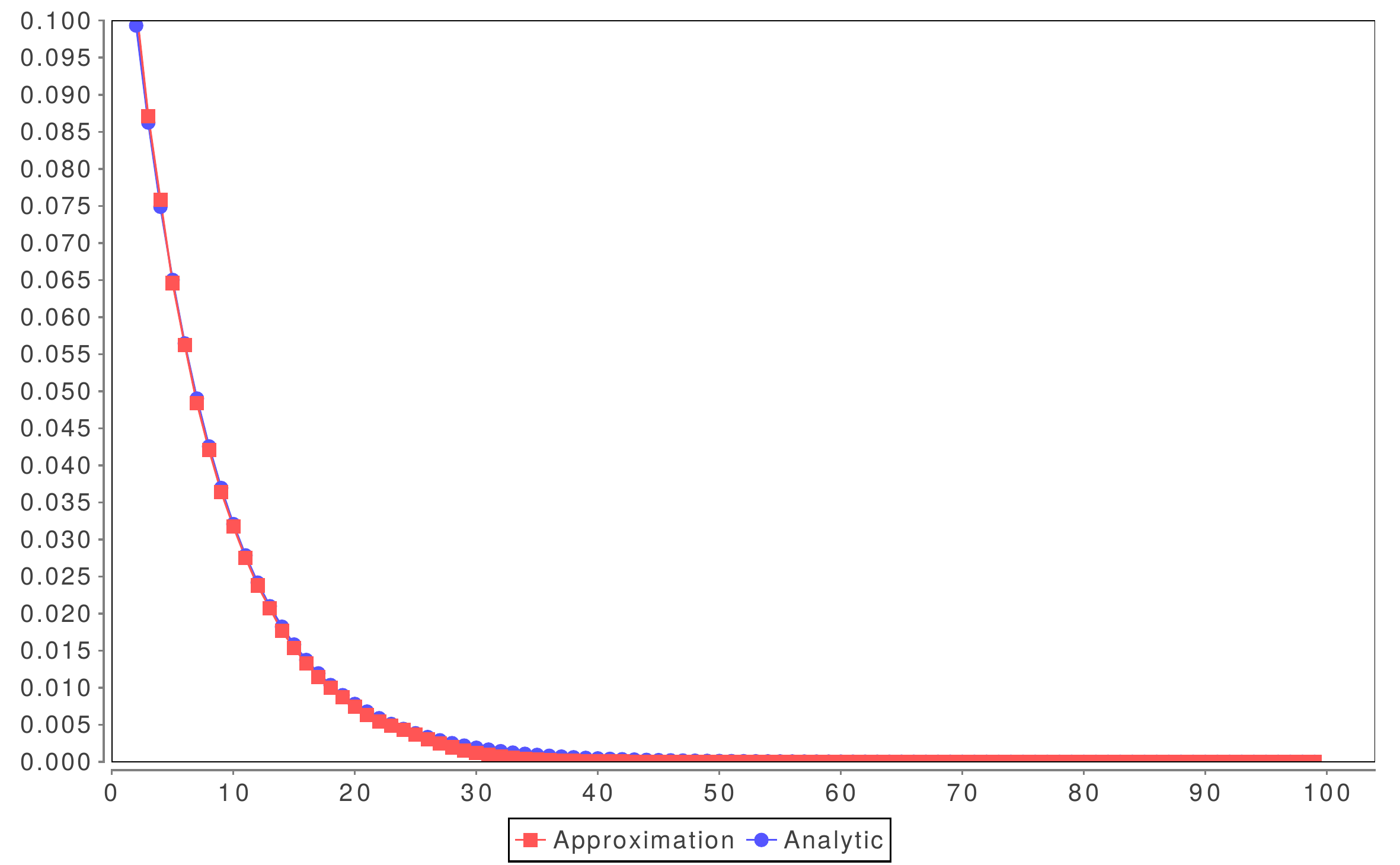} \label{fig:eigenvalues-empirical-200}} 
\subfloat[Spectrum $(n = 1000)$]{\includegraphics[width=0.5\columnwidth]{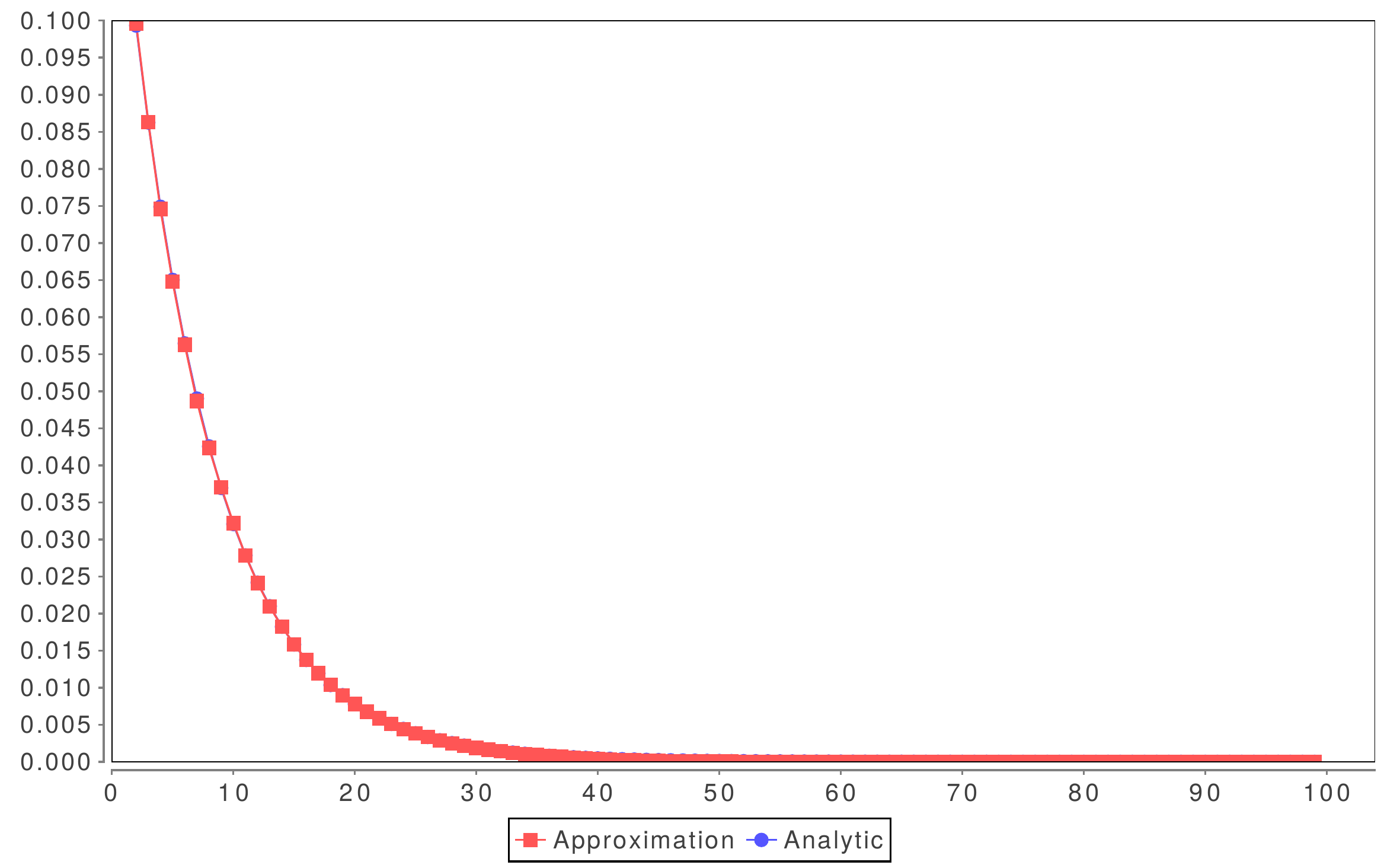} \label{fig:eigenvalues-empirical-1000}} 
\\
\subfloat[$\phi_1: n = 200$]{\includegraphics[width=0.5\columnwidth]{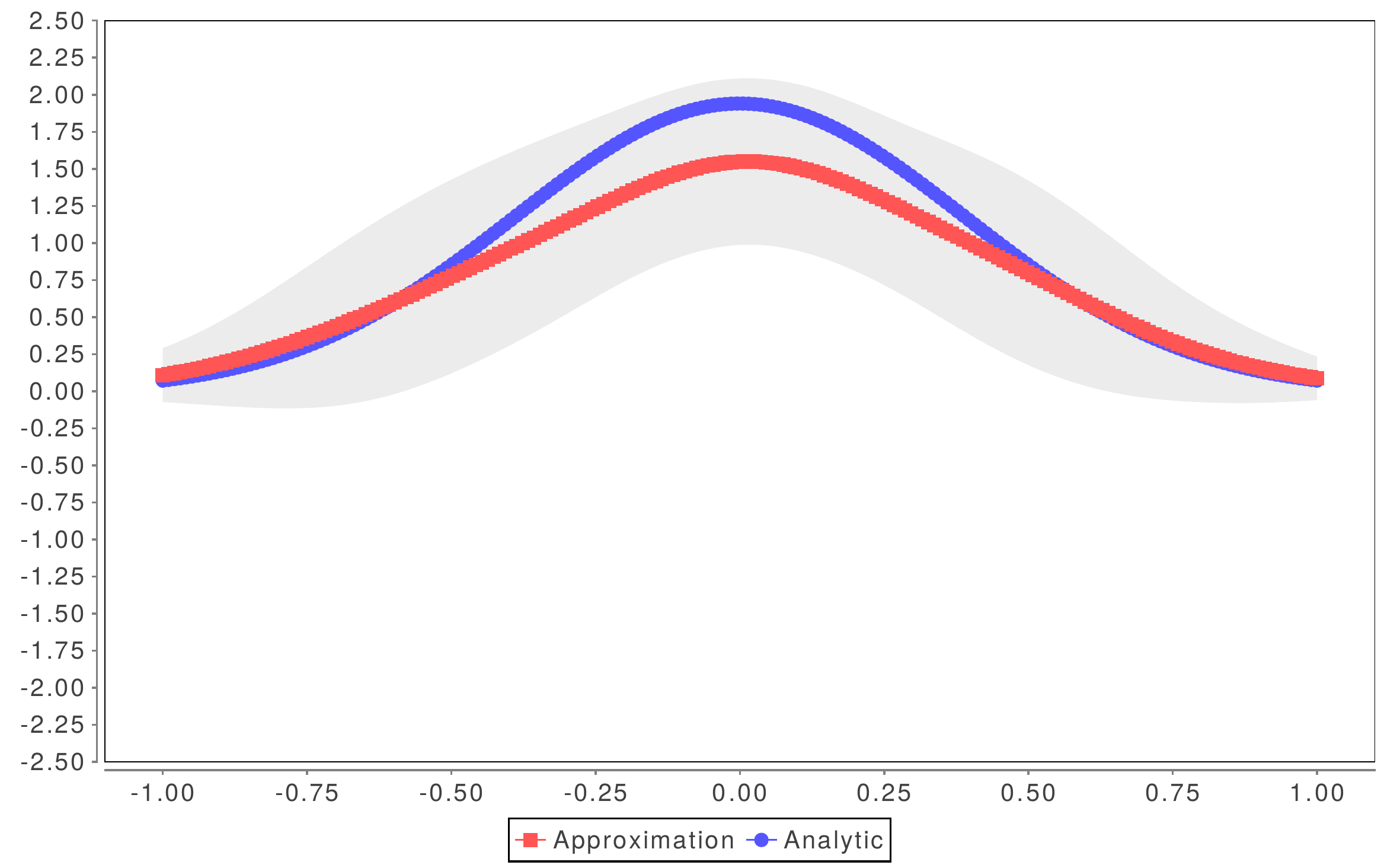} \label{fig:eigenfunctions-empirical-200-0-sigma02}}
\subfloat[$\phi_1: n = 1000$]{\includegraphics[width=0.5\columnwidth]{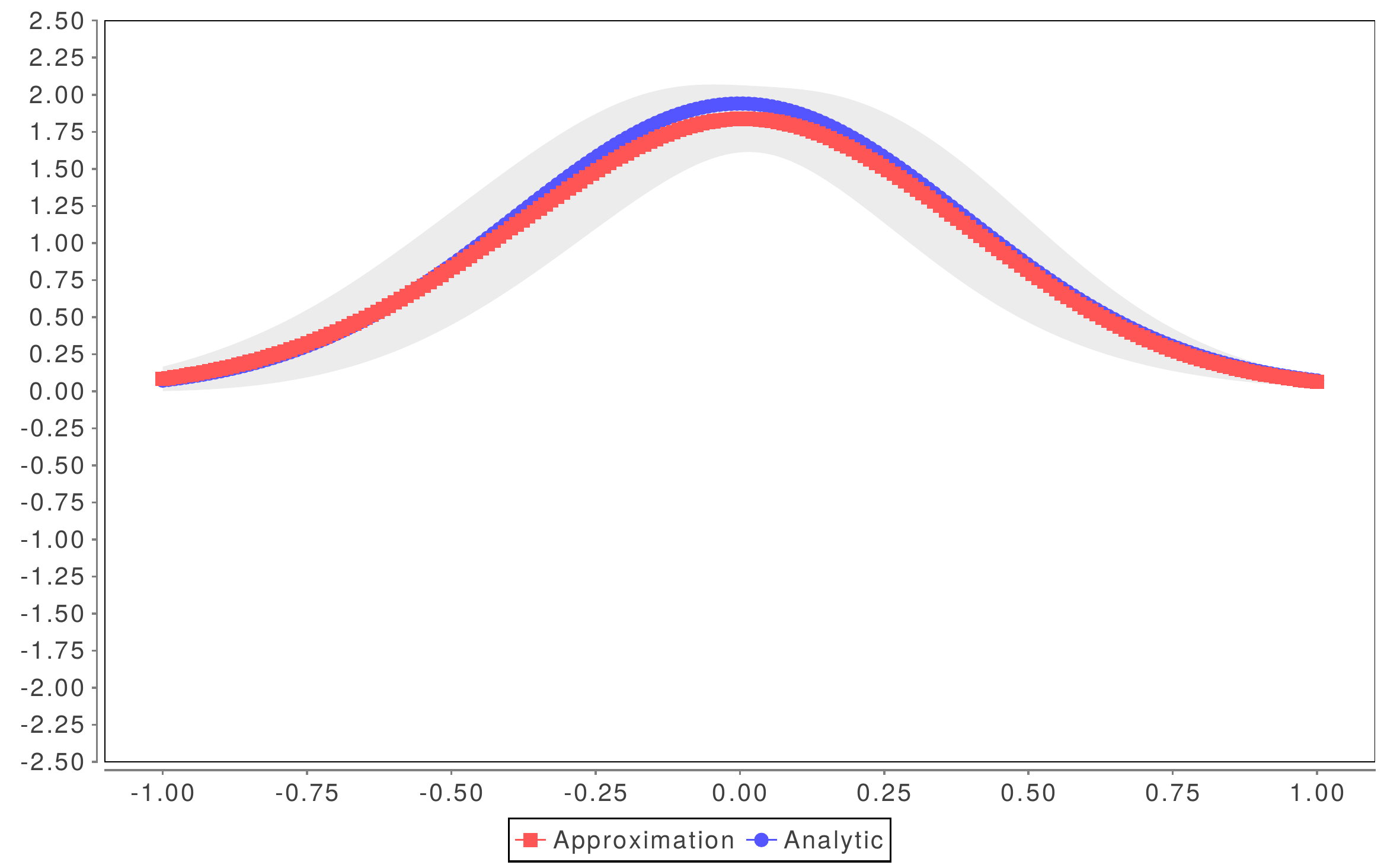} \label{fig:eigenfunctions-empirical-1000-0-sigma02}}
\\
\subfloat[$\phi_{10}: n = 200$]{\includegraphics[width=0.5\columnwidth]{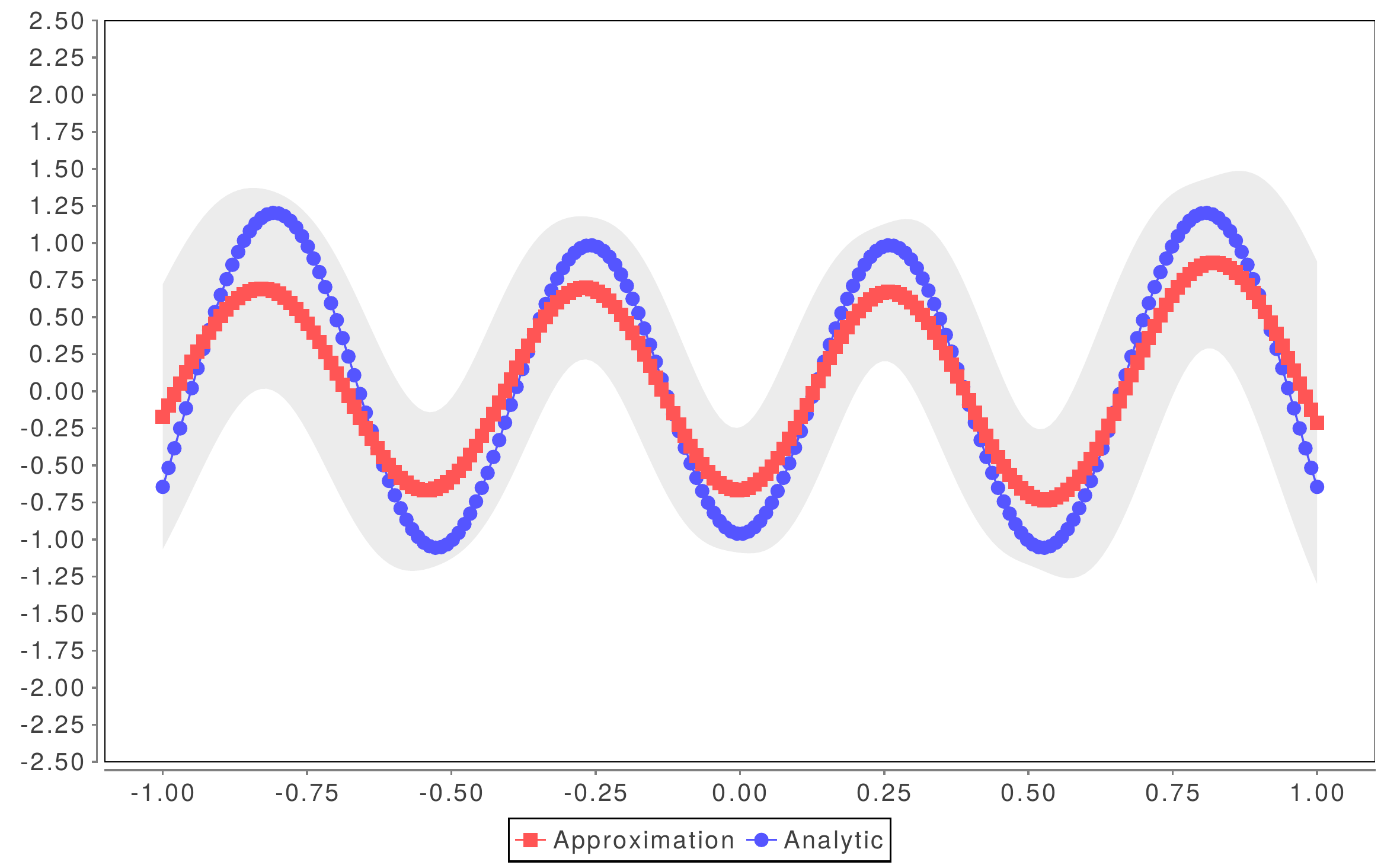} \label{fig:eigenfunctions-empirical-200-10-sigma02}}
\subfloat[$\phi_{10}: n = 1000$]{\includegraphics[width=0.5\columnwidth]{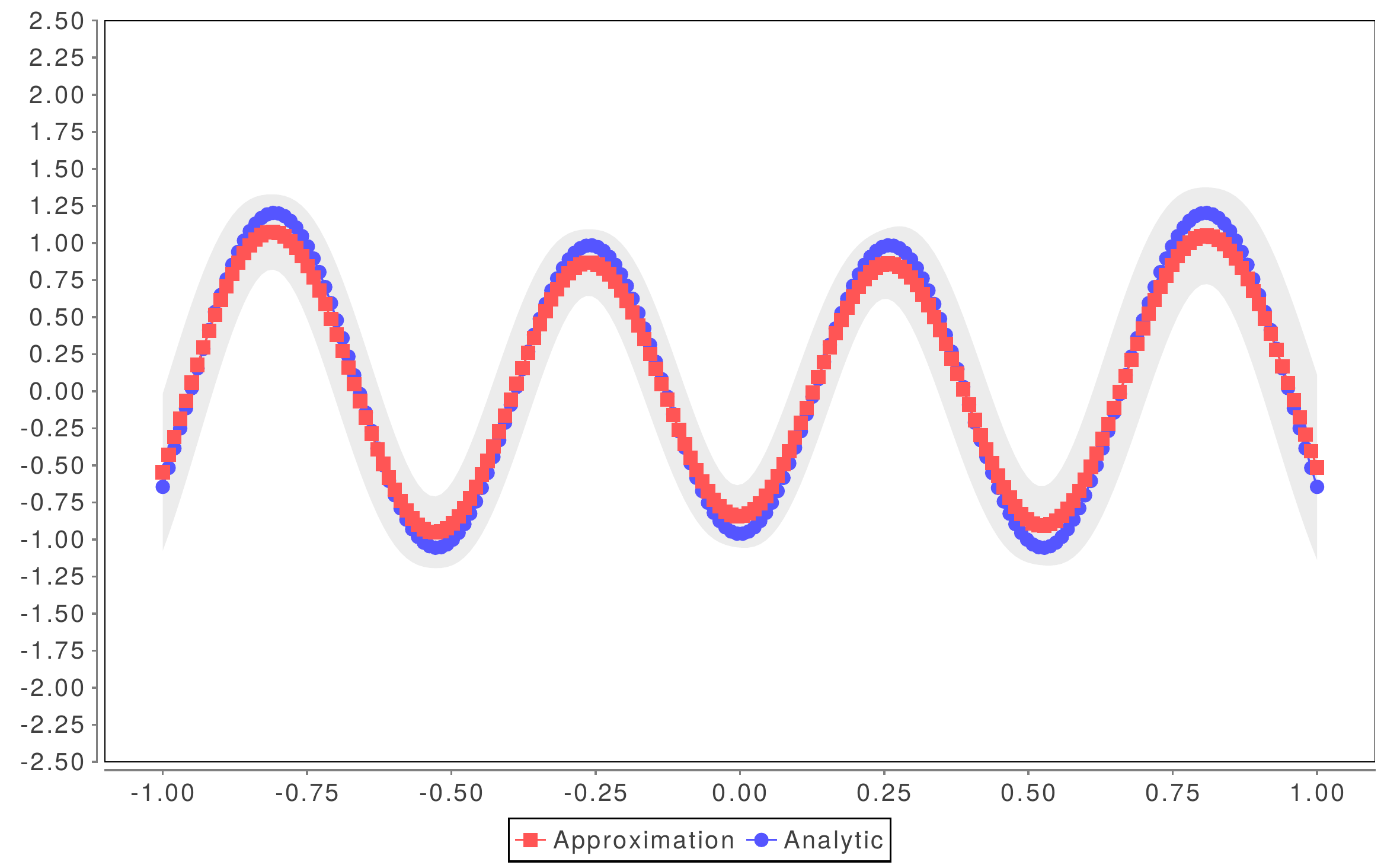} \label{fig:eigenfunctions-empirical-1000-10-sigma02}}
\\
\subfloat[$\phi_{40}: n = 200$]{\includegraphics[width=0.5\columnwidth]{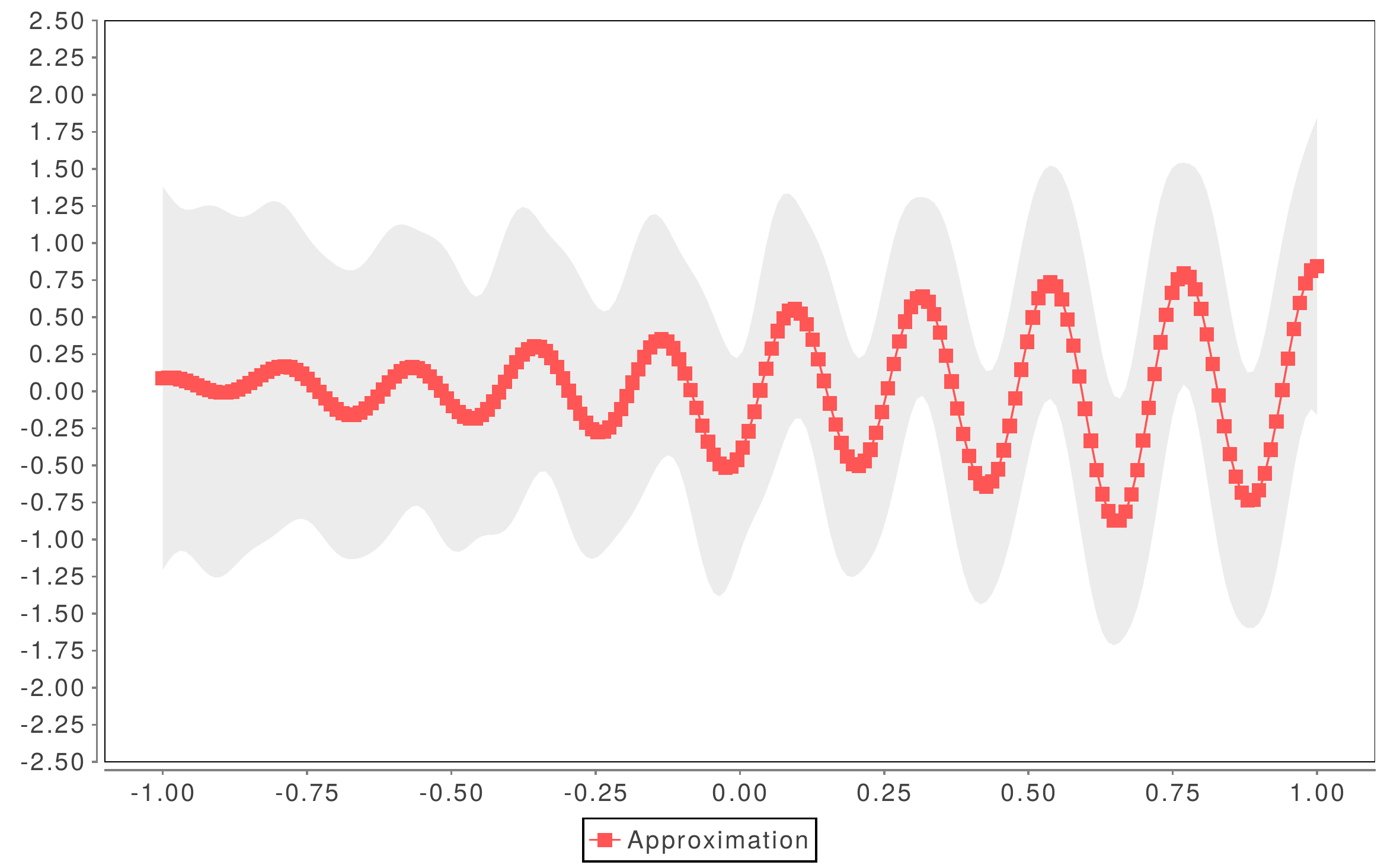} \label{fig:eigenfunctions-empirical-200-40-sigma02}}
\subfloat[$\phi_{40}: n = 1000$]{\includegraphics[width=0.5\columnwidth]{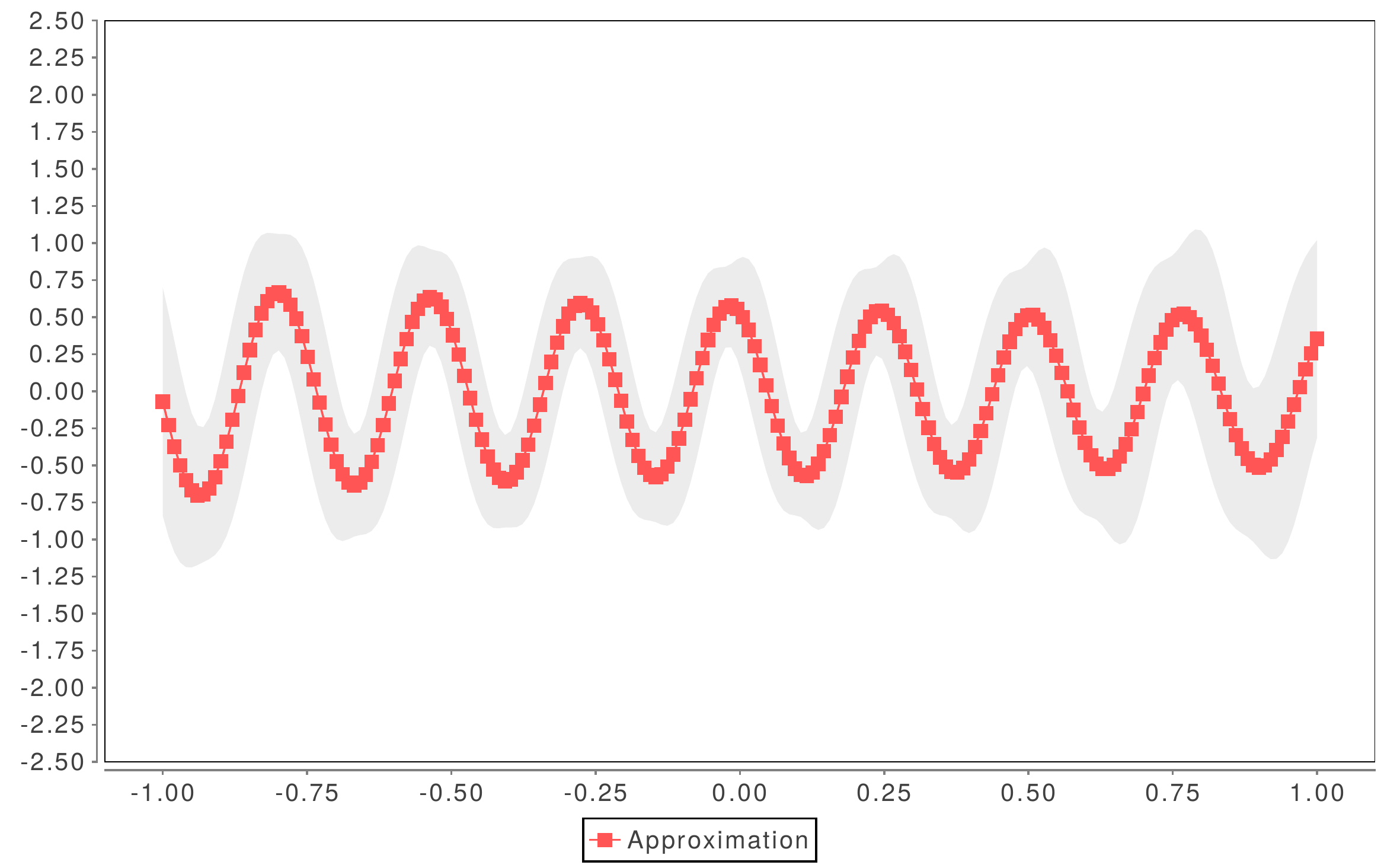} \label{fig:eigenfunctions-empirical-1000-40-sigma02}}
\caption{Estimation of the eigenvalues and eigenfunctions. The plots show the average and standard-deviation estimated from 100 runs, with parameters $\sigma=1$ and $\sigma=0.2$.
  The blue line shows the analytic solution. (The analytic solution for $\phi_{40}$ is not available, as the numeric computation of the $40-$th hermite polynomial is not feasible.)}
\label{fig:eigenfunctions-empirical}
\end{figure}
Figure~\ref{fig:eigenfunctions-empirical} shows approximations of the eigenvalues and eigenfunctions for the case where $\sigma=1$ and $\sigma=0.2$. We see that the eigenvalues are extremely well approximated, even for $n=200$. As the bound for the eigenfunctions predicts,  the eigenfunctions are approximated reasonably well as long as the spectrum decays quickly. Where the decay is slow (i.e.\ for eigenfunction $\phi_{40}$ the approximation with $n=200$ is not good enough anymore and more points need to be chosen to obtain a reasonable approximation. In this case, $n=1000$ leads to a satisfactory approximation. We also see in this example that the bounds are much too pessimistic. For $n=200$, Theorem~\ref{thm:bound-eigenvalues} asserts that with probability $0.99$, the largest difference between the true and the approximated eigenvalues $\sup_j (\lambda_j - \hat{\lambda}_j)$ for $n=200$ is $0.48$ and for $n=1000$ is $0.2$, where in reality the values are at least an order of magnitude better.

\subsection{Choosing the approximation parameters}
\begin{figure}
\subfloat[$\sigma=0.005$]{\includegraphics[width=0.5\columnwidth]{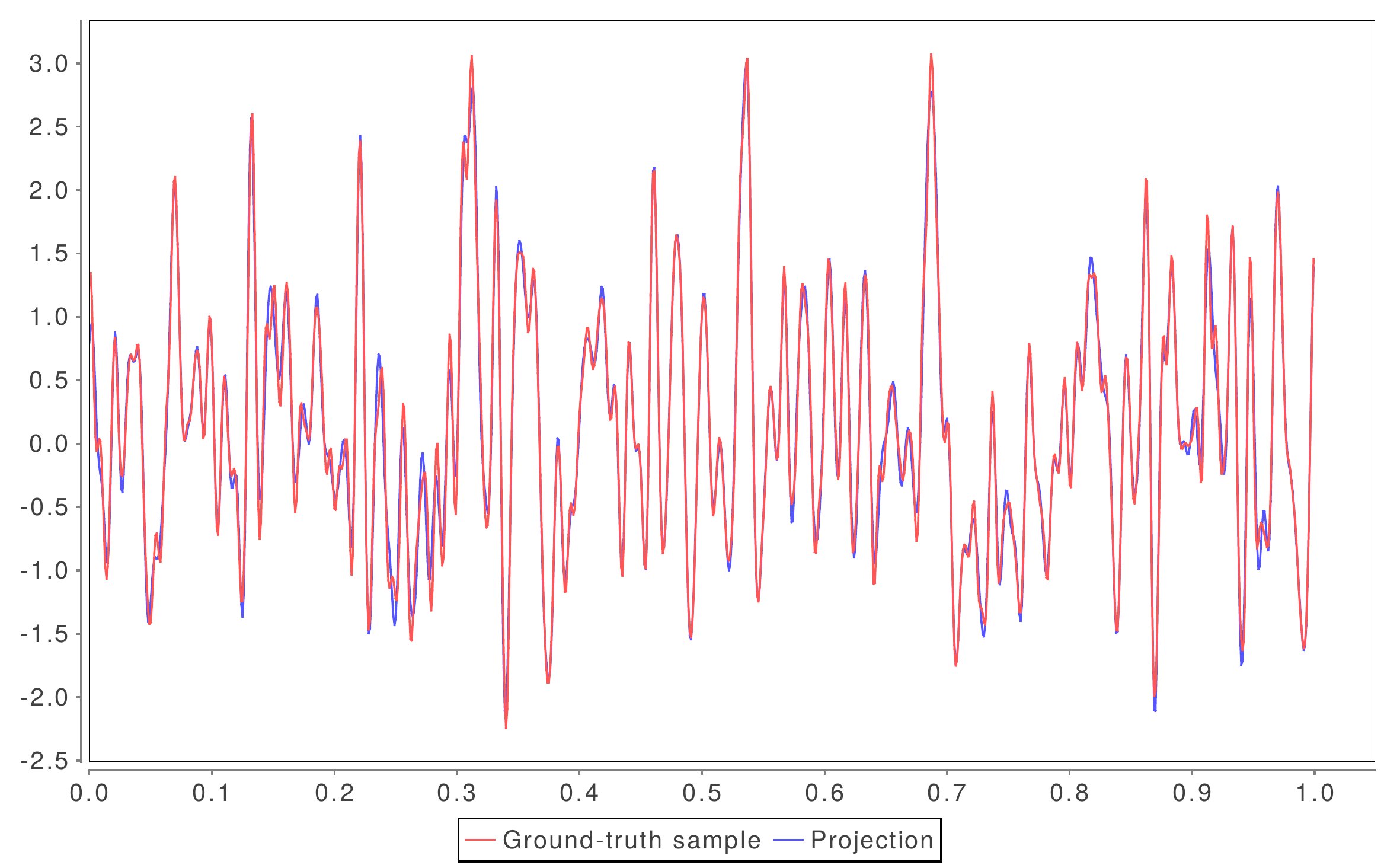} \label{fig:sample-gpwiggly}}
\subfloat[$\sigma=0.5$]{\includegraphics[width=0.5\columnwidth]{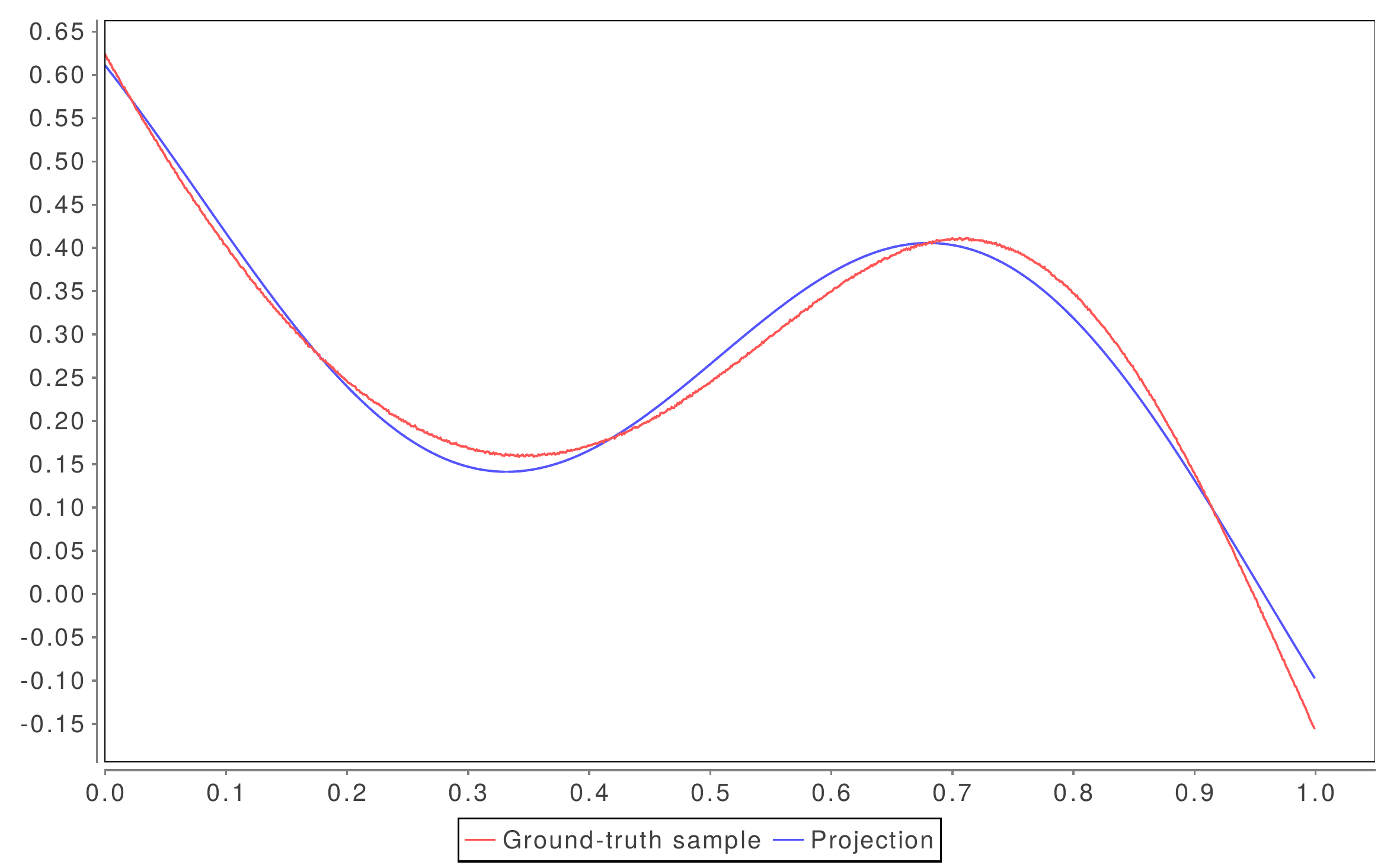}\label{fig:sample-gpsmooth}}\\
\subfloat[]{\includegraphics[width=\columnwidth]{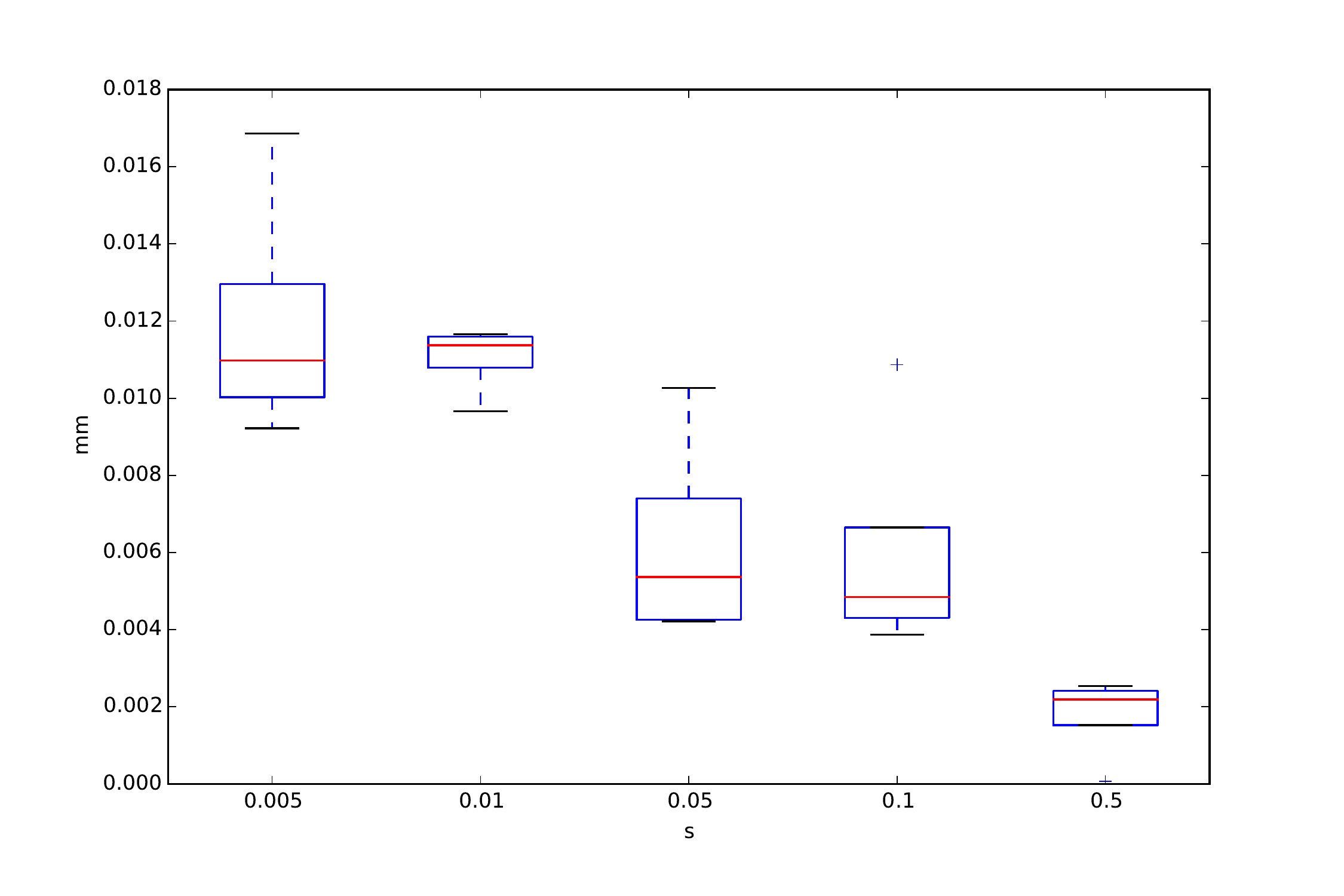}\label{fig:overall-error-estimate}}
\caption{\protect\subref{fig:sample-gpwiggly}, \protect\subref{fig:sample-gpsmooth}: Random samples from the Gaussian process with kernel $k(x,x') = \exp({-\norm{x-x'/\sigma^2}})$, for $\sigma=0.005$ and $\sigma=0.5$. 
\protect\subref{fig:overall-error-estimate} The projection error of random samples when using a low-rank approximation that covers $99\%$ of the variance. We see that despite the various numerical approximations, the desired approximation error of $0.01$ is attained even for functions that enforce very little smoothness.} 
\label{fig:approx-accuracy}
\end{figure}

The above considerations allow us to come up with an approach of choosing the parameters for the approximation. 
We suggest the following procedure:
\begin{enumerate}
\item Define a Gaussian process model $\GP(\mu, k)$, which represents the prior knowledge
\item Determine the total variance that should be covered by the approximation
\item Compute from the eigenvalues the number of eigenfunctions $r$ that are needed to retain a certain fraction of the variance,
  such that the desired fraction of variance $p$ from the total variance $var(k)$ is retained, i.e. 
\[
r = \argmin_m \frac{\sum_{j=1}^m \hat{\lambda}_j }{var(k)} > p.
\]
 \label{enum:fraction-variance}
\item Choose the number of points that we need for the Nystr\"om approximation, such that the $i-$th eigenfunction, for which 
  $\lambda_i - \lambda_{i+1}$ is minimal, leads to a stable approximation.\label{enum:nystrom}
\end{enumerate}
The total variance $var(k)$ in step~\ref{enum:fraction-variance} can be estimated  by noting that
\[
var(k) = \sum_{i=1}^\infty \lambda_i = \int_{\Omega}k(x,x)\, dp(x),
\]
and use numerical integration to estimate the value of the
integral. For stationary kernels, this last step can even be avoided,
by noting that $k(x,x)$ is constant and hence the total sum is simply
$k(x,x)$ for some $x \in \Omega$. For estimating the leading
eigenvalues, we use the Nystr\"om approximation, for which we know
that it can accurately estimate the eigenvalues.  Choosing the number
of points for the Nystr\"om approximation (step~\ref{enum:nystrom}) is
more difficult. The bound given in
Theorem~\ref{thm:bound-eigenfunctions} is not tight enough to be of
practical use. To determine whether we have chosen sufficiently many points for the
Nystr\"om approximation, we propose the following
experiment: We define the full Gaussian process
$\GP(\mu,k)$ and draw random samples for a discrete number of points $x_1, \ldots, x_N \subset \Omega$, toobtain the sample values $\hat{u}(x_1), \ldots, \hat{u}(x_N)$. 
This can be done by sampling from the multivariate normal distribution, $\N(\mu, K)$, where K is the Kernel matrix $K_{ij}=k(x_i, x_j), \; i,j=1 \ldots, N$. 
The overall error of the low rank approximation can be assessed by determining how well we can approximate the samples with the low-rank Gaussian process $\GP(\mu, \tilde{k})$. 
This can be efficiently done by performing a Gaussian process regression
with training data $\{(x_1, \hat{u}(x_1),\ldots, (x_N, \hat{u}(x_N)\}$
(see e.g. \cite{rasmussen_gaussian_2006}, Chapter 2 for
details). Figure~\ref{fig:approx-accuracy} shows this approach for a 1D example. We have tested an approximation that covers $99 \%$ of the variance for different kernels. The error was estimated using $1000$ points of the domain. 
We see that for $\sigma=0.005$, for which the sample is very wiggly (Figure~\ref{fig:sample-gpwiggly}), the approximation error is slightly larger
than the $1\%$, and hence we would need to increase the number of points used for the Nystr\"om approximation. For all other kernels, the error is very close or even below the desired $1\%$ approximation error. That the value is sometimes smaller than $1 \%$ is because the eigenvalues are discrete, and hence we might actually cover more than $99 \%$ by choosing the $r$ first eigenfunctions. 



  \section*{Acknowledgment}

This work has been funded as part of the Swiss National Science foundation project in the context of the project
SNF153297. We thank Sandro Sch\"onborn and Volker Roth for interesting and enlightning discussion. 
A special thanks goes to Ghazi Bouabene and Christoph Langguth, for their work on the  scalismo software, in which all the methods are implemented.





%
\bibliographystyle{IEEEtran}
\bibliography{ref}

\begin{thebibliography}{10}
\providecommand{\url}[1]{#1}
\csname url@samestyle\endcsname
\providecommand{\newblock}{\relax}
\providecommand{\bibinfo}[2]{#2}
\providecommand{\BIBentrySTDinterwordspacing}{\spaceskip=0pt\relax}
\providecommand{\BIBentryALTinterwordstretchfactor}{4}
\providecommand{\BIBentryALTinterwordspacing}{\spaceskip=\fontdimen2\font plus
\BIBentryALTinterwordstretchfactor\fontdimen3\font minus
  \fontdimen4\font\relax}
\providecommand{\BIBforeignlanguage}[2]{{%
\expandafter\ifx\csname l@#1\endcsname\relax
\typeout{** WARNING: IEEEtran.bst: No hyphenation pattern has been}%
\typeout{** loaded for the language `#1'. Using the pattern for}%
\typeout{** the default language instead.}%
\else
\language=\csname l@#1\endcsname
\fi
#2}}
\providecommand{\BIBdecl}{\relax}
\BIBdecl

\bibitem{grenander2007pattern}
U.~Grenander and M.~I. Miller, \emph{Pattern theory: from representation to
  inference}.\hskip 1em plus 0.5em minus 0.4em\relax Oxford university press
  Oxford, 2007, vol.~1.

\bibitem{cootes_active_1995}
T.~F. Cootes, C.~J. Taylor, D.~H. Cooper, J.~Graham, and {others}, ``Active
  shape models-their training and application,'' \emph{Computer Vision and
  Image Understanding}, vol.~61, no.~1, 1995.

\bibitem{blanz_morphable_1999}
V.~Blanz and T.~Vetter, ``A morphable model for the synthesis of 3d faces,'' in
  \emph{{SIGGRAPH} '99: Proceedings of the 26th annual conference on Computer
  graphics and interactive techniques}.\hskip 1em plus 0.5em minus 0.4em\relax
  {ACM} Press, 1999, pp. 187--194.

\bibitem{li_making_2010}
M.~Li and J.~T.-Y. Kwok, ``Making large-scale nystrom approximation possible,''
  2010.

\bibitem{myronenko2010point}
A.~Myronenko and X.~Song, ``Point set registration: Coherent point drift,''
  \emph{Pattern Analysis and Machine Intelligence, IEEE Transactions on},
  vol.~32, no.~12, pp. 2262--2275, 2010.

\bibitem{luthi_statismo-framework_2012}
M.~L{\"u}thi, R.~Blanc, T.~Albrecht, T.~Gass, O.~Goksel, P.~Büchler,
  M.~Kistler, H.~Bousleiman, M.~Reyes, P.~C. Cattin, and {others}, ``Statismo-a
  framework for {PCA} based statistical models.'' \emph{Insight Journal}, 2012.

\bibitem{scalismo}
``Scalismo - scalable image analysis and shape modelling,''
  http://github.com/unibas-gravis/scalismo.

\bibitem{wang_boundary_2000}
Y.~Wang and L.~H. Staib, ``Boundary finding with prior shape and smoothness
  models,'' \emph{{IEEE} Transactions on Pattern Analysis and Machine
  Intelligence}, vol.~22, no.~7, 2000.

\bibitem{grenander_computational_1998}
U.~Grenander and M.~I. Miller, ``Computational anatomy: An emerging
  discipline,'' \emph{Quarterly of applied mathematics}, vol.~56, no.~4, pp.
  617--694, 1998.

\bibitem{amit_structural_1991}
Y.~Amit, U.~Grenander, and M.~Piccioni, ``Structural image restoration through
  deformable templates,'' \emph{Journal of the American Statistical
  Association}, vol.~86, no. 414, pp. 376--387, 1991.

\bibitem{joshi1997gaussian}
S.~C. Joshi, A.~Banerjee, G.~E. Christensen, J.~G. Csernansky, J.~W. Haller,
  M.~I. Miller, and L.~Wang, ``Gaussian random fields on sub-manifolds for
  characterizing brain surfaces,'' in \emph{Information Processing in Medical
  Imaging}.\hskip 1em plus 0.5em minus 0.4em\relax Springer, 1997, pp.
  381--386.

\bibitem{holden2008review}
M.~Holden, ``A review of geometric transformations for nonrigid body
  registration,'' \emph{IEEE TRANSACTIONS ON MEDICAL IMAGING}, vol.~27, no.~1,
  p. 111, 2008.

\bibitem{rasmussen_gaussian_2006}
C.~E. Rasmussen and C.~K. Williams, \emph{Gaussian processes for machine
  learning}.\hskip 1em plus 0.5em minus 0.4em\relax Springer, 2006.

\bibitem{zhu2009nonrigid}
J.~Zhu, S.~C. Hoi, and M.~R. Lyu, ``Nonrigid shape recovery by gaussian process
  regression,'' in \emph{Computer Vision and Pattern Recognition, 2009. CVPR
  2009. IEEE Conference on}.\hskip 1em plus 0.5em minus 0.4em\relax IEEE, 2009,
  pp. 1319--1326.

\bibitem{scholkopf_object_2005}
B.~Sch{\"o}lkopf, F.~Steinke, and V.~Blanz, ``Object correspondence as a
  machine learning problem,'' in \emph{{ICML} '05: Proceedings of the 22nd
  international conference on Machine learning}.\hskip 1em plus 0.5em minus
  0.4em\relax New York, {NY}, {USA}: {ACM} Press, 2005, pp. 776--783.

\bibitem{younes2010shapes}
L.~Younes, \emph{Shapes and diffeomorphisms}.\hskip 1em plus 0.5em minus
  0.4em\relax Springer, 2010, vol. 171.

\bibitem{bruveris2012mixture}
M.~Bruveris, L.~Risser, and F.-X. Vialard, ``Mixture of kernels and iterated
  semidirect product of diffeomorphisms groups,'' \emph{Multiscale Modeling \&
  Simulation}, vol.~10, no.~4, pp. 1344--1368, 2012.

\bibitem{sommer2013sparse}
S.~Sommer, F.~Lauze, M.~Nielsen, and X.~Pennec, ``Sparse multi-scale
  diffeomorphic registration: the kernel bundle framework,'' \emph{Journal of
  mathematical imaging and vision}, vol.~46, no.~3, pp. 292--308, 2013.

\bibitem{schmah2013left}
T.~Schmah, L.~Risser, and F.-X. Vialard, ``Left-invariant metrics for
  diffeomorphic image registration with spatially-varying regularisation,'' in
  \emph{Medical Image Computing and Computer-Assisted Intervention--MICCAI
  2013}.\hskip 1em plus 0.5em minus 0.4em\relax Springer, 2013, pp. 203--210.

\bibitem{cootes_combining_1995}
T.~F. Cootes and C.~J. Taylor, ``Combining point distribution models with shape
  models based on finite element analysis,'' \emph{Image and Vision Computing},
  vol.~13, no.~5, 1995.

\bibitem{zhao2005novel}
Z.~Zhao, S.~R. Aylward, and E.~K. Teoh, ``A novel 3d partitioned active shape
  model for segmentation of brain mr images,'' in \emph{Medical Image Computing
  and Computer-Assisted Intervention--MICCAI 2005}.\hskip 1em plus 0.5em minus
  0.4em\relax Springer, 2005, pp. 221--228.

\bibitem{davatzikos2003hierarchical}
C.~Davatzikos, X.~Tao, and D.~Shen, ``Hierarchical active shape models, using
  the wavelet transform,'' \emph{Medical Imaging, IEEE Transactions on},
  vol.~22, no.~3, pp. 414--423, 2003.

\bibitem{nain2007multiscale}
D.~Nain, S.~Haker, A.~Bobick, and A.~Tannenbaum, ``Multiscale 3-d shape
  representation and segmentation using spherical wavelets,'' \emph{Medical
  Imaging, IEEE Transactions on}, vol.~26, no.~4, pp. 598--618, 2007.

\bibitem{albrecht2008statistical}
T.~Albrecht, M.~L{\"u}thi, and T.~Vetter, ``A statistical deformation prior for
  non-rigid image and shape registration,'' in \emph{Computer Vision and
  Pattern Recognition, 2008. CVPR 2008. IEEE Conference on}.\hskip 1em plus
  0.5em minus 0.4em\relax IEEE, 2008, pp. 1--8.

\bibitem{kainmueller2013omnidirectional}
D.~Kainmueller, H.~Lamecker, M.~O. Heller, B.~Weber, H.-C. Hege, and S.~Zachow,
  ``Omnidirectional displacements for deformable surfaces,'' \emph{Medical
  image analysis}, vol.~17, no.~4, pp. 429--441, 2013.

\bibitem{le_pdm-enlor:_2013}
Y.~Le, U.~Kurkure, and I.~Kakadiaris, ``{PDM}-{ENLOR}: Learning ensemble of
  local {PDM}-based regressions,'' in \emph{2013 {IEEE} Conference on Computer
  Vision and Pattern Recognition ({CVPR})}, Jun. 2013, pp. 1878--1885.

\bibitem{luthi_using_2011}
M.~L{\"u}thi, C.~Jud, and T.~Vetter, ``Using landmarks as a deformation prior
  for hybrid image registration,'' \emph{Pattern Recognition}, pp. 196--205,
  2011.

\bibitem{gerig2014spatially}
T.~Gerig, K.~Shahim, M.~Reyes, T.~Vetter, and M.~L{\"u}thi, ``Spatially varying
  registration using gaussian processes,'' in \emph{Medical Image Computing and
  Computer-Assisted Intervention--MICCAI 2014}.\hskip 1em plus 0.5em minus
  0.4em\relax Springer, 2014, pp. 413--420.

\bibitem{luthi2013unified}
M.~L{\"u}thi, C.~Jud, and T.~Vetter, ``A unified approach to shape model
  fitting and non-rigid registration,'' in \emph{Machine Learning in Medical
  Imaging}.\hskip 1em plus 0.5em minus 0.4em\relax Springer, 2013, pp. 66--73.

\bibitem{jolliffe2002principal}
I.~Jolliffe, \emph{Principal component analysis}.\hskip 1em plus 0.5em minus
  0.4em\relax Wiley Online Library, 2002.

\bibitem{berlinet2004reproducing}
A.~Berlinet and C.~Thomas-Agnan, \emph{Reproducing kernel Hilbert spaces in
  probability and statistics}.\hskip 1em plus 0.5em minus 0.4em\relax Springer,
  2004, vol.~3.

\bibitem{halko_finding_2011}
\BIBentryALTinterwordspacing
N.~Halko, P.-G. Martinsson, and J.~A. Tropp, ``Finding structure with
  randomness: Probabilistic algorithms for constructing approximate matrix
  decompositions,'' \emph{{SIAM} review}, vol.~53, no.~2, pp. 217--288, 2011.
  [Online]. Available: \url{http://epubs.siam.org/doi/abs/10.1137/090771806}
\BIBentrySTDinterwordspacing

\bibitem{duvenaud2014automatic}
D.~Duvenaud, ``Automatic model construction with gaussian processes,'' Ph.D.
  dissertation, University of Cambridge, 2014.

\bibitem{paysan20093d}
P.~Paysan, R.~Knothe, B.~Amberg, S.~Romdhani, and T.~Vetter, ``A 3d face model
  for pose and illumination invariant face recognition,'' in \emph{Advanced
  Video and Signal Based Surveillance, 2009. AVSS'09. Sixth IEEE International
  Conference On}.\hskip 1em plus 0.5em minus 0.4em\relax IEEE, 2009, pp.
  296--301.

\bibitem{davis1995elastic}
M.~H. Davis, A.~Khotanzad, D.~P. Flamig, and S.~E. Harms, ``Elastic body
  splines: a physics based approach to coordinate transformation in medical
  image matching,'' in \emph{Computer-Based Medical Systems, 1995., Proceedings
  of the Eighth IEEE Symposium on}.\hskip 1em plus 0.5em minus 0.4em\relax
  IEEE, 1995, pp. 81--88.

\bibitem{kybic2003fast}
J.~Kybic and M.~Unser, ``Fast parametric elastic image registration,''
  \emph{Image Processing, IEEE Transactions on}, vol.~12, no.~11, pp.
  1427--1442, 2003.

\bibitem{rohr2001landmark}
K.~Rohr, H.~S. Stiehl, R.~Sprengel, T.~M. Buzug, J.~Weese, and M.~Kuhn,
  ``Landmark-based elastic registration using approximating thin-plate
  splines,'' \emph{Medical Imaging, IEEE Transactions on}, vol.~20, no.~6, pp.
  526--534, 2001.

\bibitem{shawe2004kernel}
J.~Shawe-Taylor and N.~Cristianini, \emph{Kernel methods for pattern
  analysis}.\hskip 1em plus 0.5em minus 0.4em\relax Cambridge university press,
  2004.

\bibitem{opfer_multiscale_2006}
R.~Opfer, ``Multiscale kernels,'' \emph{Advances in Computational Mathematics},
  vol.~25, no.~4, pp. 357--380, 2006.

\bibitem{xu2009refinement}
Y.~Xu and H.~Zhang, ``Refinement of reproducing kernels,'' \emph{The Journal of
  Machine Learning Research}, vol.~10, pp. 107--140, 2009.

\bibitem{wahba_spline_1990}
G.~Wahba, \emph{Spline models for observational data}.\hskip 1em plus 0.5em
  minus 0.4em\relax Society for Industrial Mathematics, 1990.

\bibitem{sotiras2013deformable}
A.~Sotiras, C.~Davatzikos, and N.~Paragios, ``Deformable medical image
  registration: A survey,'' \emph{Medical Imaging, IEEE Transactions on},
  vol.~32, no.~7, pp. 1153--1190, 2013.

\bibitem{klein_evaluation_2007}
S.~Klein, M.~Staring, and J.~P. Pluim, ``Evaluation of optimization methods for
  nonrigid medical image registration using mutual information and b-splines,''
  \emph{Image Processing, {IEEE} Transactions on}, vol.~16, no.~12, pp.
  2879--2890, 2007.

\bibitem{johnson_consistent_2002}
H.~J. Johnson and G.~E. Christensen, ``Consistent landmark and intensity-based
  image registration,'' \emph{Medical Imaging, {IEEE} Transactions on},
  vol.~21, no.~5, pp. 450--461, 2002.

\bibitem{fischer_combination_2003}
B.~Fischer and J.~Modersitzki, ``Combination of automatic non-rigid and
  landmark based registration: the best of both worlds,'' \emph{Medical
  imaging}, pp. 1037--1048, 2003.

\bibitem{biesdorf_hybrid_2009}
A.~Biesdorf, S.~W{\textbackslash}örz, H.~J. Kaiser, C.~Stippich, and K.~Rohr,
  ``Hybrid spline-based multimodal registration using local measures for joint
  entropy and mutual information,'' \emph{Medical Image Computing and
  Computer-Assisted Intervention–{MICCAI} 2009}, pp. 607--615, 2009.

\bibitem{lorensen1987marching}
W.~E. Lorensen and H.~E. Cline, ``Marching cubes: A high resolution 3d surface
  construction algorithm,'' in \emph{ACM siggraph computer graphics}, vol.~21,
  no.~4.\hskip 1em plus 0.5em minus 0.4em\relax ACM, 1987, pp. 163--169.

\bibitem{umeyama1991least}
S.~Umeyama, ``Least-squares estimation of transformation parameters between two
  point patterns,'' \emph{IEEE Transactions on Pattern Analysis \& Machine
  Intelligence}, no.~4, pp. 376--380, 1991.

\bibitem{klein2010elastix}
S.~Klein, M.~Staring, K.~Murphy, M.~Viergever, J.~Pluim \emph{et~al.},
  ``Elastix: a toolbox for intensity-based medical image registration,''
  \emph{IEEE transactions on medical imaging}, vol.~29, no.~1, pp. 196--205,
  2010.

\bibitem{styner2003evaluation}
M.~A. Styner, K.~T. Rajamani, L.-P. Nolte, G.~Zsemlye, G.~Sz{\'e}kely, C.~J.
  Taylor, and R.~H. Davies, ``Evaluation of 3d correspondence methods for model
  building,'' in \emph{Information processing in medical imaging}.\hskip 1em
  plus 0.5em minus 0.4em\relax Springer, 2003, pp. 63--75.

\bibitem{rueckert_automatic_2001}
D.~Rueckert, A.~F. Frangi, and J.~A. Schnabel, ``Automatic construction of 3d
  statistical deformation models using non-rigid registration,'' in
  \emph{{MICCAI} '01: Medical Image Computing and Computer-Assisted
  Intervention}, 2001, pp. 77--84.

\bibitem{zhu_gaussian_1998}
H.~Zhu, C.~K. Williams, R.~Rohwer, and M.~Morciniec, ``Gaussian regression and
  optimal finite dimensional linear models,'' \emph{Neural Networks and Machine
  Learning}, 1998.

\bibitem{williams_effect_2000}
C.~Williams and M.~Seeger, ``The effect of the input density distribution on
  kernel-based classifiers,'' in \emph{Proceedings of the 17th International
  Conference on Machine Learning}, 2000, pp. 1159--1166.

\bibitem{rosasco_learning_2010}
\BIBentryALTinterwordspacing
L.~Rosasco, M.~Belkin, and E.~D. Vito, ``On learning with integral operators,''
  \emph{The Journal of Machine Learning Research}, vol.~11, pp. 905--934, 2010.
  [Online]. Available: \url{http://dl.acm.org/citation.cfm?id=1756036}
\BIBentrySTDinterwordspacing

\end{thebibliography}

%








\end{document}